\documentclass{article}
\pdfoutput=1
\usepackage{arxiv}
\usepackage[utf8]{inputenc} 
\usepackage[T1]{fontenc}    
\usepackage{hyperref}       
\usepackage{url}            
\usepackage{booktabs}       
\usepackage{nicefrac}       
\usepackage{microtype}      
\usepackage{graphicx}
\usepackage{fullpage}
\usepackage{float}

\usepackage{amsmath,amssymb,amsfonts,amsthm}

\usepackage[titletoc,toc,title]{appendix}
\usepackage{textpos}
\usepackage{array} 
\usepackage{listings}
\usepackage{mathtools}
\usepackage{pdfpages}
\usepackage[textsize=footnotesize,color=green]{todonotes}
\usepackage{bm}
\usepackage{bbm}

\usepackage{tikz}
\usepackage[normalem]{ulem}
\usepackage{hhline}

\usepackage{graphicx}
\usepackage{subfig}
\usepackage{color}

\usepackage{pgfplots}
\usepackage{pgfplotstable}
\definecolor{markercolor}{RGB}{124.9, 255, 160.65}
\pgfplotsset{
compat=1.3,
width=10cm,
tick label style={font=\small},
label style={font=\small},
legend style={font=\small}
}
\newcommand{\myfrac}[2]{\displaystyle \frac{#1}{#2}}

\usetikzlibrary{calc}
\usetikzlibrary{intersections} 

\usepackage{stmaryrd}

\title{Physics-informed neural network for ultrasound nondestructive quantification of surface breaking cracks}

\author{ {Khemraj Shukla}\\
	Division of Applied Mathematics\\
	Brown University\\
	Providence, RI 02906\\
	\And
    {Patricio Clark Di Leoni} \\
	Department of Mechanical Engineering\\
	John Hopkins University\\
	Baltimore, MD 21218\\
	\And
    {James Blackshire} \\
	Air Force Research Laboratory\\
	OH 45433
	\And
    {Daniel Sparkman} \\
	Air Force Research Laboratory\\
	OH 45433
	\And
    {George Em Karniadakis}\thanks{Corresponding author.\textit{E-mail address:} george\_karniadakis@brown.edu}\\
    Division of Applied Mathematics\\
	Brown University\\
	Providence, RI 02906\\
}

\begin{document}
\maketitle

\begin{abstract}
We introduce an optimized physics-informed neural network (PINN)  trained to solve the problem of identifying and characterizing a surface breaking crack in a metal plate. PINNs are neural networks that can combine data and physics in the learning process by adding the residuals of a system of Partial Differential Equations to the loss function. Our PINN is supervised with realistic ultrasonic surface acoustic wave data acquired at a frequency of 5 MHz. The ultrasonic surface wave data is represented as a surface deformation on the top surface of a metal plate, measured by using the method of laser vibrometry. The PINN is physically informed by the acoustic wave equation and its convergence is sped up using adaptive activation functions. The adaptive activation function 
uses a scalable hyperparameter in the activation function, which is optimized to achieve best performance of the network as it changes
dynamically the topology of the loss function involved in the optimization process. The usage of adaptive activation function  significantly improves the convergence, notably observed in the current study. We use PINNs to estimate the speed of sound of the metal plate, which we do with an error of 1\%, and then, by allowing the speed of sound to be space dependent, we identify and characterize the crack as the positions where the speed of sound has decreased. Our study also shows the effect of sub-sampling of the data on the sensitivity of sound speed estimates. More broadly, the resulting model shows a promising deep neural network model for ill-posed inverse problems.
\end{abstract}
\section{Introduction}
The recent advances in the machine learning algorithms along with the growth in data and computing resources \cite{wei2019benchmarking} have transformative results across various scientific disciplines including but not limited to image recognition \cite{krizhevsky2012imagenet}, cognitive sciences \cite{lake2015human}, genomics \cite{alipanahi2015predicting} and bioinformatics \cite{nielsen2009nn}. Unlike these sciences, cost of acquisition of data becomes prohibitive while analyzing the physical and biological systems. The prohibitive nature of the data results in failures of a vast majority of state-of-the-art machine-learning algorithms. The failures are typically due to an unconstrained search of non-linear mapping between high dimensional (a large ratio between number of features per sample) input-output pairs of the data, resulting in a very large space of admissible solutions. To circumvent the effect of data sparsity of the data, Raissi et al. \cite{raissi2019physics} proposed a physics informed neural network (PINN), which uses the physical laws governing the dynamics of a time dependent system as a regularization agent resulting in a reduction of the space of admissible solutions to a manageable size. Since the advent of the PINN, applicability of PINN is being explored for various real world problems and based on this premise, our work in this paper presents applicability of PINN to solve the problems of wavefield imaging at ultrasonic frequencies.

Wavefield imaging methods have acquired recent popularity and gradually are becoming a standard tool for ultrasonic non-destructive evaluation research \cite{blackshire2002near, pohl2013laser}. In practice, wavefield imaging methods use a scanning laser vibrometry system to detect motions on a material surface generated by a stationary ultrasound excitation source. Repetitions of the ultrasonic excitation for different $(x,y)$ vibrometry detection points enables the collection of time series data (displacement amplitude vs time signals) at discrete $(x,y)$ spatial locations, resulting in a set of measurement data captured at regularized grid points. Additionally, this process also results in the collection of time-evolving snapshots of ultrasound wavefield data showing important details of the ultrasound-crack interaction i.e., crack scattering and back-scattering events. Researchers have successfully used wavefield imaging methods to quantitatively study ultrasonic wave dispersion (variation of phase velocity with frequency), attenuation (due to material property heterogeneity), damage scattering events, and time history details of elastic waves propagating in complex material systems \cite{dawson2016isolation, blackshire2017enhanced}. Wavefield imaging methods have, for example, been successfully used to study crack morphology details, e.g. direct visualization of ultrasonic wave interactions with realistic damage features resulting in an improved understanding of crack morphology and crack closure effects on the transmitted and reflected ultrasound wave characteristics \cite{blackshire2012ultrasonic}. In a related effort, frequency-wavenumber analysis was used to discriminate between bulk and surface wave modes scattered from a surface-breaking crack \cite{flynn2014embedded, tian2014lamb}. When these results are combined with the detailed information regarding the damage features e.g. size, shape and morphology of a crack, the scattering process of ultrasonic waves can be correlated with the locally developed damage features.  

Thus far, studies conducted to determine the detailed characteristics of a crack and its morphology using ultrasonic surface acoustic waves have been primarily based on direct visualization of the wavefields or an analysis of its time series. In the present study, we have used the wavefield data to completely characterize the incipient damage in an aerospace material using a PINN. The detection of a crack is based on the premise that the presence of a crack will result in attenuation of the wave energy, thus reducing the local sound speed. In this study, the sound speed, a function of space $v(x, y)$, is learned from the PINN.  We performed computational experiments on three sets of wavefield data acquired at different incidence angles (angle between incident wavefield and axis of the crack, measured from normal) of $0^o,~45^o,$ and $90^o$. The speed of sound recovered from these data sets represents the location and extent of the fracture quite accurately.     

\section{Ultrasonic surface acoustic wave data}
Wavefield imaging methods involve a quantitative measurement of out-of-plane particle displacements on the surface of a material caused by elastic wave motions. A typical wavefield imaging system utilizes a contact transducer to introduce transient ultrasonic waves on a material substrate. These motions are detected by a focused laser vibrometry beam at various spatial positions $(x,y)$ along the material's surface, where optical interferometric principles are used to relate phase and amplitude changes in the detected laser light to surface displacement motions \cite{blackshire2017enhanced}. A detailed study discussing the wavefield imaging method for acquiring ultrasonic surface acoustic wave data is provided by Blackshire \cite{blackshire2017enhanced}. The data used in this study was acquired at three incidence angles of $0^o$, $45^o$ and $90^o$, and it 
includes 2D time snapshots of the surface acoustic waves representing the particle displacements at $(x,y)$ regularized spatial grid locations for out-of-plane motions in the $z-$direction.
\begin{figure}
\centering
\subfloat[$t=11.58~\mu s$, $\text{at}=0^o$ without the crack.]{
\begin{tikzpicture}
\begin{axis}[enlargelimits=false, axis on top, axis equal image, xlabel=x (mm), ylabel=y (mm), width=0.45\textwidth]

\addplot graphics [xmin=0,xmax=12,ymin=0,ymax=12] {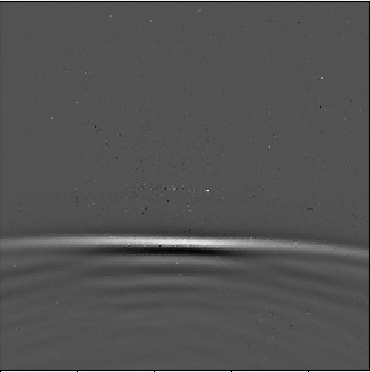};
\end{axis}
\end{tikzpicture}
}
\subfloat[$t=13.58~\mu s$, $\text{at}=0^o$ with the crack.]{
\begin{tikzpicture}
\begin{axis}[enlargelimits=false, axis on top, axis equal image, xlabel=x (mm), ylabel=y (mm), width=0.45\textwidth]
\addplot graphics [xmin=0,xmax=12,ymin=0,ymax=12] {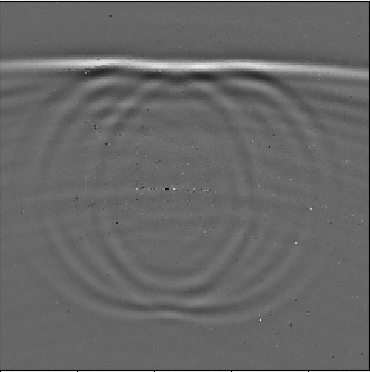};
\end{axis}
\end{tikzpicture}
}\\
\subfloat[$t=9.98~\mu s$, $\text{at}=45^o$ with out the crack.]{
\begin{tikzpicture}
\begin{axis}[enlargelimits=false, axis on top, axis equal image, xlabel=x (mm), ylabel=y (mm), width=0.45\textwidth]

\addplot graphics [xmin=0,xmax=12,ymin=0,ymax=12] {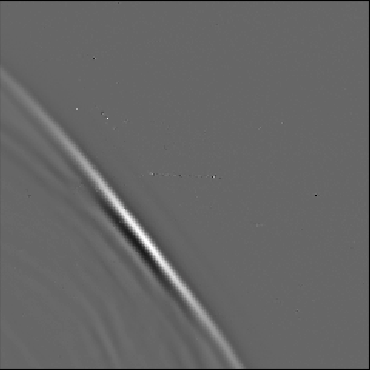};
\end{axis}
\end{tikzpicture}
}
\subfloat[$t=11.98~\mu s$, $\text{at}45^o$ with the crack.]{
\begin{tikzpicture}
\begin{axis}[enlargelimits=false, axis on top, axis equal image, xlabel=x (mm), ylabel=y (mm), width=0.45\textwidth]

\addplot graphics [xmin=0,xmax=12,ymin=0,ymax=12] {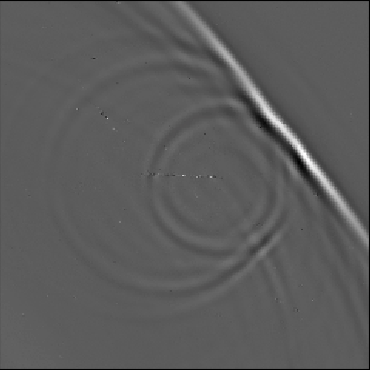};
\end{axis}
\end{tikzpicture}
}\\
\subfloat[$t=8.78~\mu s$, $\text{at}=90^o$ without the crack.]{
\begin{tikzpicture}
\begin{axis}[enlargelimits=false, axis on top, axis equal image, xlabel=x (mm), ylabel=y (mm), width=0.45\textwidth]

\addplot graphics [xmin=0,xmax=12,ymin=0,ymax=12] {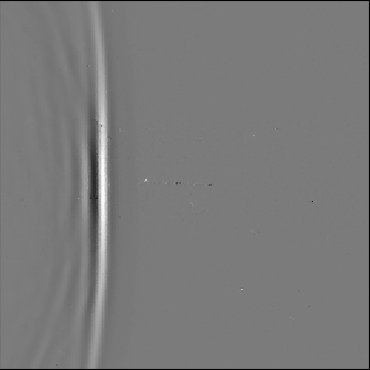};
\end{axis}
\end{tikzpicture}
}
\subfloat[$t=10.98~\mu s$, $\text{at}=90^o$ without the crack.]{
\begin{tikzpicture}
\begin{axis}[enlargelimits=false, axis on top, axis equal image, xlabel=x (mm), ylabel=y (mm), width=0.45\textwidth]

\addplot graphics [xmin=0,xmax=12,ymin=0,ymax=12] {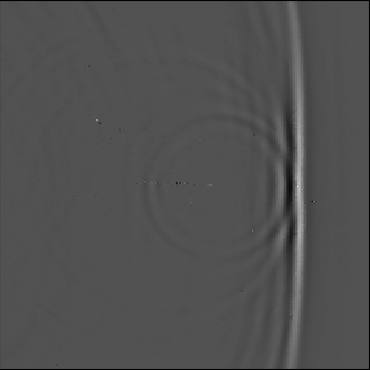};
\end{axis}
\end{tikzpicture}
}

\caption{Representative snapshots of particle displacement in metal plate, where (a) and (b) corresponds to $0^o$ incidence angle and showing the displacement with and without the effects of crack on wavefield, respectively. Similar representations for incidence angles of $45^o$ and $90^o$ are shown by pairs of [(c), (d)], and [(e), (f)], respectively. The data are from \cite{blackshire2017enhanced}.}
\end{figure}
The data was acquired using a National Institute of Standards and Technology (NIST) surface-breaking crack reference standard (RM 8458) in a $7~\text{cm} \times 7~\text{cm} \times 2~\text{cm}$ block of 7075-T651 aluminum alloy substrate material, with wavefield imaging measurements taken over a $12~\text{mm} \times 12~ \text{mm}$ region centered over the crack with a scanning step resolution of $50~\text{microns} (240 \times 240~x-y~\text{scan points})$, and a time-step data acquisition interval of $0.02~\mu\text{s}$ ($1024$ total time-step points or $20.48~\mu\text{s}$). A wave packet of 5 MHz was propagated as initial conditions and data were arranged in a dimension $[\text{Nx}, \text{Ny}, \text{Nt}]$ of $240 \times 240 \times 1024$, with Nx, Ny being the grid points in $x$ and $y$ directions and Nt is number of time samples. These datasets were acquired for three sets of incidence angles $0^0$, $45^o$ and $90^o$. The angle of incidence is the angle between the wave vector and the axis of the fracture and measured from the normal. Examples of snapshots of wavefield representing the particle displacement are shown in Figure 1. Figures 1a and 1b represent the snapshots of wavefield at 11.58 $\mu s$ and 13.58 $\mu s$, before and after the interaction of the wavefield with the crack  at $0^o$ of incidence angle, respectively. Similarly, Figures 1c and 1d represent the snapshots of the wavefield at $45^o$ (at 9.98 $\mu s$) with and without the effects of crack on the wavefield, respectively. Figures 1e and 1f represent the snapshots for incidence angle of $90^o$. It is quite evident that the back-scattering phenomena are more prominent for angles $0^o$ and $45^o$ than those of recorded at angle of $90^o$.

\section{Problem setup and Physics Informed Neural Network (PINN)}
To characterize the crack in terms of location and extent, we postulate that the speed, $v(x,y)$,  of the wave will be the key feature. As ultrasonic acoustic waves propagate in a medium with crack(s), the speed of wave decreases due to inelastic attenuation and scattering of the wave energy \cite{carcione2007wave}. Thus, a difference wave speed in and around the crack will show evidence of the presence of crack. In this study, we consider the linear second-order partial differential equation governing the propagation of acoustic wave equations, which is expressed as 

\begin{align} \label{eq1}
    u_{tt} = v^{2}(x, y) \Delta u, \qquad x \in \Omega, ~\text{and}~t \in [0, T],
\end{align}
where $u(t, x, y)$ is the solution, $v(x,y)$ is the sound speed and $\Omega \in \mathbb{R}^2$. The subscript $t$ denotes the partial differentiation in time domain. Here, we are given measurements $u$ and want to learn $v(x, y)$ that best describes equation (\ref{eq1}). These problems of learning are known as system identification or data driven discovery of partial differential of equations \cite{raissi2017machine, rudy2017data}. Neural networks (NNs) are successfully used to obtain the approximate solution of partial differential equations (PDEs). One can also construct the physics-informed machine learning using systematically structured prior information about the solutions. The work in  \cite{raissi2018hidden,raissi2017machine,raissi2019deep} have successfully demonstrated the use of PDEs as prior information to constrain the minimization process in the context of system identification (forward problem) and solution inferences (inverse problem). In particular, PINN can solve forward and inverse problem accurately. In the forward problems, the approximate solutions of the PDEs are obtained, whereas the inverse problem computes the parameters or even unknown functions involved in the PDEs. The problem of characterization of a crack falls into the  second category i.e. computation of $v(x,y)$ with $u(t,x,y)$ given by the real data. The loss function constructed in the PINN algorithm incorporates the residual term from the governing equation(s), which act as a regularization term and thus constrains the space of admissible solutions.  

\subsection{Neural network}
We consider a neural netowrk (NN) of depth $D$ corresponding to a network with an input layer, $D-1$ hidden layers and an output
layer. In the $k^{\text{th}}$ hidden layer, $N_k$ number of neurons are present. Each hidden layer of the network receives an output $x^{k-1} \in \mathbb{R}^{N_{k-1}}$ from the previous layer where an affine transformation is performed. The transformation is expressed as
\begin{align} \label{eq2}
\mathcal{L}_k(x^{k-1}):=w^kx^{k-1} + b^k.    
\end{align}
The network weights $w^k$ and bias term $b^k$ for $k^{\text{th}}$ are initially chosen from \textit{independent and identically distributed (iid)} samplings. The nonlinear activation function $\sigma(\cdot)$ is applied to  $\mathcal{L}_k(x^{k-1})$ component wise prior to sending it as input to next layer. The activation function at the output layer is an identity function. Thus, the final neural network representation is expressed as

\begin{align}\label{eq3}
u_\Theta(x)=\left(\mathcal{L}_k \circ \sigma \circ \mathcal{L}_{k-1} \circ ...\circ \sigma \circ \mathcal{L}_1\right)(x),    
\end{align}
 where the operator $\circ$ is the composition operator, $\Theta=\{w^k, b^k \}_{k=1}^D$ represents the trainable parameters in the network, $u$ is the output and $x:=x^0$ is the input to the neural network.
 
 \subsection{PINN for ultrasonic surface acoustic wave equation}
We define the residual $f(t, x)$ by (\ref{eq1}) i.e.,
\begin{align} \label{eq4}
f:= u_{tt} - v^{2}(x, y) \Delta u,
\end{align}
and proceed with approximating  $u(t,x)$ with a deep neural network as expressed in equation (\ref{eq3}). This assumption along with equation (\ref{eq4}) results in the \textit{physics informed neural network $f(t,x)$}. This network is derived by applying the chain rule for differentiating composition of functions using automatic differentiation \cite{raissi2017machine,baydin2018automatic} and has the same parameters representing the $u(t,x)$ but with different activation function due to the $\Delta$ operator in (\ref{eq1}). The shared parameters between the neural networks $u(t,x)$ and $f(t,x)$ can be learned by minimizing the mean squared loss error expressed as 
\begin{align} \label{eq5}
    MSE = \lambda MSE_u + MSE_f,
\end{align}
where $MSE_u=\myfrac{1}{N_u}\sum_{i=1}^{N_u} \left \vert u(t_u^i, x_u^i, y_u^i) - u^i \right \vert^2$, and $MSE_f = \myfrac{1}{N_f}\sum_{i=1}^{N_f} \left \vert f(t_f^i, x_f^i, y_f^i) \right \vert^2$. The parameter $\lambda > 0$ is a penalty parameter, which helps in achieving the fast convergence.  
Here, $\{t_u^i, x_u^i, u^i\}$ denote the initial and boundary training data on $u(t,x)$ and $\left\{t_f^i, x_f^i \right\}$ specify the residual points for $f(t,x)$. The loss $MSE_u$ here corresponds to the data and $MSE_f$ enforces the structure imposed by (\ref{eq4}). 

In the present study, the characterization of the crack is provided by spatially varying speeds of the ultrasonic acoustic waves i.e $v(x,y)$. Thus, to estimate the speed, we used another neural network to learn the $v(x,y)$ and subsequently fed to PINN. A block diagram representing the PINN model used in this study is shown in Figure 2a.

\subsection{Loss function and optimization algorithm}
Our aim is to find the optimal weight for which the loss function defined in (\ref{eq5}) is minimized. Thus, the definition of the resulting optimization problem is expressed as
\begin{align} \label{eq6}
w^* = \arg \min_{w \in \Theta } (J(w)); ~~~b^* =\arg \min_{b \in \Theta }(J(b))~~~v^* =\arg \min_{v \in \Theta }(J(b)).
\end{align}

One can approximate the solutions to optimization problems, defined in (\ref{eq6}), using an iterative method by using one of the forms of gradient descent algorithm. The stochastic gradient descent (SGD) algorithm is ubiquitously used by machine learning community \cite{rudy2017data}. In SGD method, the weights are updated as
\begin{align} \label{eq7}
    w^{m+1} = w^m -\eta_l \nabla_w J^m(w),
\end{align}
where $\eta_l > 0$ is the learning rate and $J^m$ is the loss function at $m^{th}$ iteration. SGD methods could be initialized with some starting value $w^0$. In this work, the ADAM optimizer \cite{kingma2014adam}, a variant of SGD method, is used unless mentioned otherwise. 

\begin{figure}
\centering
\subfloat[PINN model]{
\includegraphics[trim=0cm 0cm 0cm 0cm, clip, width=0.6\textwidth]{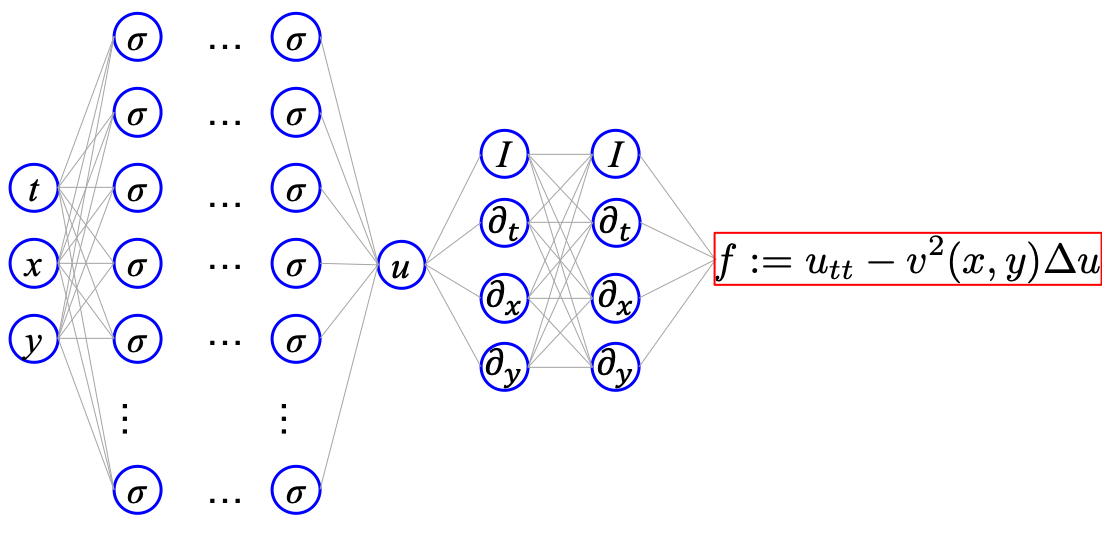}
}
\subfloat[Activation function]{
\includegraphics[trim=0cm 0cm 0cm 0cm, clip, width=0.40\textwidth]{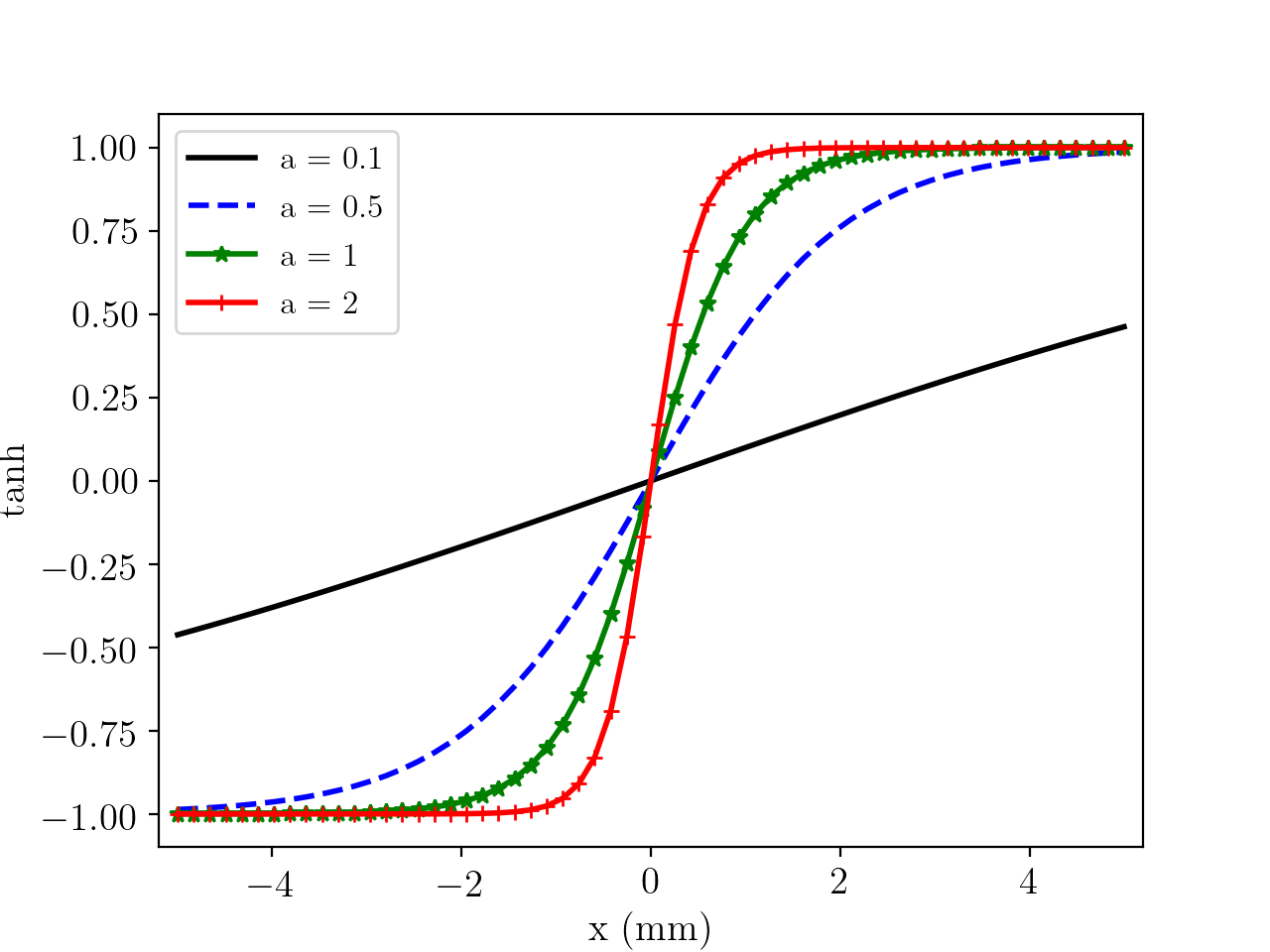}
}

\caption{(a) represents physics informed neural network model informed by the acoustic wave equation and (b) shows  comparison of activation function $\tanh$ at various values of $a$.}

\end{figure}

\subsection{Adaptive activation function}
The dependence of the derivative of the loss function on the optimization parameters defines a very important role of activation function. The regularity of the derivative of the loss function is dependent on the activation function. Various activation functions are used in PINN algorithm such as $\tanh, \sin etc$ to solve the PDEs \cite{raissi2019physics, raissi2019deep}. There is no criterion to choose these function as it solely depends on the problem in hand. In the present work, while performing the training on ultrasonic acoustic data with the $\tanh$, a very slow and non-convergence of loss function was observed. To speed up the convergence we adopted the method of the adaptive activation function \cite{jagtap2019adaptive}. In the adaptive activation method, Jagtap et al. \cite{jagtap2019adaptive} introduced the hyper-parameter $a > 0$ in the activation function as $\sigma \left( a \mathcal{L}_k \left( x^{k-1}\right) \right)$, where $a$ is subsequently learned by minimizing the loss function along with the weights and biases. Thus, the definition of optimization problem for $a$ is expressed as
\begin{align*}
    a^* = \arg \hspace*{-0.3cm}\min_{a \in \mathbb{R}^+ \backslash \{0\} } (J(a)).
\end{align*}
The parameter $a$ is updated as
\[
a^{m+1} = a^m - \eta_l \nabla_a J^m(a).
\]

To accommodate the effect of learning factor \cite{goyal2017accurate}, responsible for convergence to global minima, the hyper-parameter $a$ is multiplied by a scale factor $n \geqslant 1$ and the final activation function is recovered as 
\[
\sigma\left(na \mathcal{L}_k \left( x^{k-1}\right) \right).
\]

The effect of $a$ on the activation function defined by hyperbolic tangent $(\tanh)$ is shown in Figure 2b. It is to be noted from Figure 2b that introducing $a$ steepens the activation function, which eventually helps in achieving the fast convergence. The effect parameter $a$ on achieving the fast convergence is shown in detail by Jagtap et al. \cite{jagtap2019adaptive} for problems dealing with non-linear PDEs such as the Burger's equation, Klein-Gordon equation, as well as other standard machine learning benchmarks. 

\begin{figure}
\centering
\subfloat[Actual data at $t=11.38~\mu s$.]{
\includegraphics[trim=0cm 0cm 0cm 0cm, clip, width=0.5\textwidth]{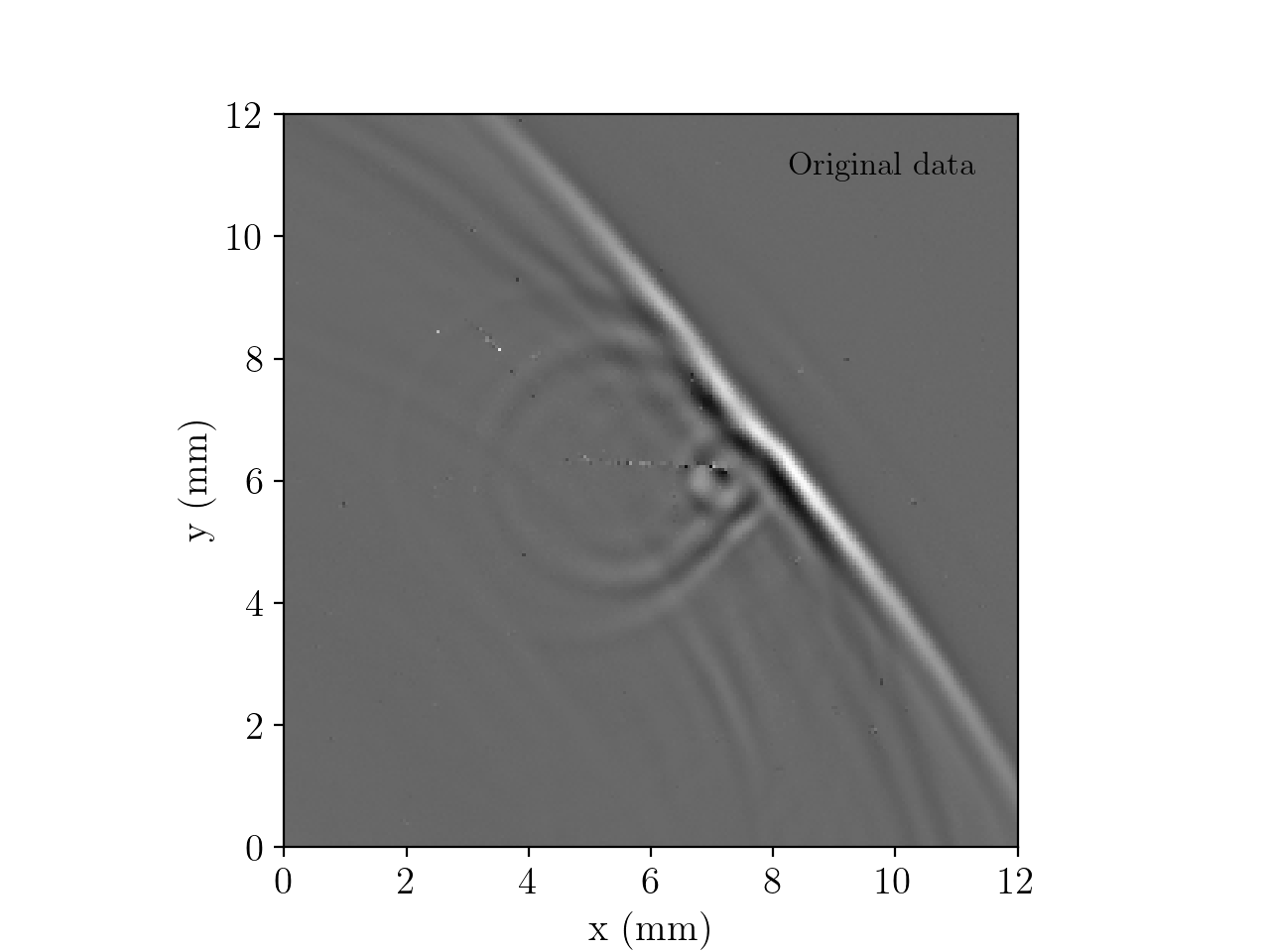}
}
\subfloat[Cumulative explained variance of (a).]{
\includegraphics[trim=0cm 0cm 0cm 0cm, clip, width=0.5\textwidth]{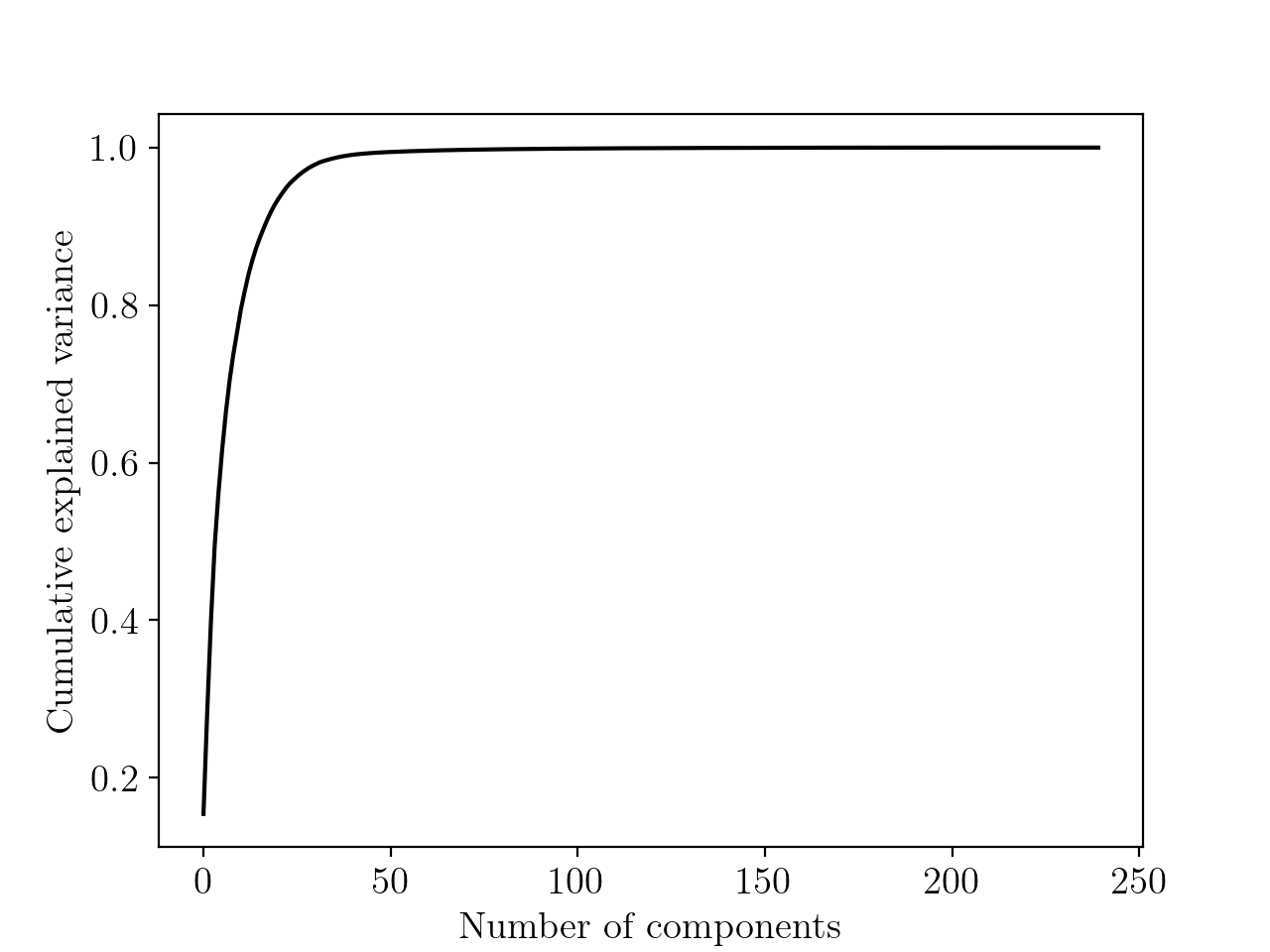}

}
\\
\subfloat[Filtered data using at $t=11.38~\mu s$.]{
\includegraphics[trim=0cm 0cm 0cm 0cm, clip, width=0.5\textwidth]{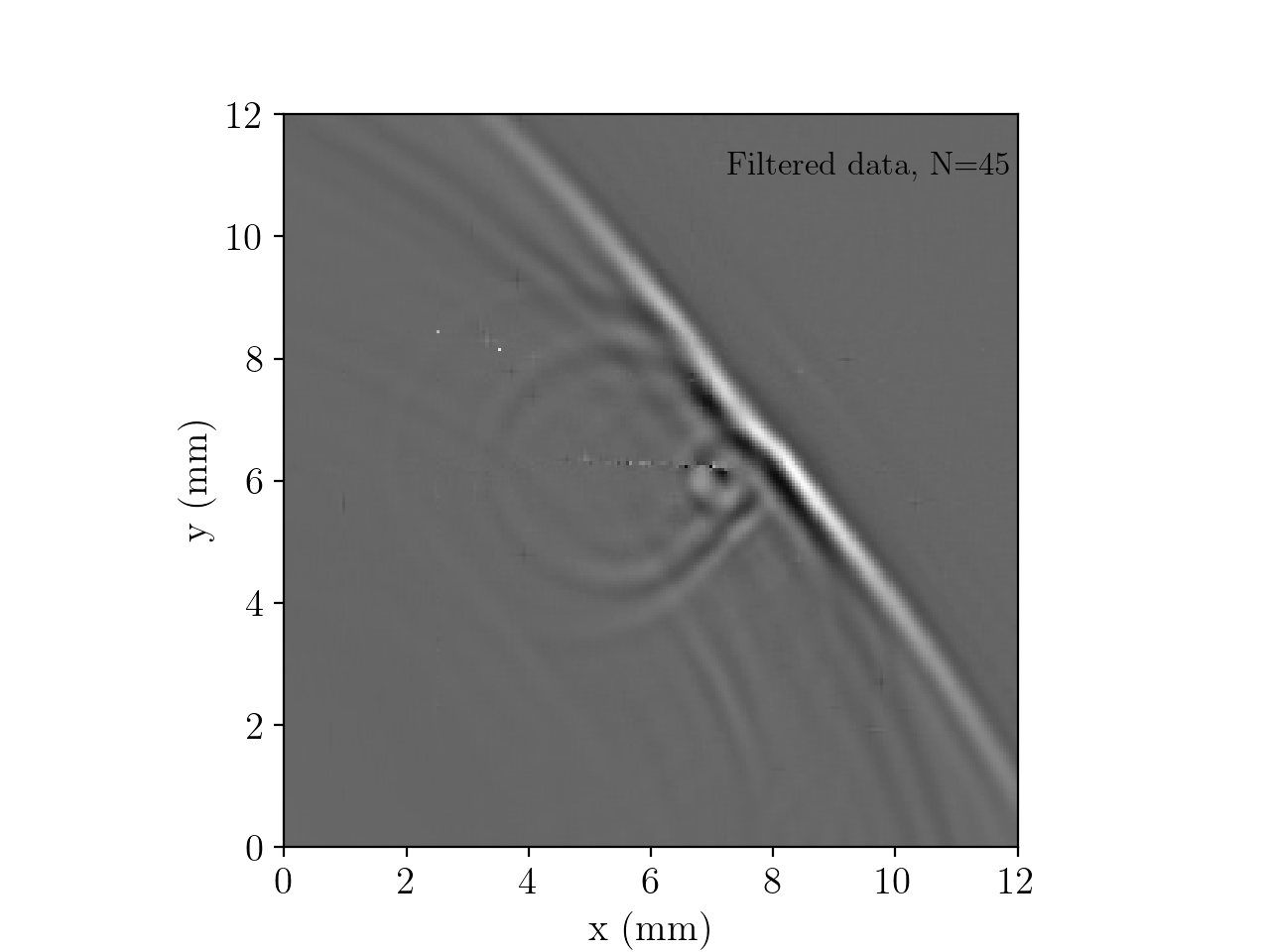}
}
\subfloat[Original vs filtered trace at $x=6.5~\text{mm}$ at $t=11.38~\mu s$.]{
\includegraphics[trim=0cm 0cm 0cm 0cm, clip, width=0.5\textwidth]{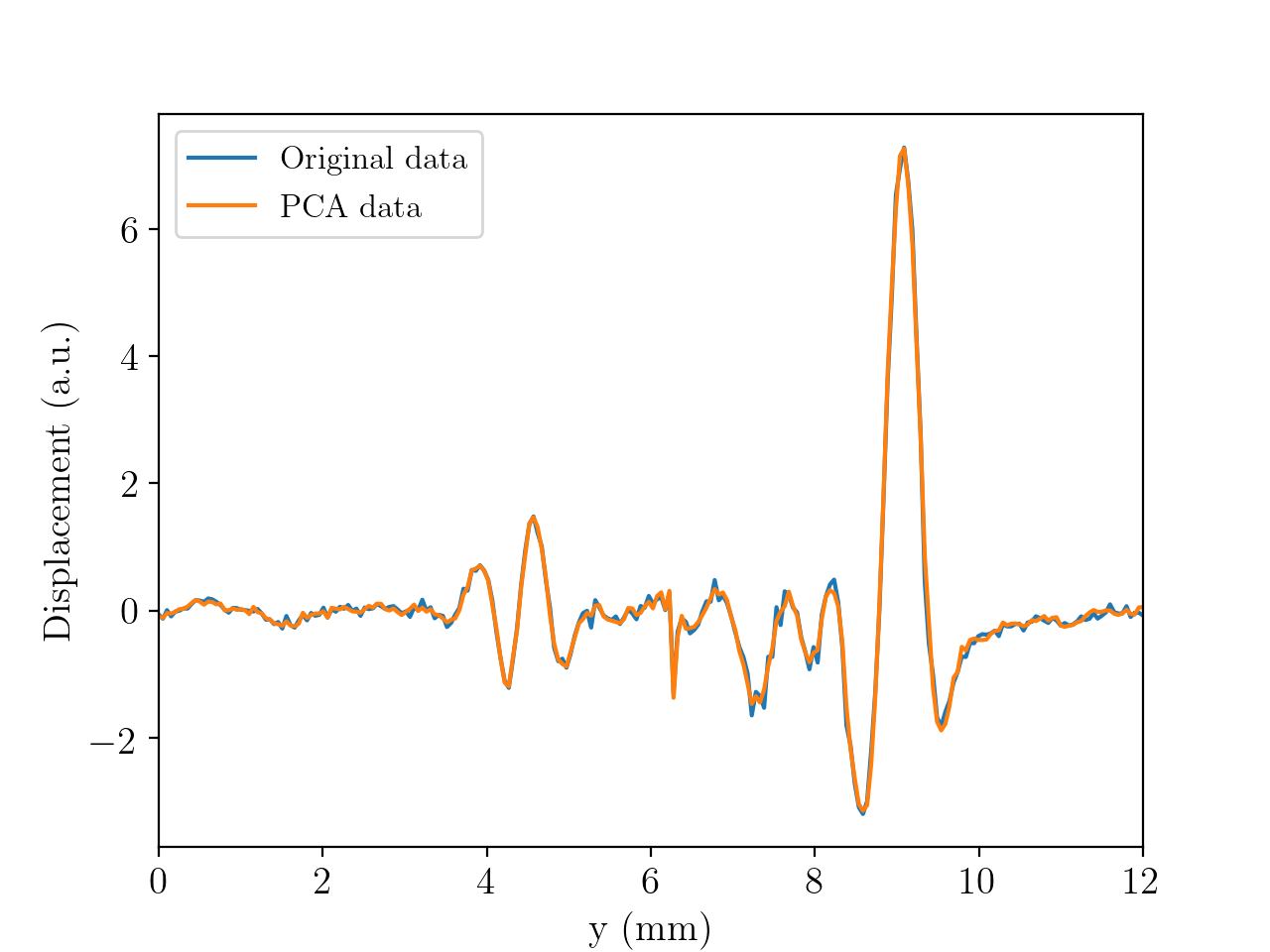}
}
\caption{Filtering of wavefield data acquired at $45^o$, using the principle component analysis (PCA). PCA requires only the first 45 components to construct the original data by zeroing out the insignificant components.}
\end{figure}

\begin{figure}
\centering
\subfloat[Actual data at $t=10.8~\mu s$.]{
\includegraphics[trim=0.5cm 2.4cm 0cm 0cm, clip, width=0.5\textwidth]{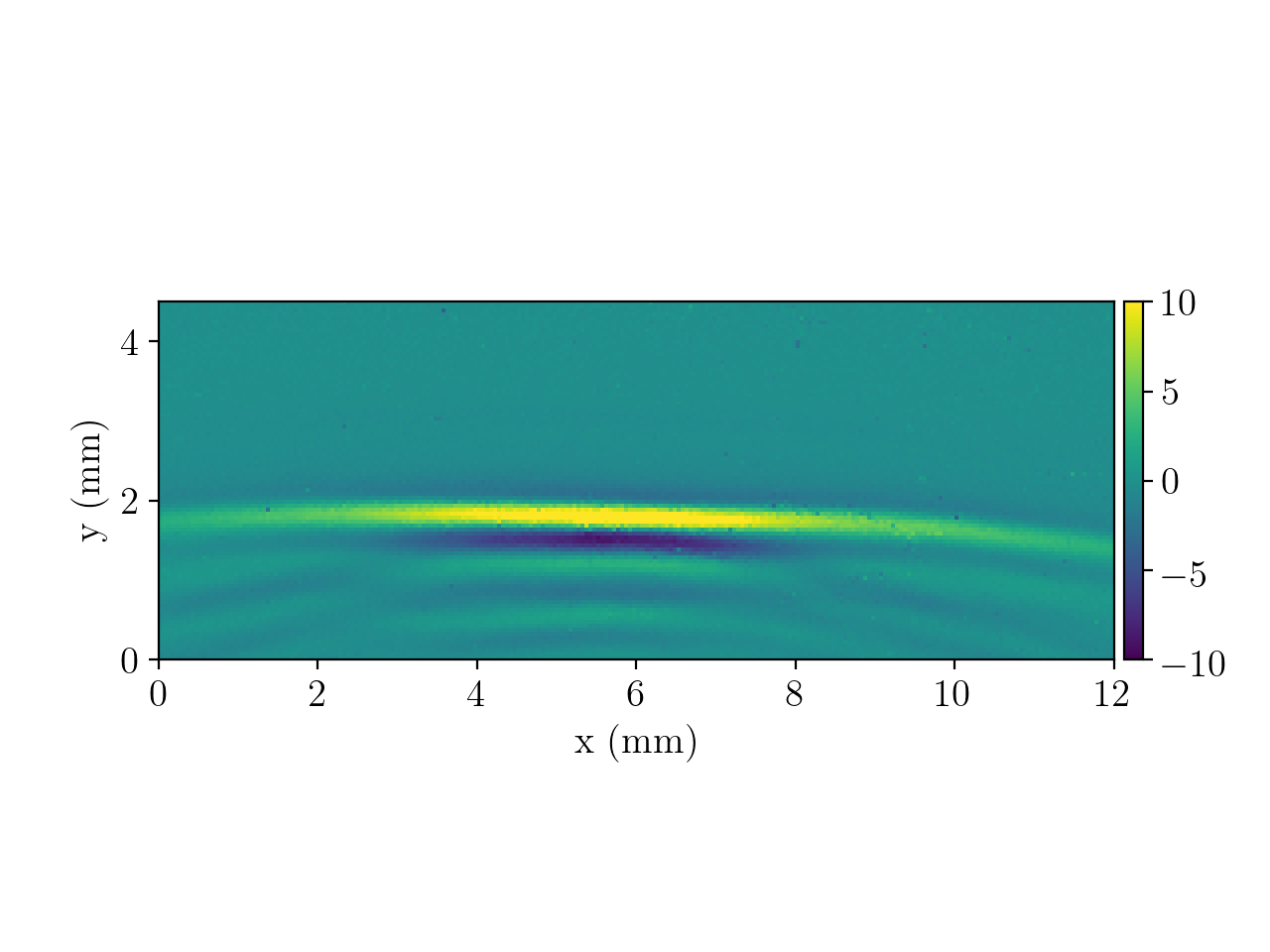}
}
\subfloat[Data recovered from PINN at $t=10.8~\mu s$.]{
\includegraphics[trim=0.5cm 2.4cm 0cm 0cm, clip, width=0.5\textwidth]{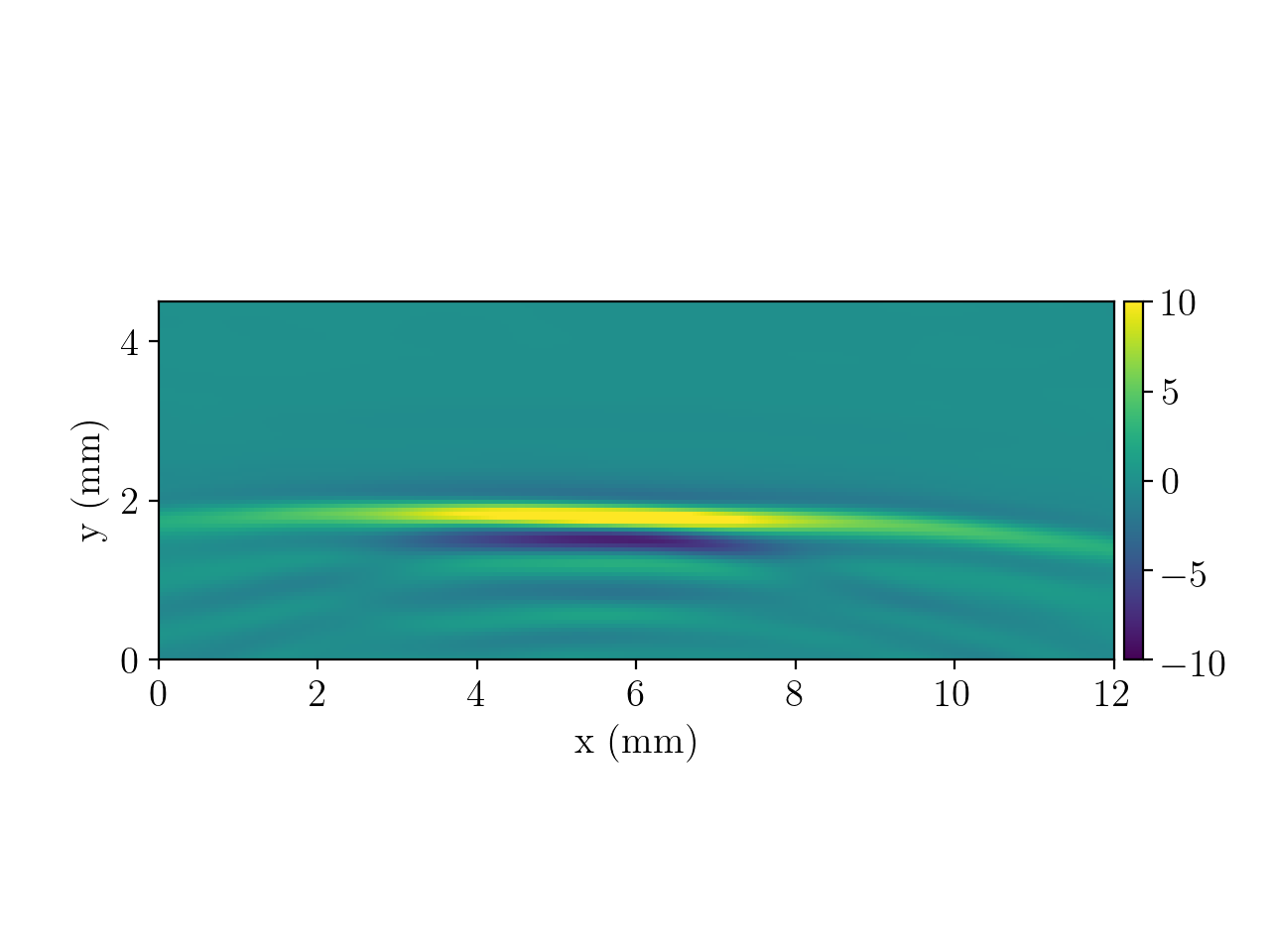}
}\\
\subfloat[Traces of (a) and (b) at $x=6~\text mm$ at $t=10.8~\mu s$.]{
\includegraphics[trim=0.5cm 0cm 0cm 0cm, clip, width=0.5\textwidth]{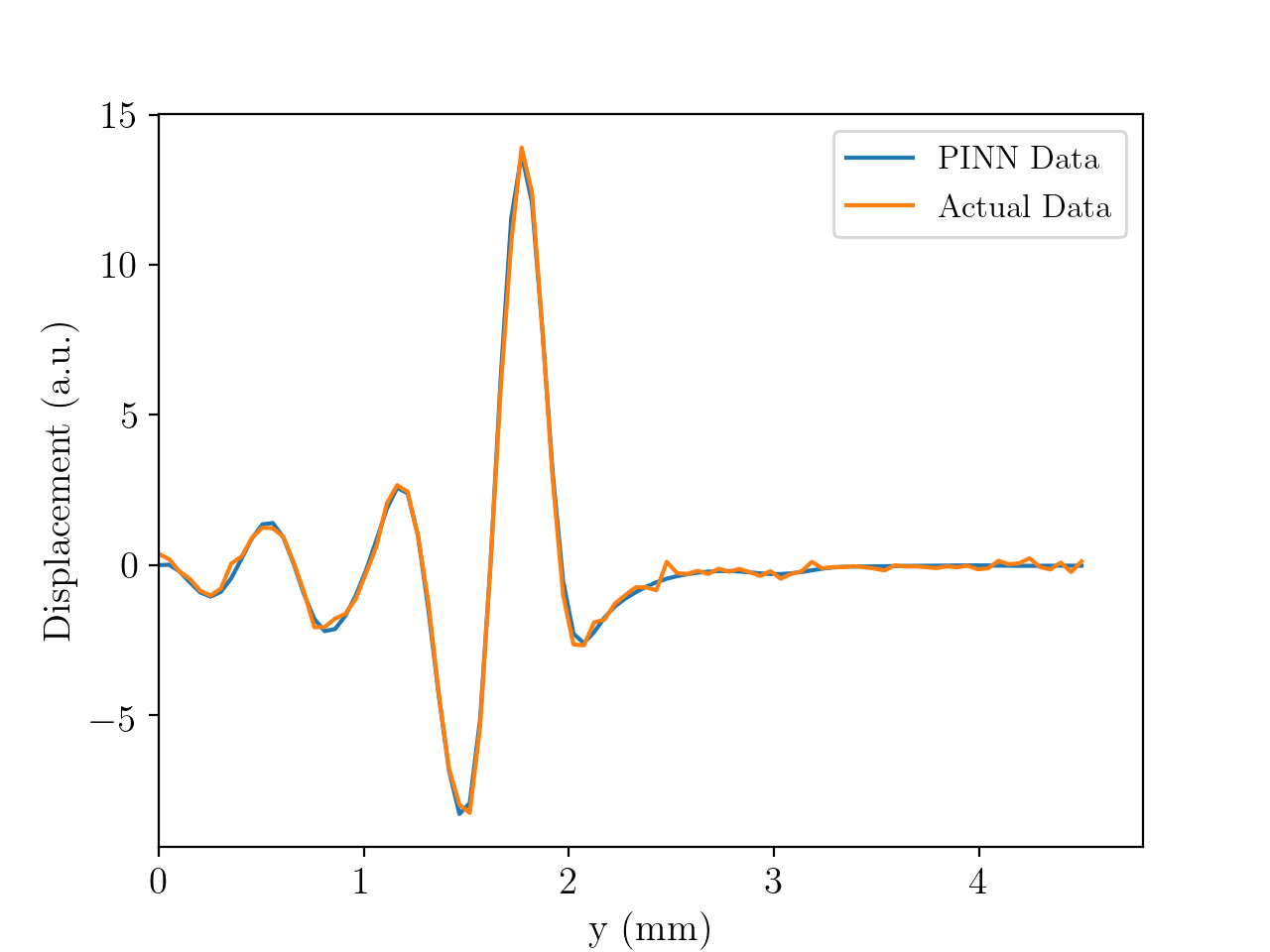}
}
\subfloat[Wave speed discovered from PINN simulation as a function of epoch.]{
\includegraphics[trim=0cm 0cm 0cm 0cm, clip, width=0.5\textwidth]{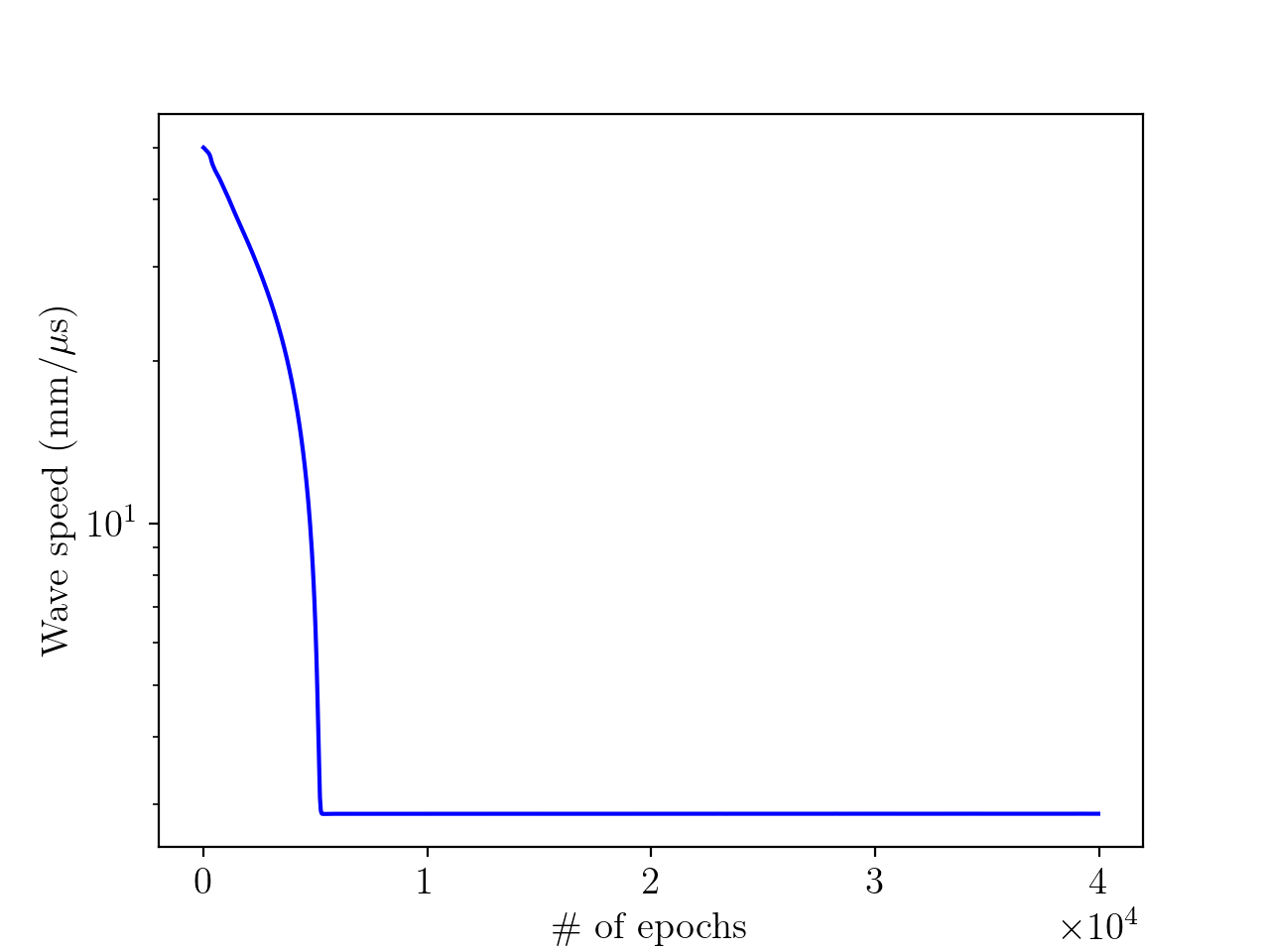}
}\\
\subfloat[Loss function of neural network model]{
\includegraphics[trim=0cm 0cm 0cm 0cm, clip, width=0.5\textwidth]{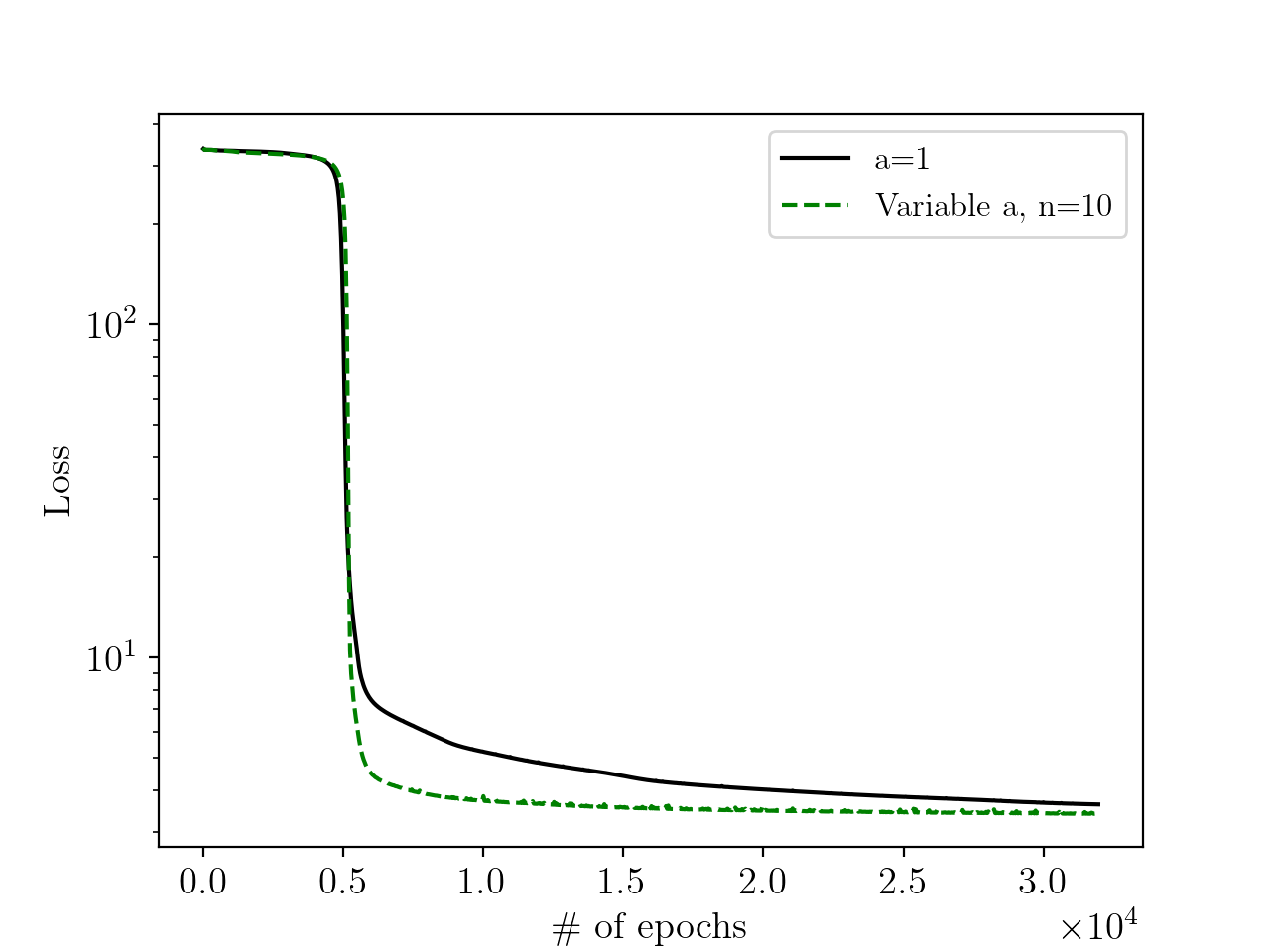}
}
\subfloat[Variation in a with $N=10$.]{
\includegraphics[trim=0cm 0cm 0cm 0cm, clip, width=0.5\textwidth]{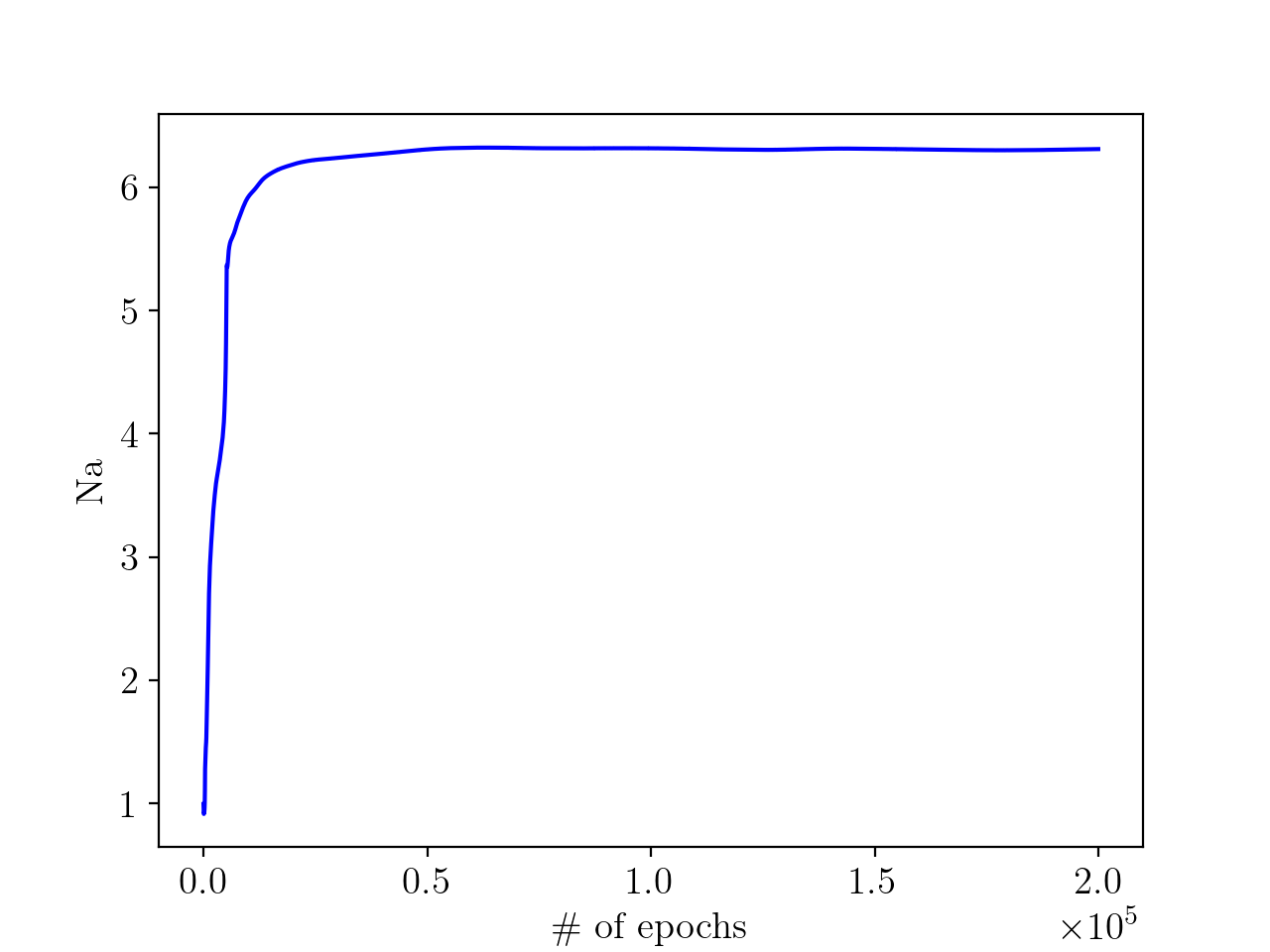}
}\\
\caption{Results from training of neural network model for data acquired at incidence angle of $0^o$ and devoid of crack, where (a) and (b) represents the snapshot of the particle displacement from obtained from actual and PINN simulated data, respectively. (c) shows a comparison of traces obtained from actual and PINN simulated data. (d) represents the speed of wave discovered from PINN simulation as a function of epoch. (e) shows a comparison of loss function for fix and adaptive activation function and (f) shows variation in $a$ with $N=10$. }
\end{figure}
\subsection{Data preconditioning}
Wavefield data are acquired through an experimental set up and prone to random and environmental noise. To filter out these random noises, we have used the method of principal component analysis (PCA). PCA is an unsupervised method for extracting the variance structure from a high dimensional data and project the data into a subspace
such that variance of projected data is maximized. Application of PCA to the data provides two types of information: principal components and explained variance ratio, which is the ratio between the variance of that principal component and the sum of variances of all individual principal components. Thus, the filtering process involves the zeroing out one or more smallest principle component by preserving the maximum data variance. We used the PCA module of sklearn package \cite{scikit-learn}, and a snippet of routine used to filter out the data is given in \ref{PCA1}. 

\begin{figure}
\centering
\subfloat[Actual data at $t=12.38~\mu s$.]{
\includegraphics[trim=0.5cm 2.4cm 0cm 0cm, clip, width=0.5\textwidth]{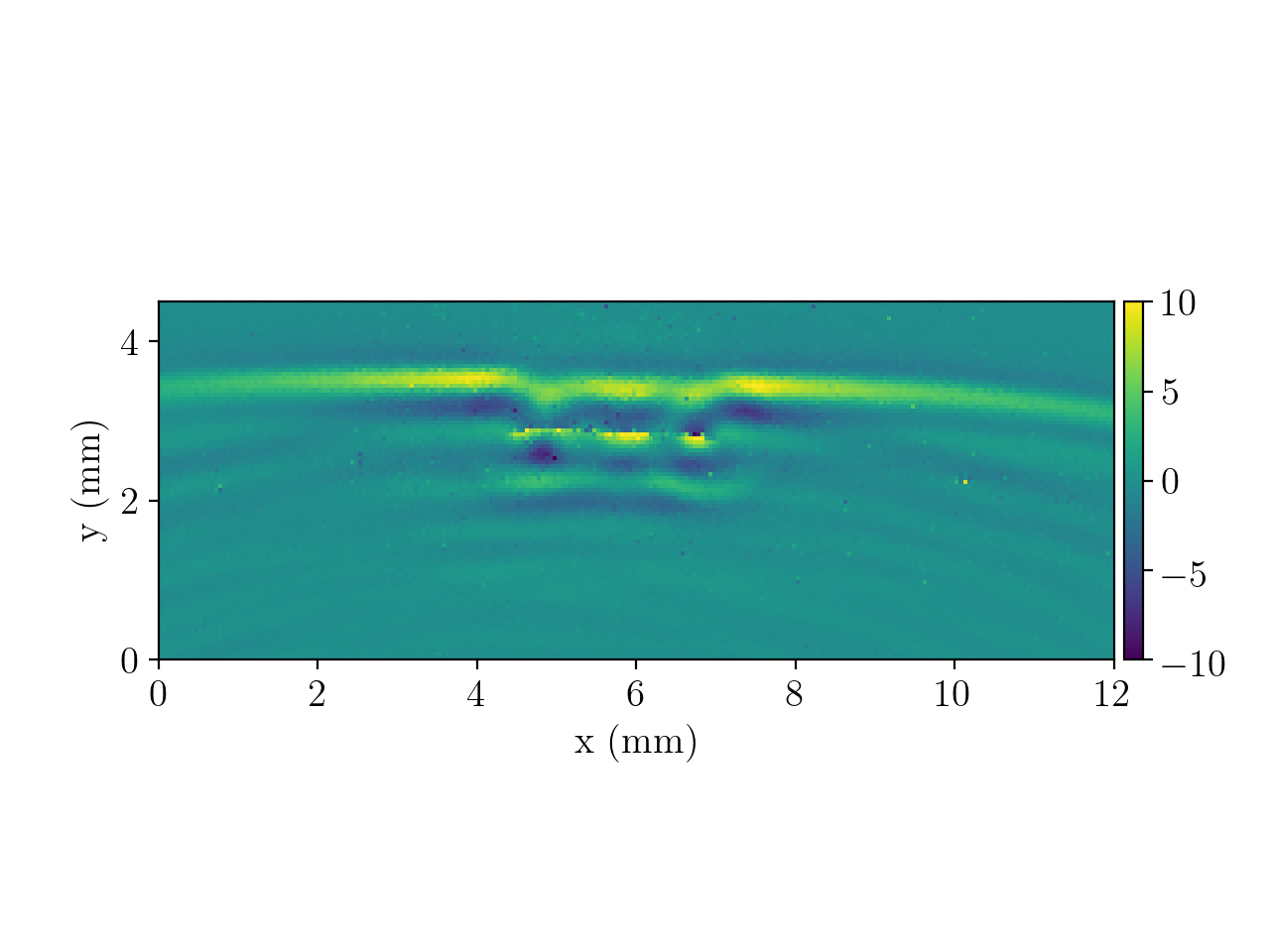}
}
\subfloat[Data recovered from PINN simulation at $t=12.38~\mu s$.]{
\includegraphics[trim=0.5cm 2.4cm 0cm 0cm, clip, width=0.5\textwidth]{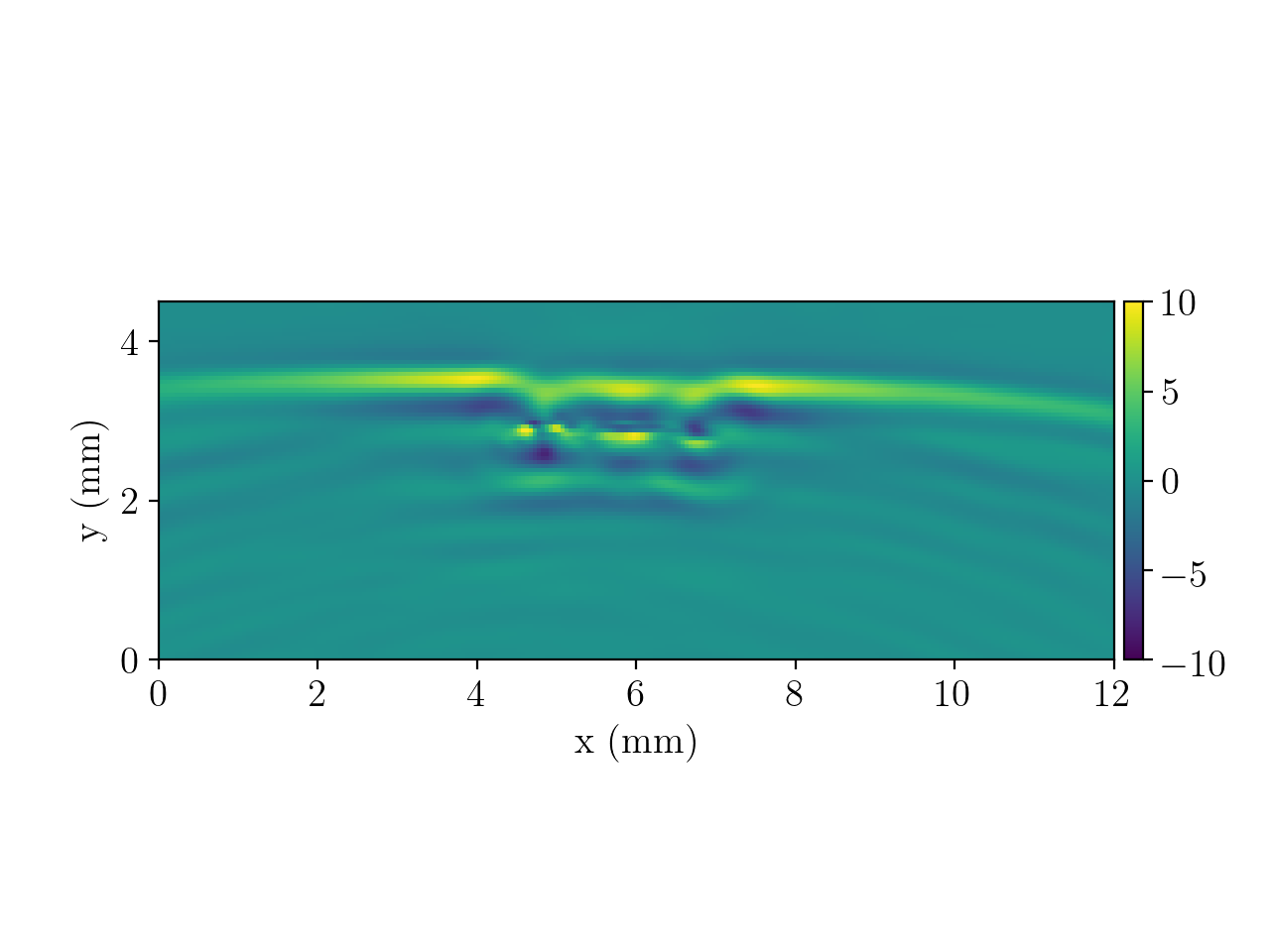}
}\\
\subfloat[Traces of (a) and (b) at $x=6~\text(mm)$ at $t=12.38~\mu s$.]{
\includegraphics[trim=0.5cm 0cm 0cm 0cm, clip, width=0.5\textwidth]{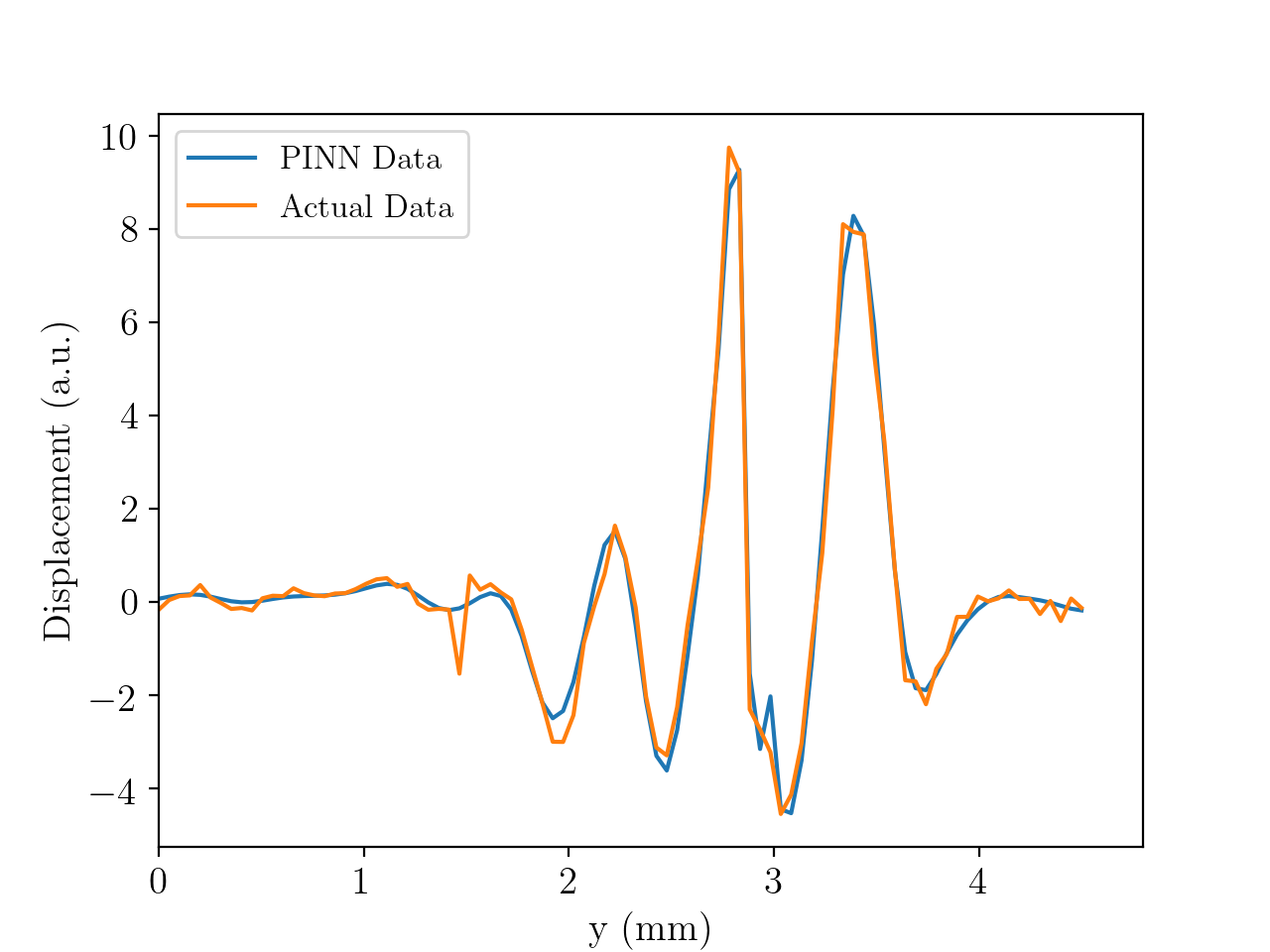}
}
\subfloat[Wave speed discovered from PINN simulation as a function of epoch.]{
\includegraphics[trim=0cm 0cm 0cm 0cm, clip, width=0.5\textwidth]{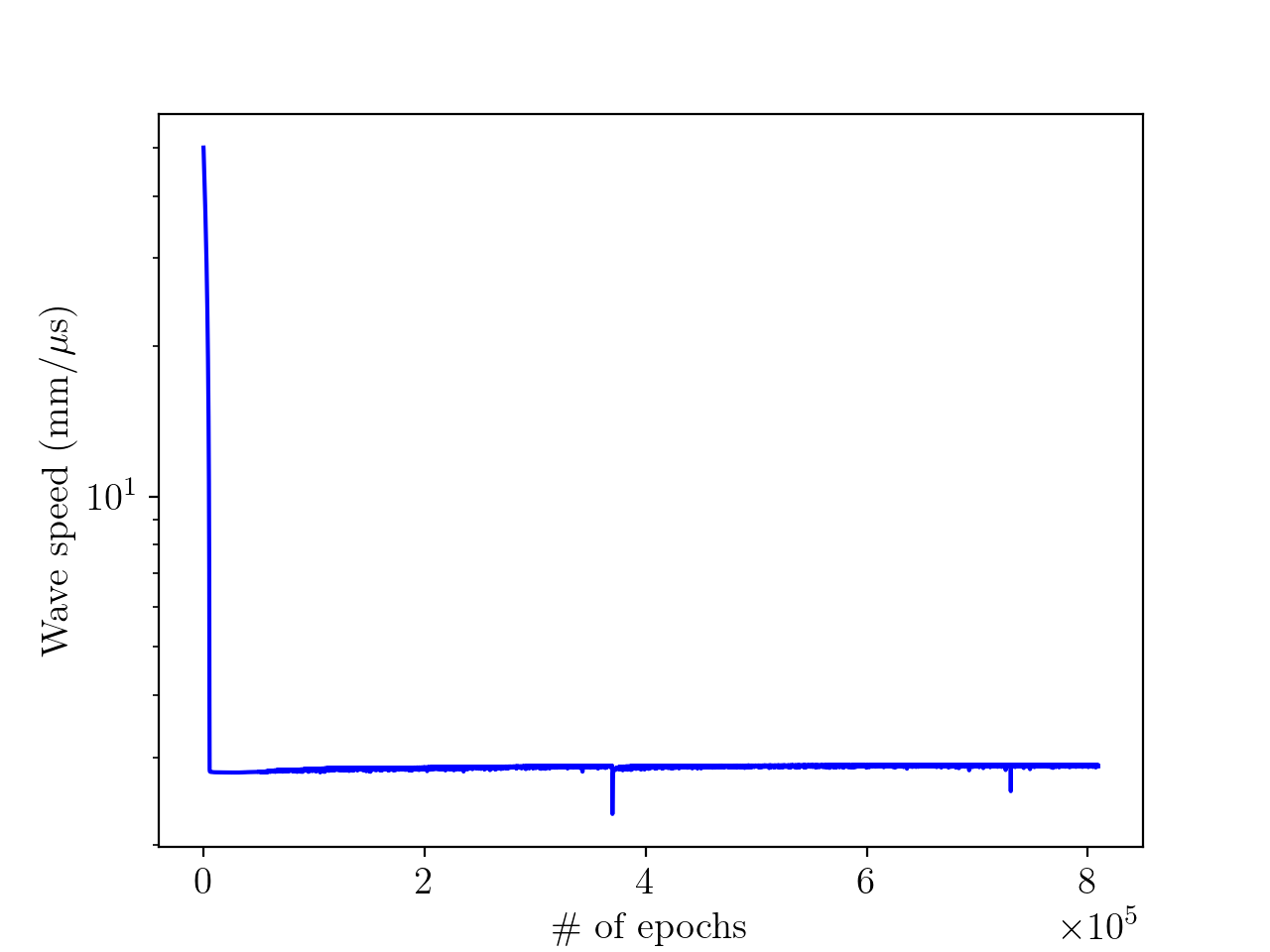}
}\\
\subfloat[Loss function of neural network model.]{
\includegraphics[trim=0cm 0cm 0cm 0cm, clip, width=0.5\textwidth]{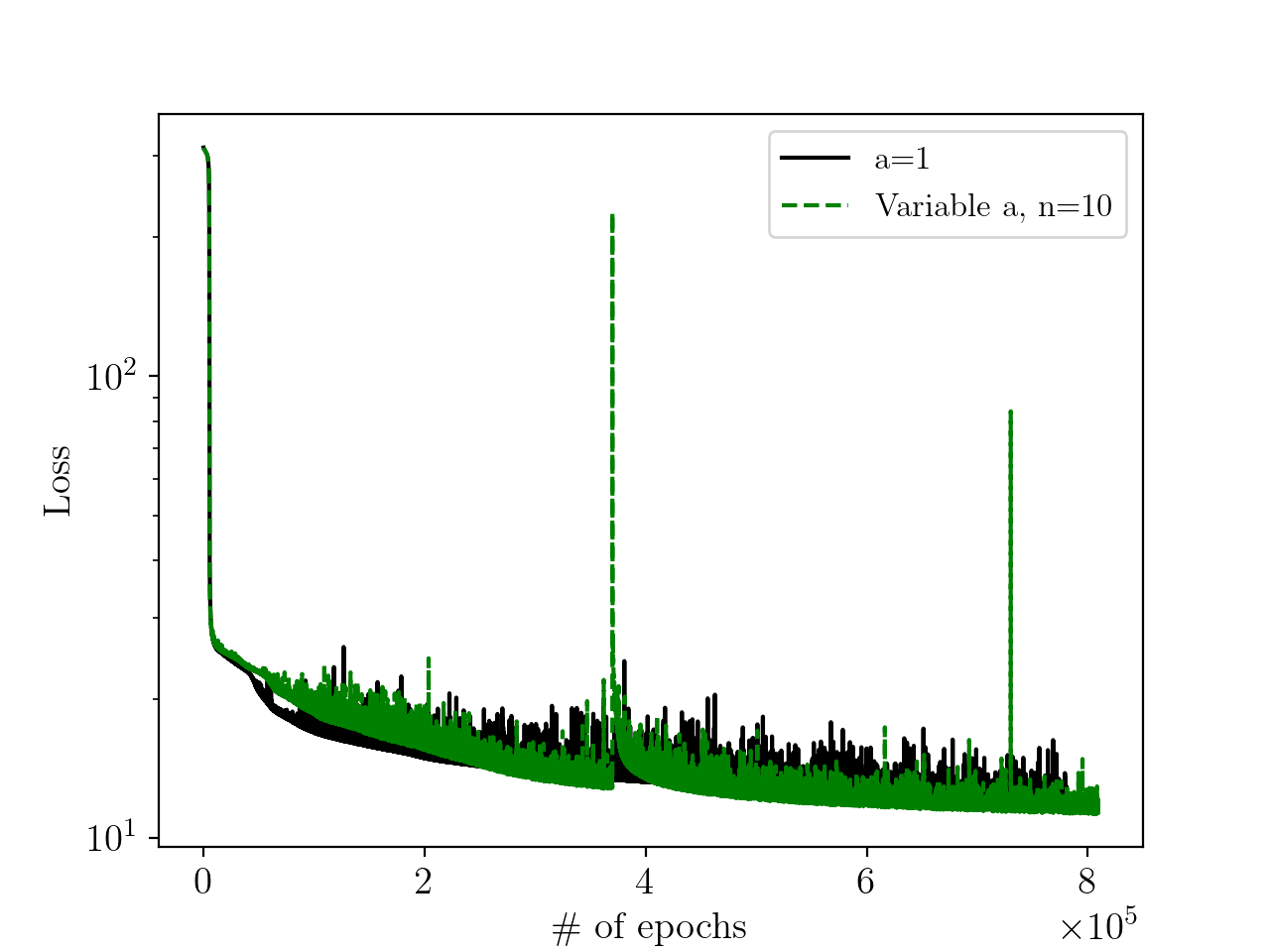}
}
\subfloat[Variation in a with $N=10$.]{
\includegraphics[trim=0cm 0cm 0cm 0cm, clip, width=0.5\textwidth]{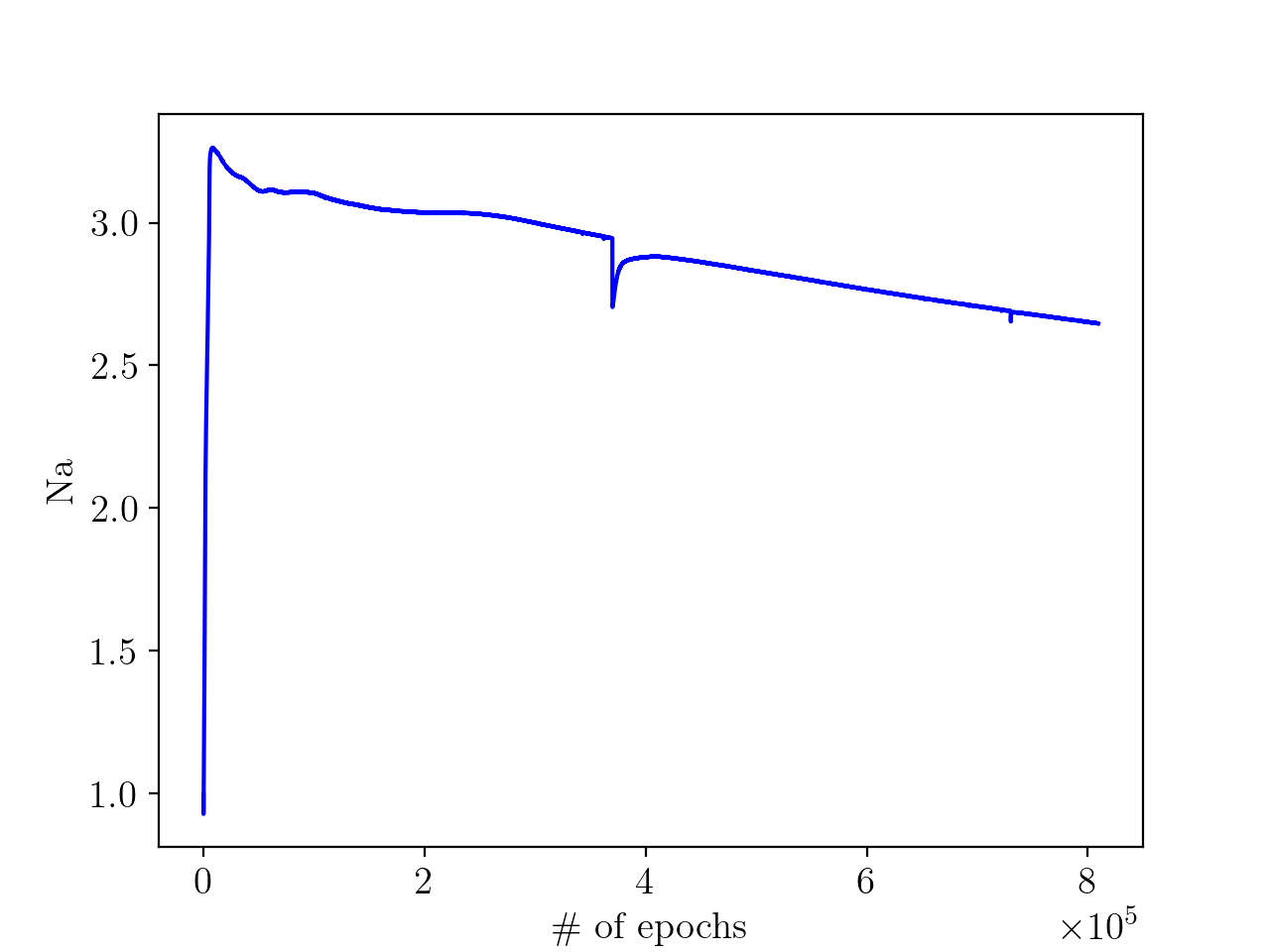}
}
\caption{Results from training of neural network model for data acquired at incidence angle of $0^o$ including the crack, where (a) and (b) represents the snapshots of the particle displacement from actual and PINN simulated data, respectively. (c) shows a comparison of traces obtained from actual and PINN simulated data and (d) represents the speed of wave discovered from PINN simulation as a function of epoch (e) shows a comparison of loss function for fix and adaptive activation function and (f) shows variation in $a$ with $N=10$.}
\end{figure}
Results obtained after application of PCA on data set for incidence angle $45^o$ are shown in Figure 3. Figure 3a represents a snapshot of the raw data at $11.38~\mu s$ for $45^o$ data set. Figure 3b shows a plot of cumulative explained variance and principle components. This quantifies how much of the total, 240-dimensional variance is contained within the first $N$ components. Figure 3b clearly shows that the first 45 components contain approximately 95\% of the variance. Figure 3c represent the filtered data obtained after discarding the components close to zero. Figure 3d represents a comparison between raw and filtered trace extracted from Figures 3a and 3c at $x=7~\text{mm}$, respectively. Figure 3d clearly shows that the filtered data is smoother in comparison to the raw data. Results for datasets acquired at incidence angle of $45^o$ and $90^o$ are shown in appendix B as Figure B1 and Figure B2, respectively. To construct the filtered data for $0^o$ case, it requires first $15$ principle components, less than those for $45^0~(N=45)$ and $90^0 (N=20)$ cases. This is due to the fact that back scattering effects are more prominent for $45^o$ than $0^o$ and $90^o$.

\section{Results}
\begin{figure}
\centering
\subfloat[PCA filtered data at $t=12.38~\mu s$.]{
\includegraphics[trim=0cm 2.4cm 0cm 0cm, clip, width=0.5\textwidth]{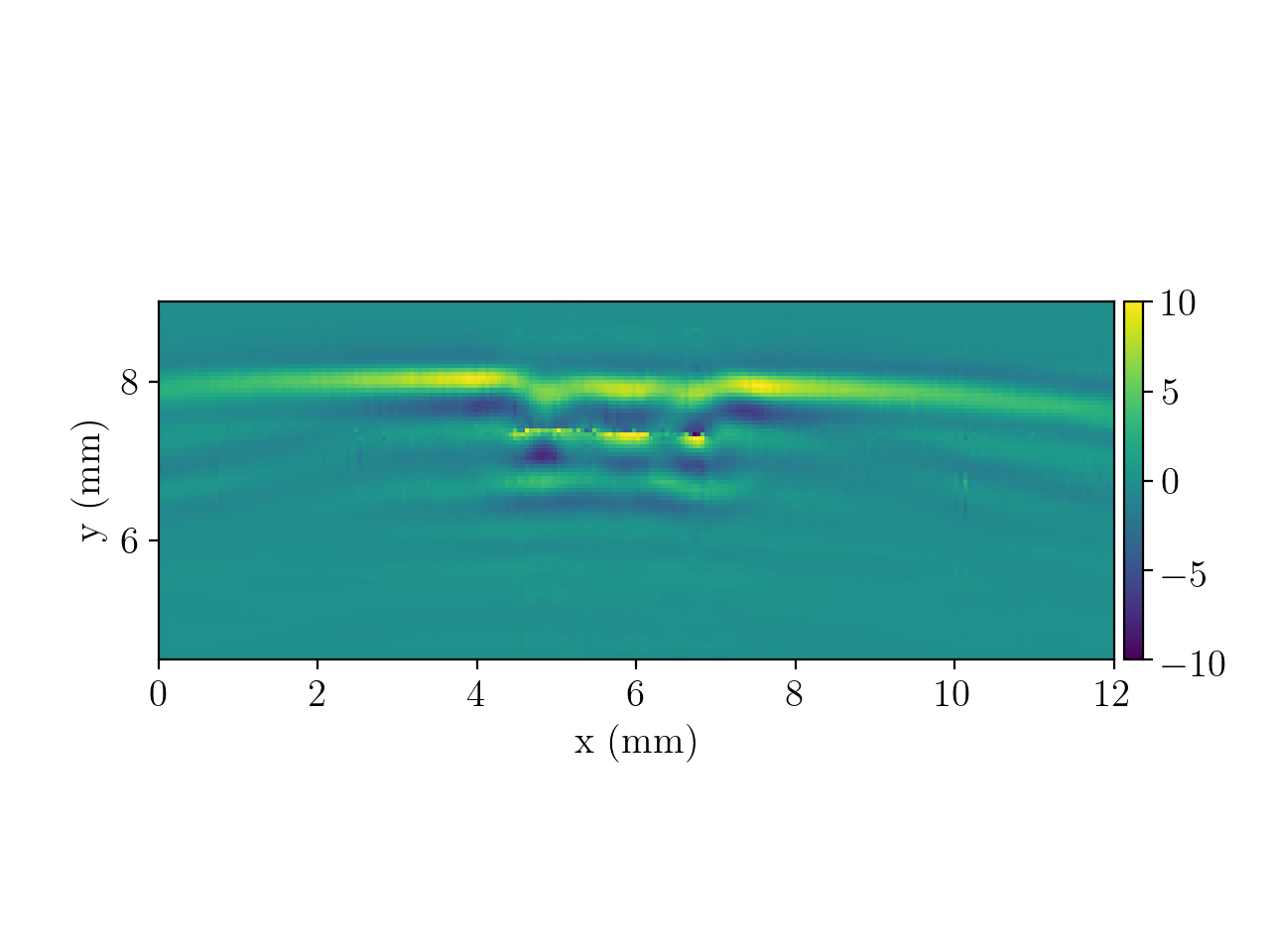}
}
\subfloat[Data discovered from PINN simulation at $t=12.38~\mu s$.]{
\includegraphics[trim=0cm 2.3cm 0cm 0cm, clip, width=0.5\textwidth]{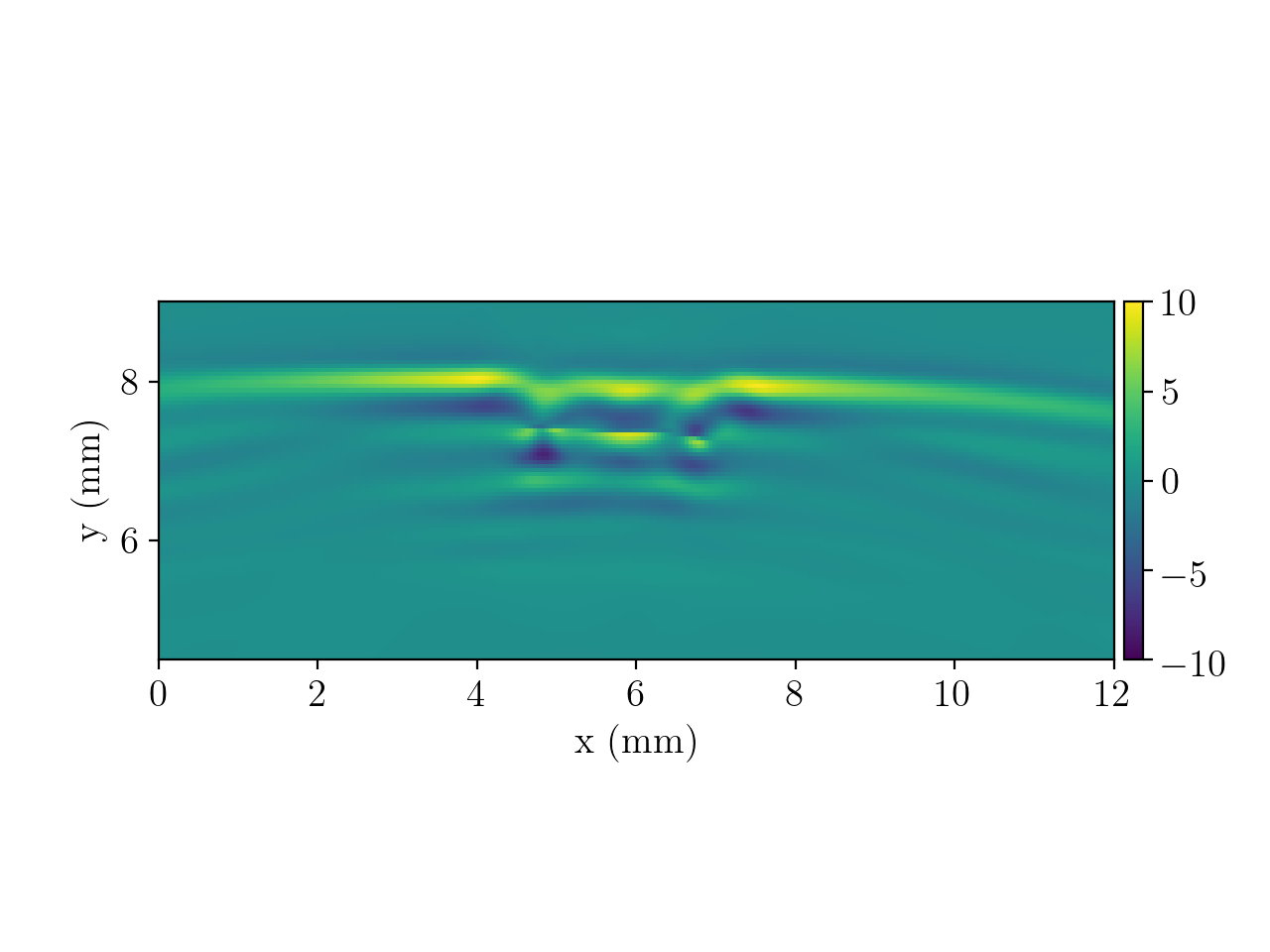}
}\\
\subfloat[Traces of (a) and (b) at $x=6~\text{mm}$ at $t=12.38~\mu s$.]{
\includegraphics[trim=0cm 0cm 0cm 0cm, clip, width=0.5\textwidth]{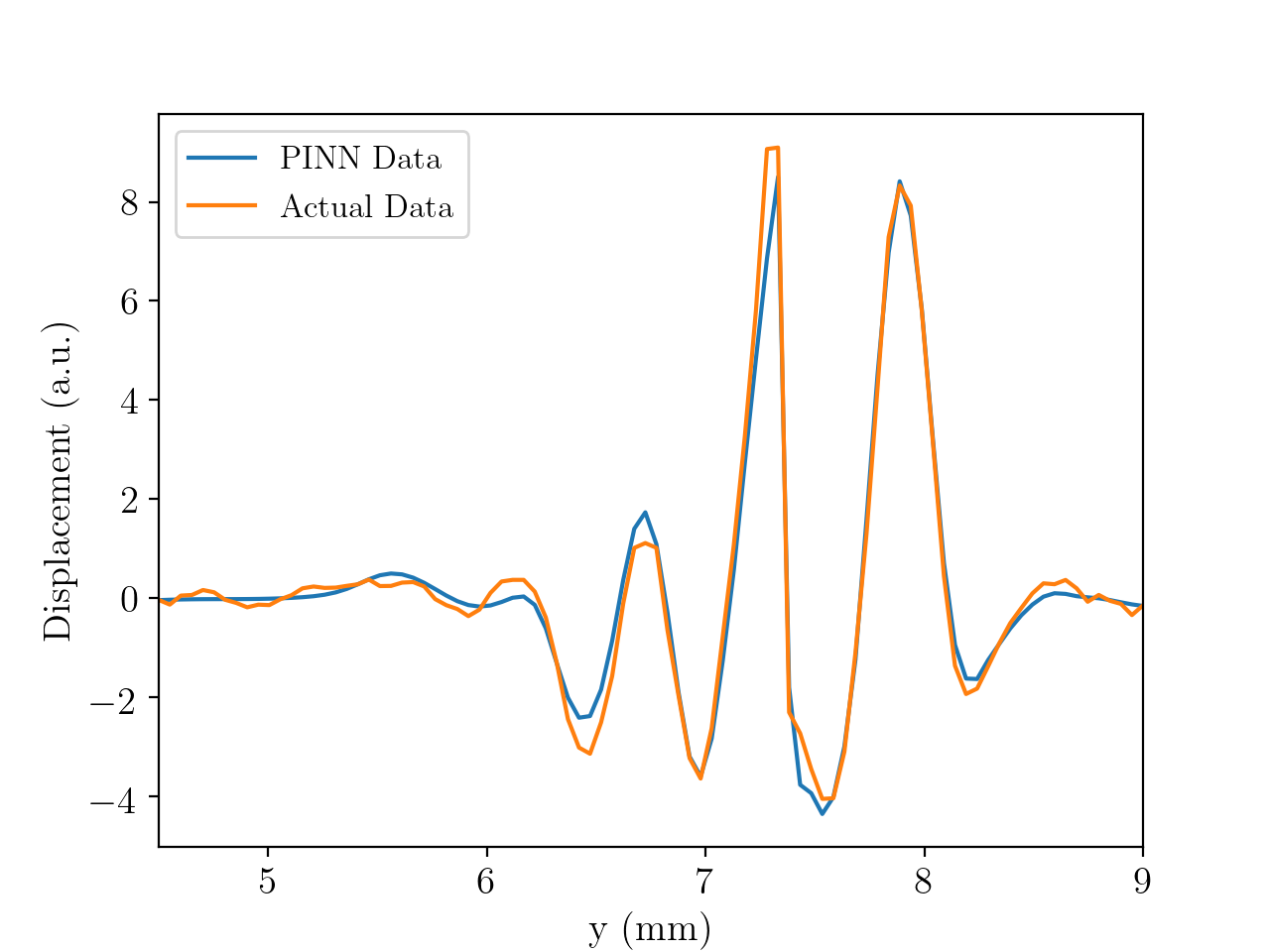}
}
\subfloat[Speed $v(x, y)$ recovered from PINN simulation.]{
\includegraphics[trim=0.5cm 2.4cm 0cm 0cm, clip, width=0.5\textwidth]{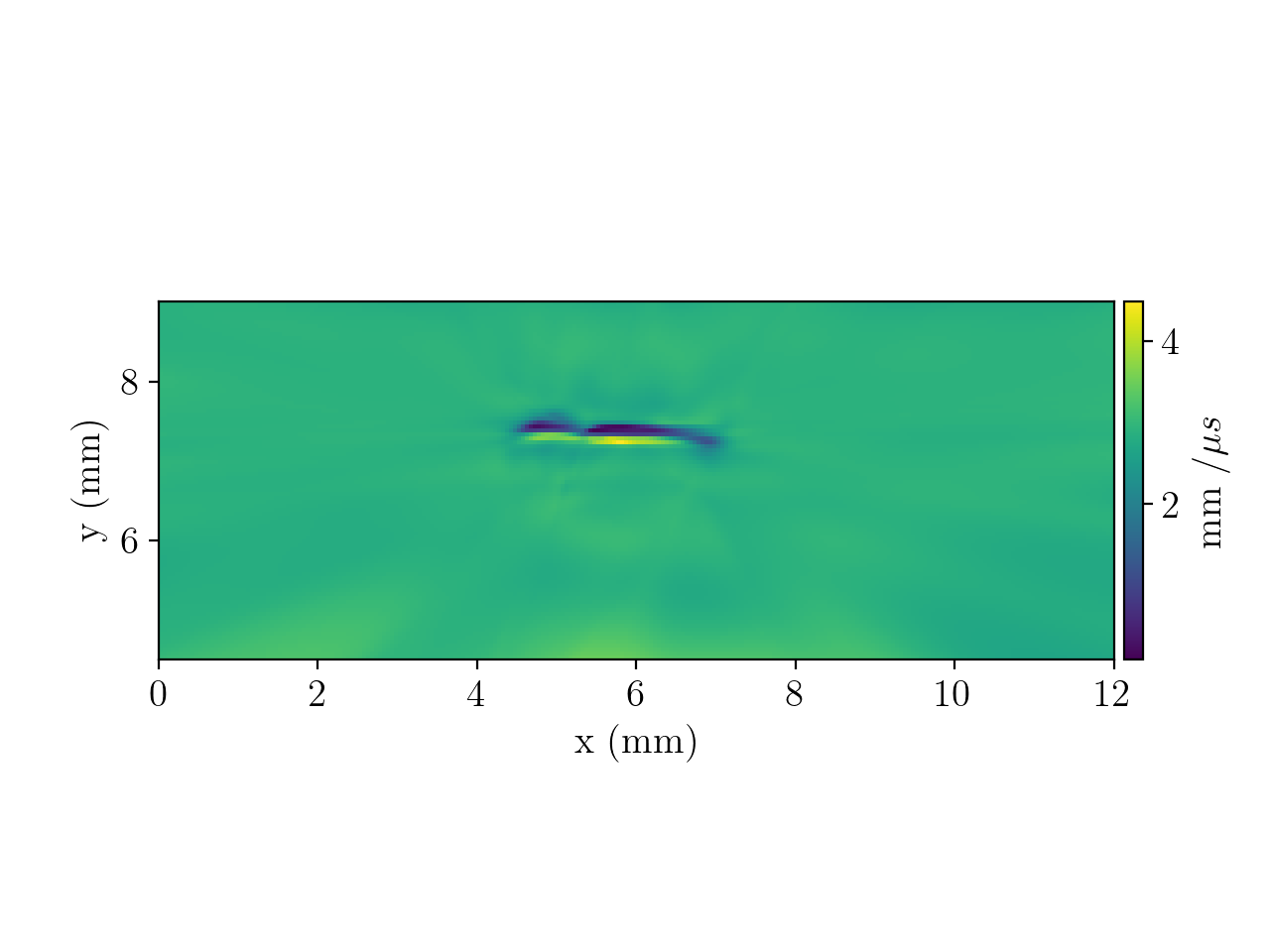}
}\\
\subfloat[Loss function of neural network model.]{
\includegraphics[trim=0cm 0cm 0cm 0cm, clip, width=0.5\textwidth]{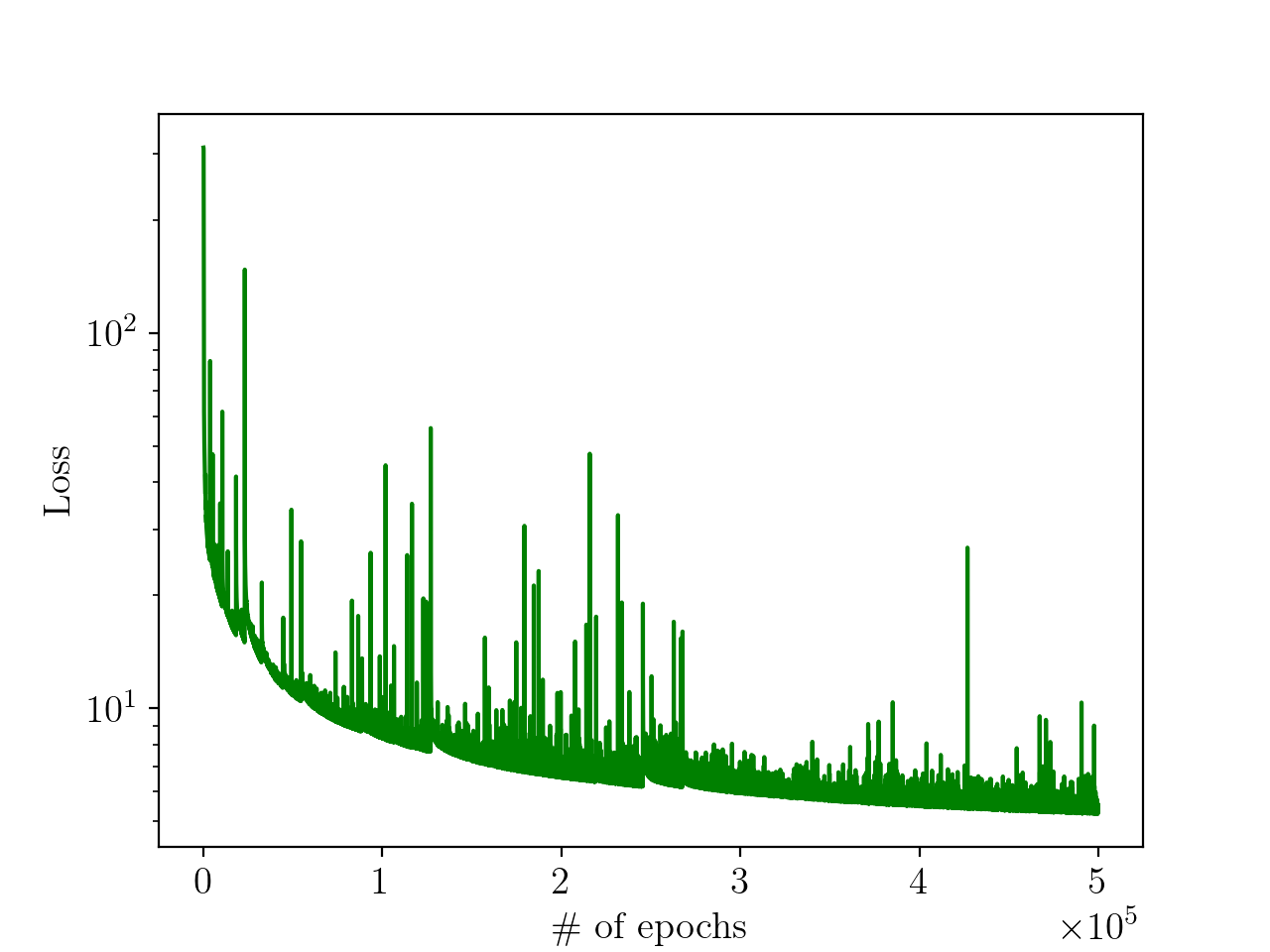}
}
\subfloat[Variation in a with $N=10$.]{
\includegraphics[trim=0cm 0cm 0cm 0cm, clip, width=0.5\textwidth]{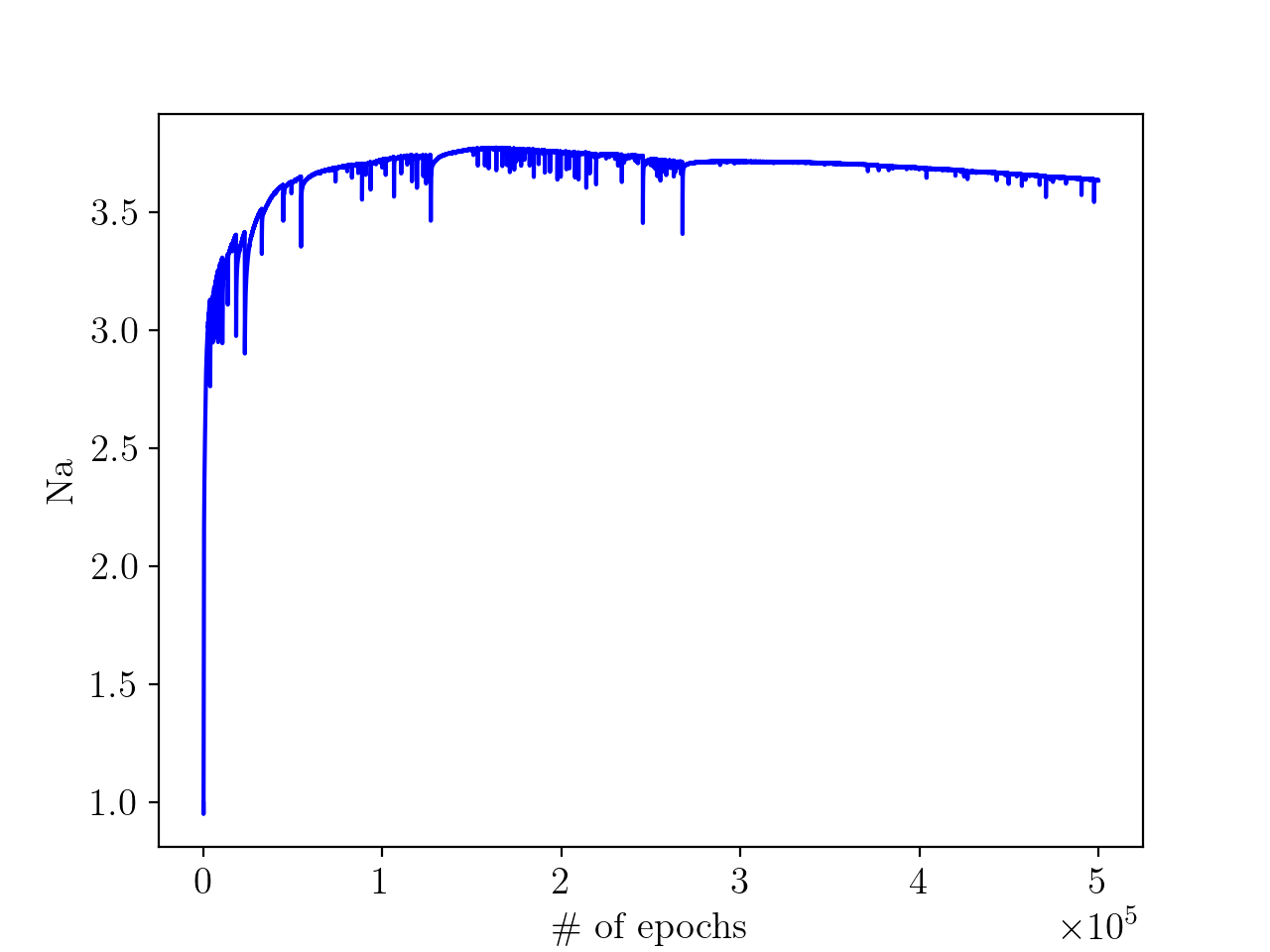}
}\\
\caption{Results from training of neural network model for data acquired at incidence angle of $0^o$ , where (a) and (b) represents the snapshot of the particle displacement from actual and PINN simulated data, respectively. (c) shows a comparison of traces obtained from actual and PINN simulated data. (d) represents the speed of wave as a function of space, where crack is reflected as low speed zone. shows the loss function corresponding to neural network model and (f) shows variation in $a$ with $N=10$.}
\end{figure}

To compute the speed of sound in the 7075-T651 aluminum alloy substrate material, we started the training experiments with the wavefield data acquired at $0^o$ angle and trained the network for the region where the wavefield had not encountered the crack. We used 30 out of the 1000 snapshots (we removed those where the wave was not present or where the wave had already hit the crack), and randomly picked about 20\% of the locations at each snapshot for the training procedure. The neural network in this case had 2 layers each with 32 neurons with a fully connected architecture and an initial learning rate of $5\text{e}-4$. Figures 4a and 4b represent snapshots of the wavefield at $10.8~\mu \text{s}$, obtained from real and PINN simulated data. Figure 4c shows a comparison between traces of the data (real and PINN recovered), extracted at $x=6~\text{mm}$ from Figure 4a and 4b. This shows that traces recovered from the PINN simulated data are in a very good agreement with the real data. Figure 4d provides a plot of the recovered wave speed and the number of training epochs. The sound speed converges to 2.9 $\text{mm}/\mu \text{s}$, which is basically the speed of the surface acoustic wave in an Aluminium alloy, computed from non-destructive testing and reported in \cite{NDTProp}. A comparison of loss function computed with fix and variable $a$ is shown in Figure 4e. Figure 4e clearly shows that usage of adaptive activation function accelerates the convergence. Figure 4f shows a plot of $a$  against number of epochs.

\begin{figure}
\centering
\subfloat[PCA filtered data at $t=11.18~\mu s$.]{
\includegraphics[trim=0cm 0cm 0cm 0cm, clip, width=0.5\textwidth]{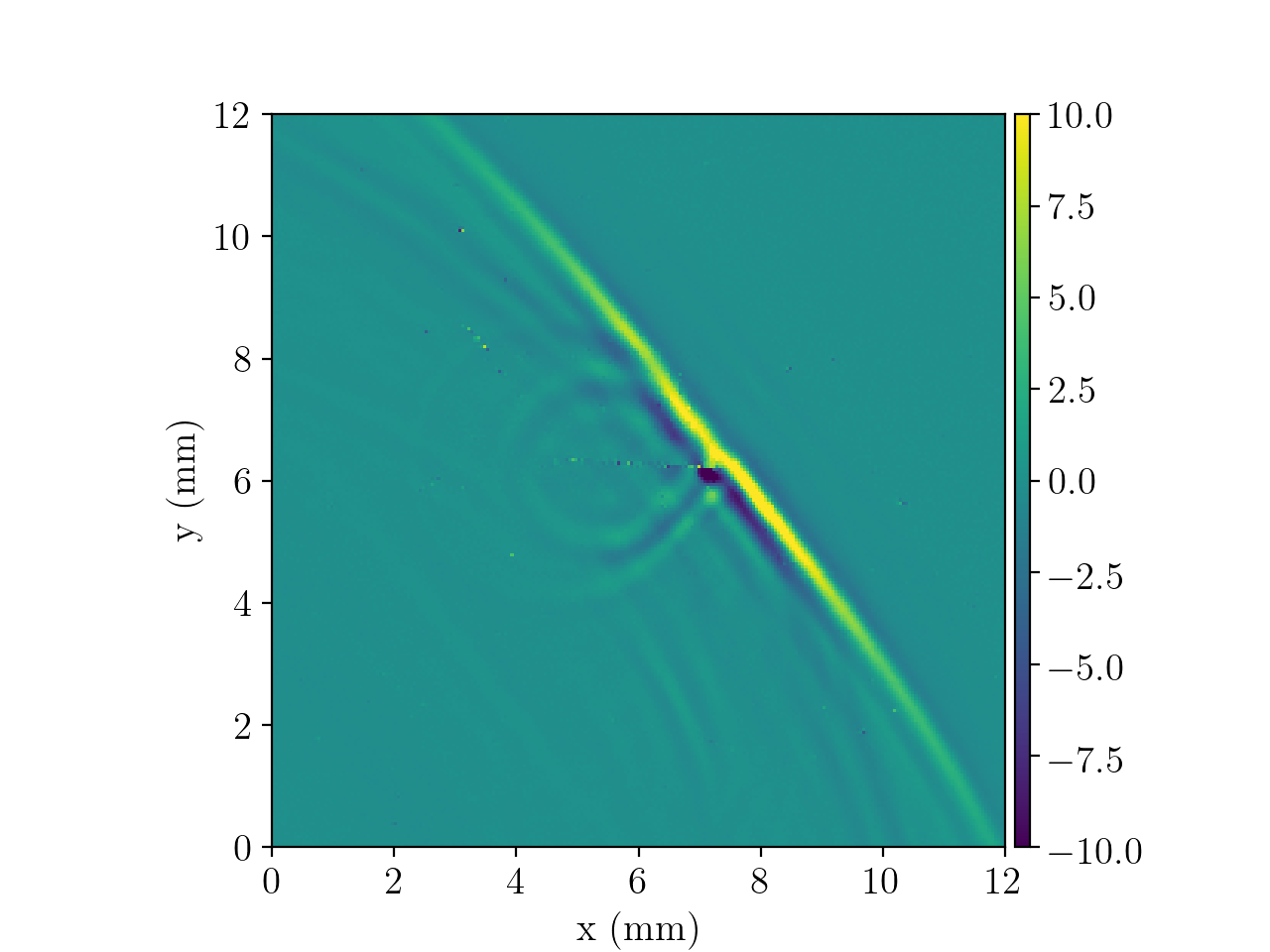}
}
\subfloat[Data recovered from PINN at $t=11.18~\mu s$.]{
\includegraphics[trim=0cm 0cm 0cm 0cm, clip, width=0.5\textwidth]{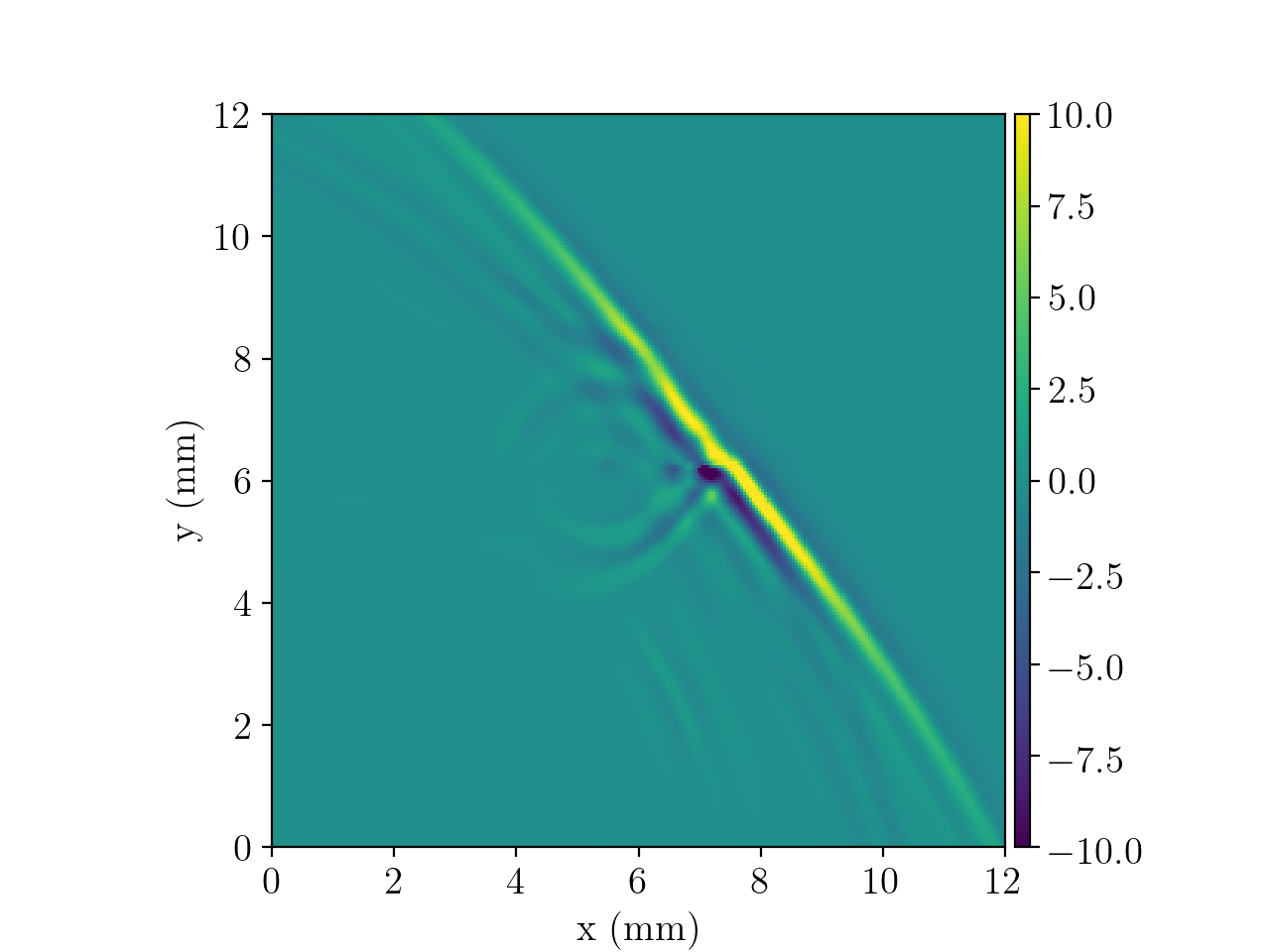}
}\\
\subfloat[ Traces of (a) and (b) at $x=6~\text{mm}$.]{
\includegraphics[trim=0cm 0cm 0cm 0cm, clip, width=0.5\textwidth]{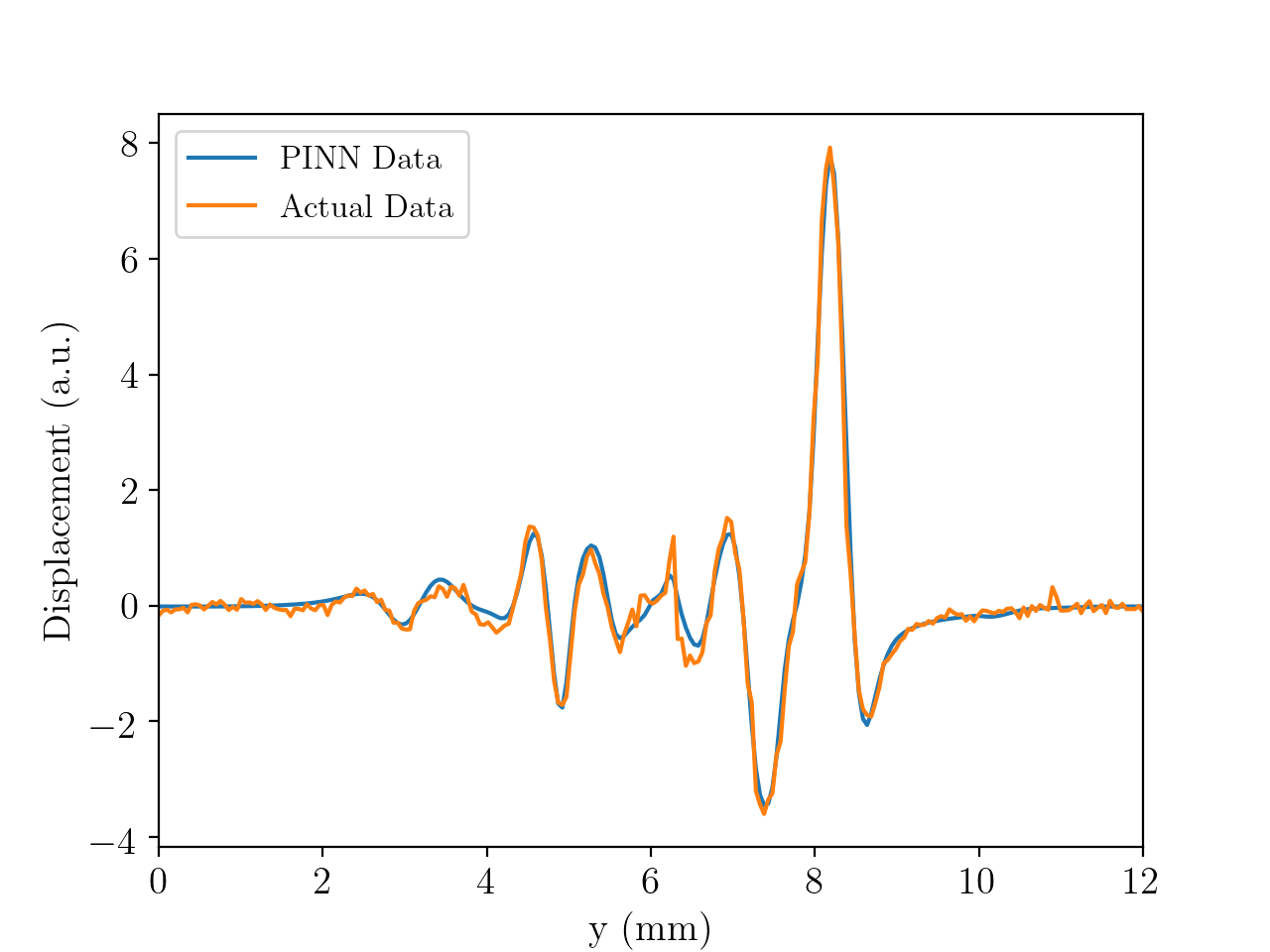}
}
\subfloat[Speed $v(x, y)$ recovered from PINN simulation.]{
\includegraphics[trim=0cm 0cm 0cm 0cm, clip, width=0.5\textwidth]{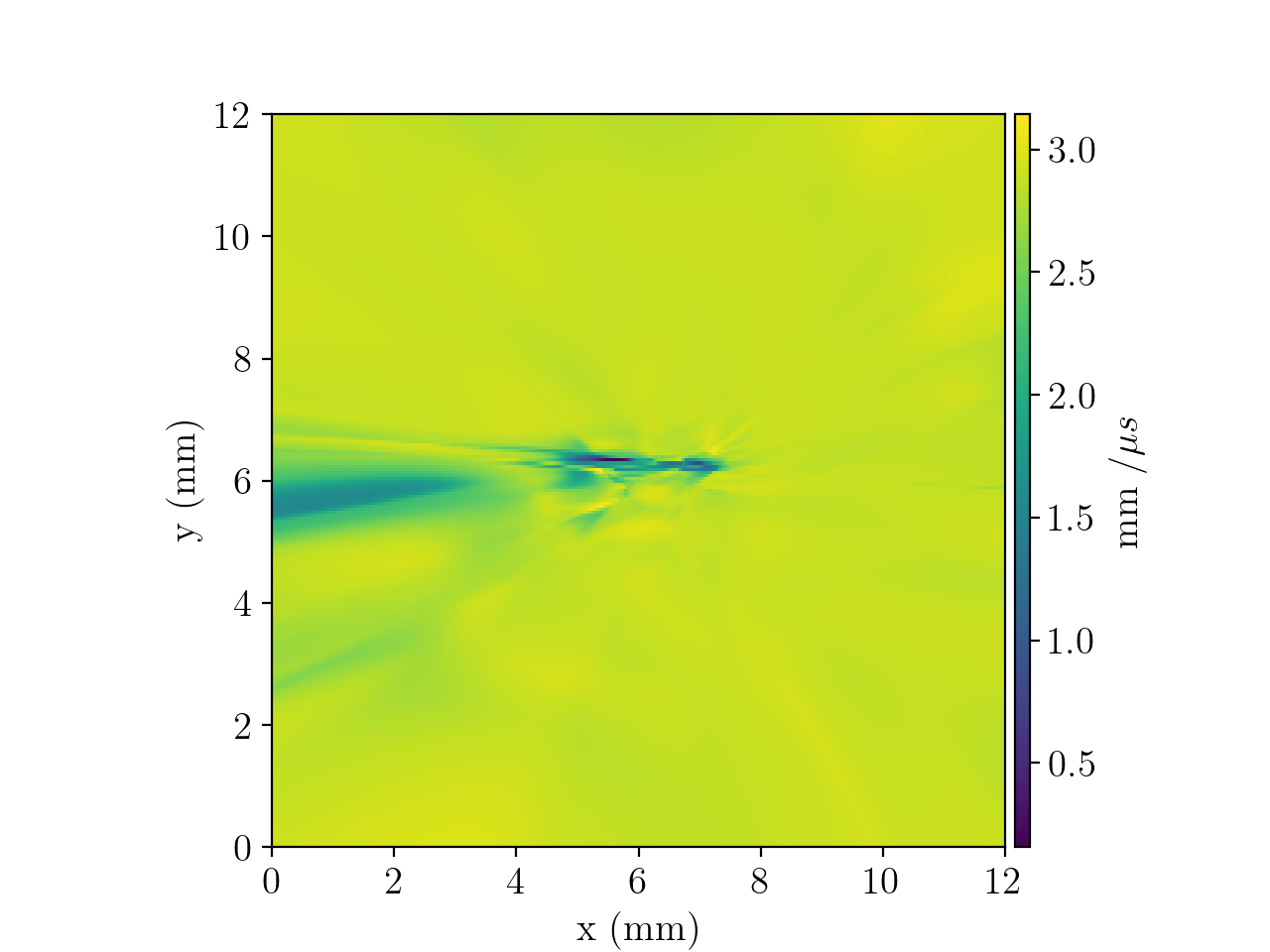}
}
\\
\subfloat[Loss function of neural network model.]{
\includegraphics[trim=0cm 0cm 0cm 0cm, clip, width=0.5\textwidth]{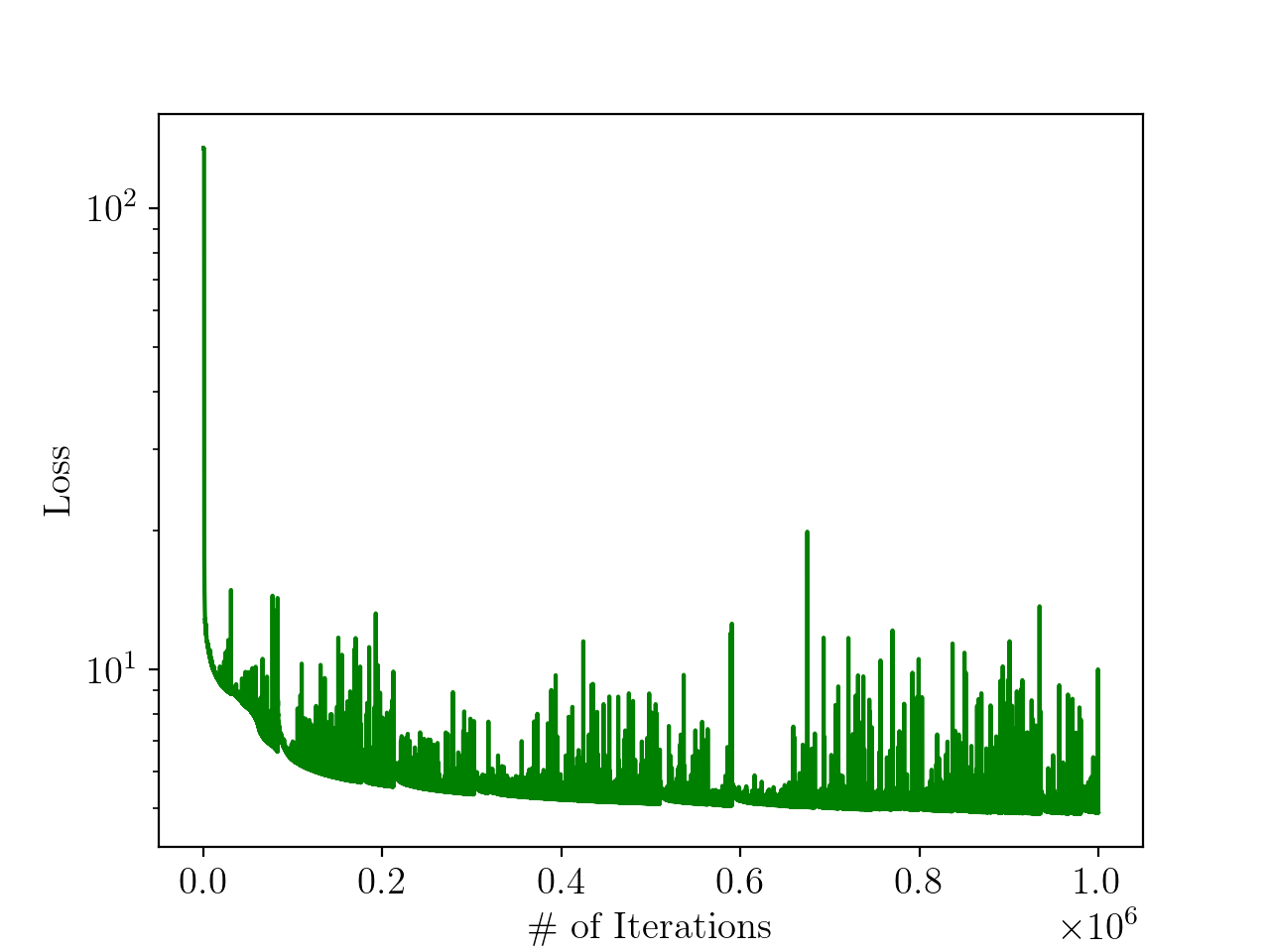}
}
\subfloat[Variation in a with $N=10$.]{
\includegraphics[trim=0cm 0cm 0cm 0cm, clip, width=0.5\textwidth]{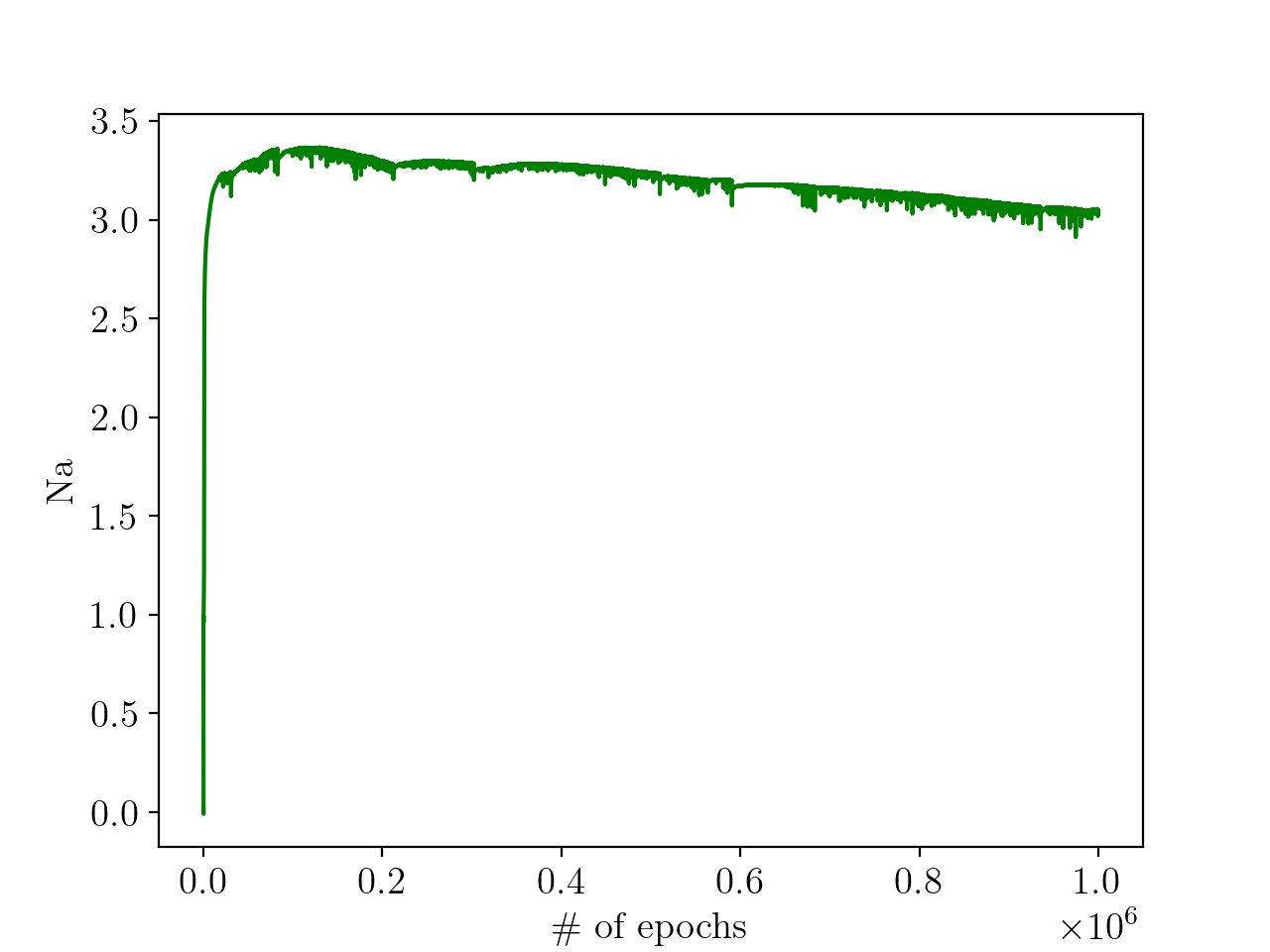}
}\\
\caption{Results from training of neural network model for data acquired at incidence angle of $45^o$ , where (a) and (b) represents the snapshots of the particle displacement from actual and PINN simulated data, respectively. (c) shows a comparison of traces obtained from actual and PINN simualted data and (d) represents the speed of wave as a function of space, where crack is reflected as low speed zone. (e) shows the loss function corresponding to neural network model and (f) shows variation in $a$ with $N=10$. }
\end{figure}
We perform the modeling of the data incorporating the back-scattering coming off the crack. The input to the network includes 40 snapshots and again uses only 40\% of the total points in each snapshot. A fully connected network with width of 96 neurons and depth of 4 layers is used along with an initial learning rate of $5\text{e}-4$. Figure 5a and 5b show snapshots from actual and PINN  simulated data, respectively. To show the accuracy of the PINN predicted results, a trace along $x=6~\text{mm}$ is extracted from actual (Figure 5a) and PINN predicted model (Figure 5b) and plots of these traces are shown in Figure 5c. The prominent mode, represented by maximum amplitude in Figure 5c, is basically back-scattered events coming off the crack. This mode takes maximum computation time to get reconstructed. Figure 5d provides a plot of the wave speeds, recovered as a global variable, against number of training epochs. Similar to the case shown in Figure 4, the speed of the sound converges to 2.9 $\text{mm}/\mu \text{s}$, whereas the sensitivity of the speed of the crack is expressed by a decrease in the speed, shown in the plot Figure 5d. This phenomena motivates the training of the model to recover the speed dependence on space i.e $v(x,y)$ in \ref{eq1}. Figure 5e represents a comparison of loss function computed with fix and variable $a$, which shows that implementation of adaptive activation function helps in accelerating the convergence. Figure 5f show a plot of $a$  against number of epochs.    
\begin{figure}
\centering
\subfloat[PCA filtered data at $t=11.08~\mu s$.]{
\includegraphics[trim=0cm 0cm 0cm 0cm, clip, width=0.5\textwidth]{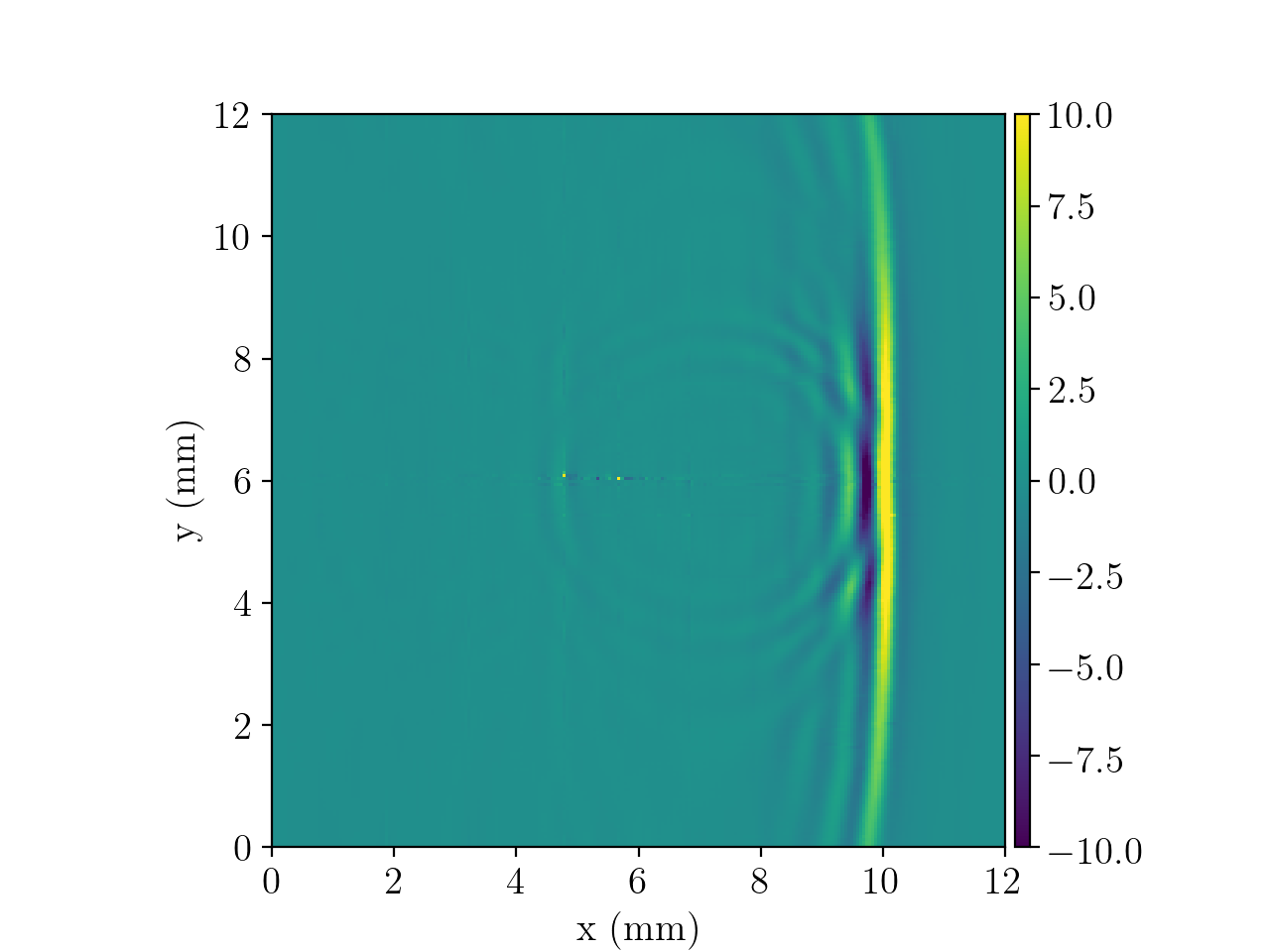}
}
\subfloat[Data recovered from PINN at $t=11.08~\mu s$.]{
\includegraphics[trim=0cm 0cm 0cm 0cm, clip, width=0.5\textwidth]{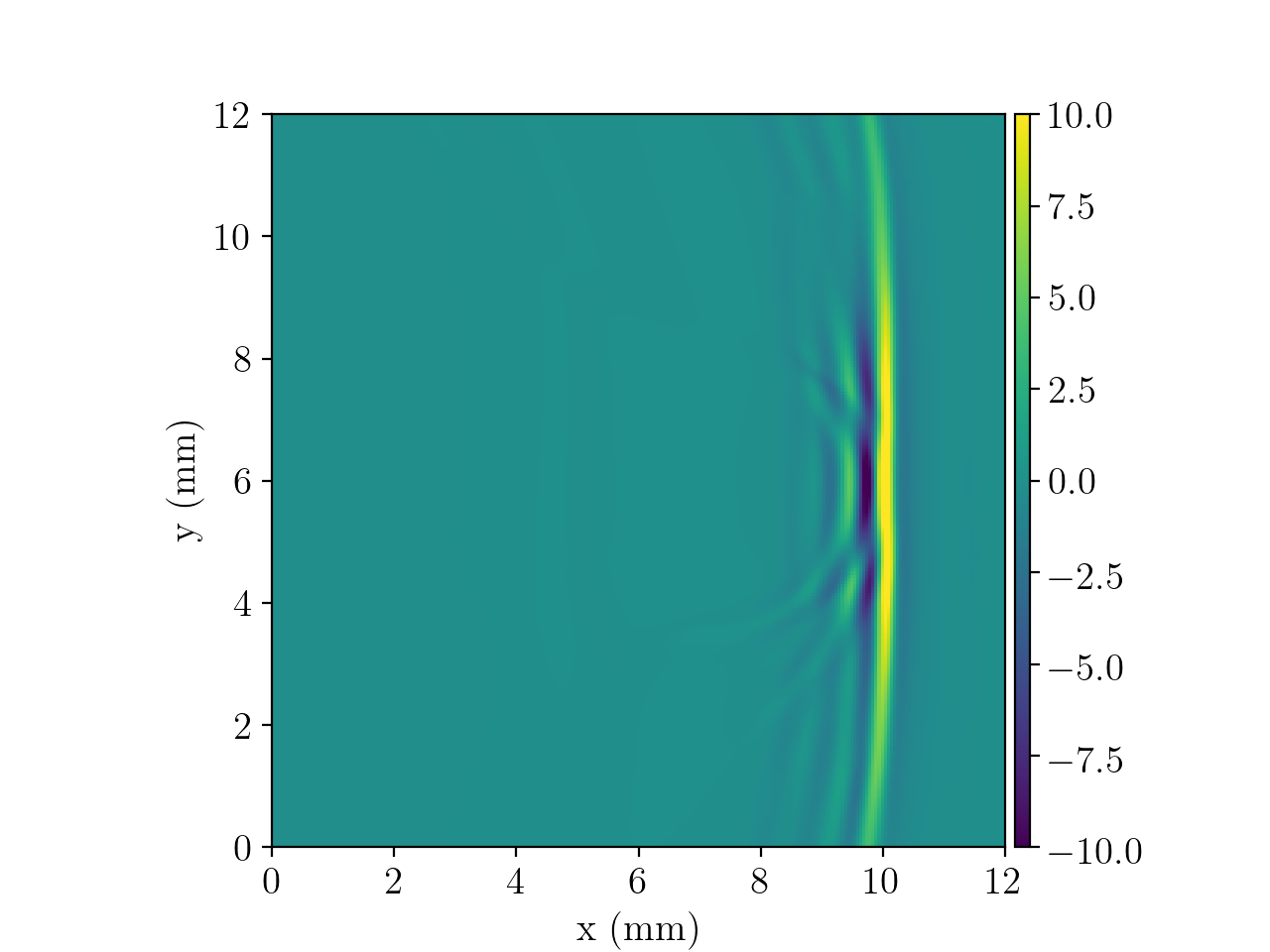}
}\\
\subfloat[Traces of (a) and (b) at $y=11.25~\text{mm}$.]{
\includegraphics[trim=0.5cm 0cm 0cm 0cm, clip, width=0.5\textwidth]{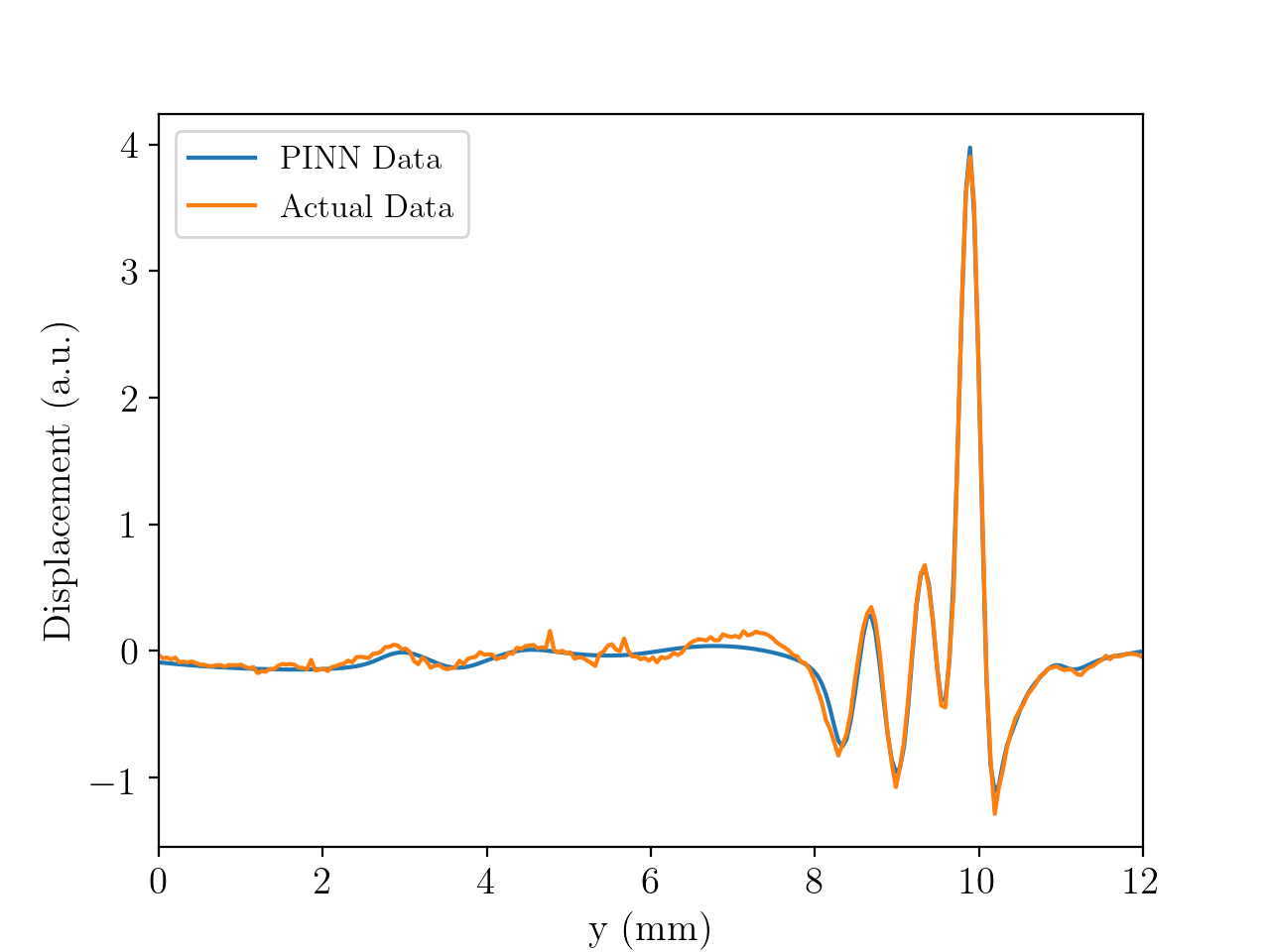}
}
\subfloat[Speed, $v(x, y)$ discovered from PINN simulation.]{
\includegraphics[trim=0cm 0cm 0cm 0cm, clip, width=0.5\textwidth]{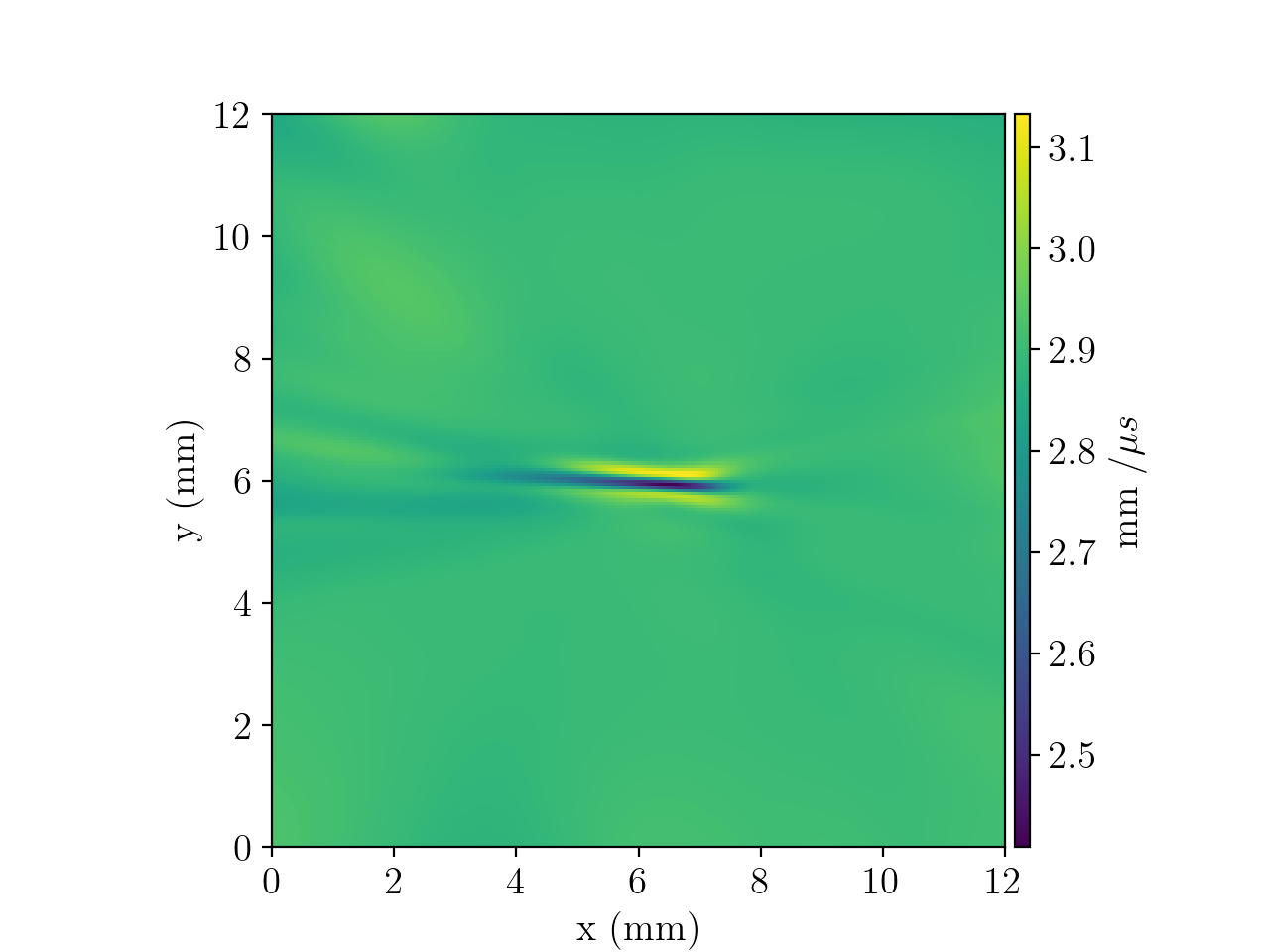}
}\\
\subfloat[Loss function of neural network model.]{
\includegraphics[trim=0cm 0cm 0cm 0cm, clip, width=0.5\textwidth]{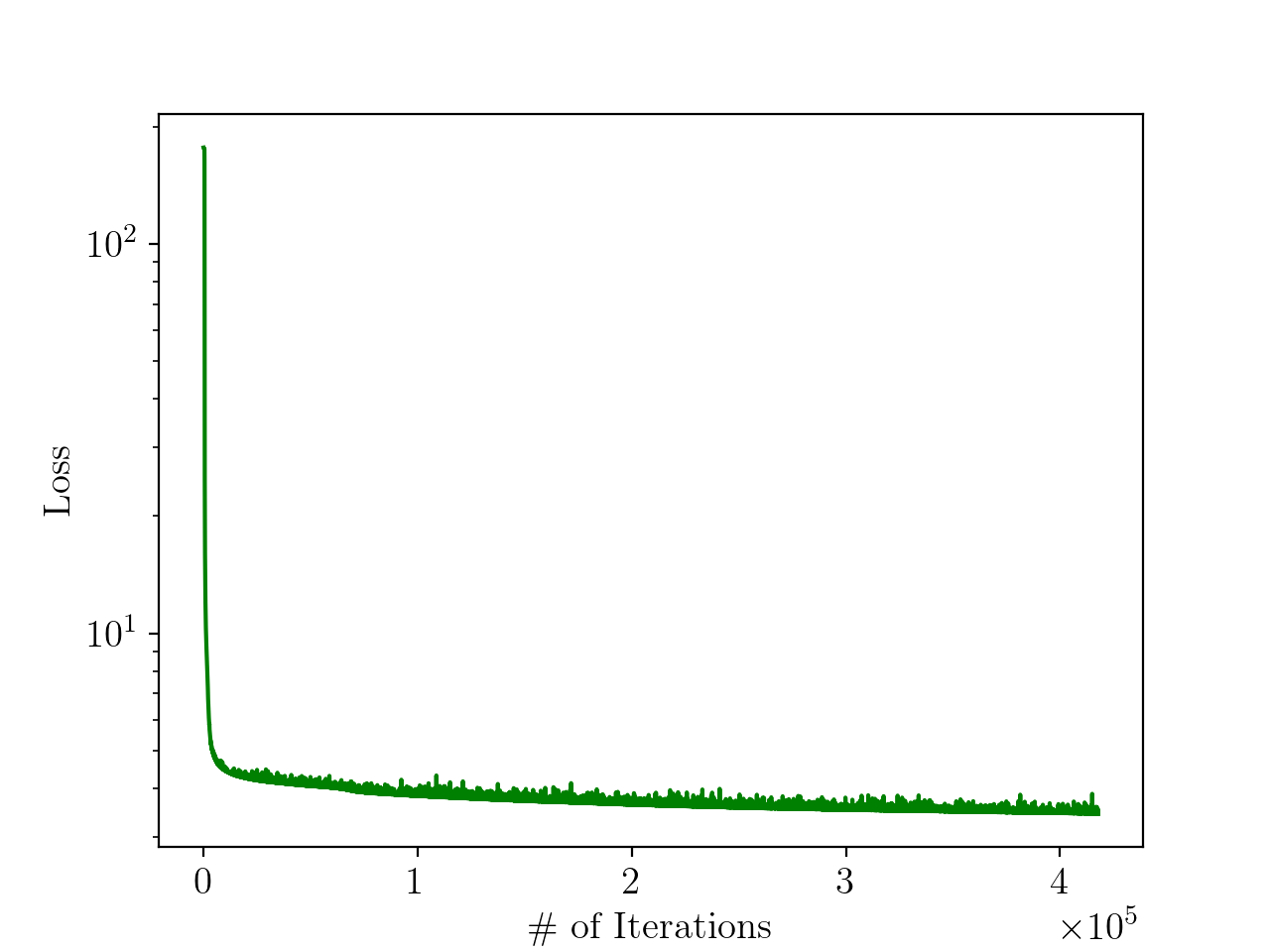}
}
\subfloat[Variation in $a$ with $N=10$.]{
\includegraphics[trim=0cm 0cm 0cm 0cm, clip, width=0.5\textwidth]{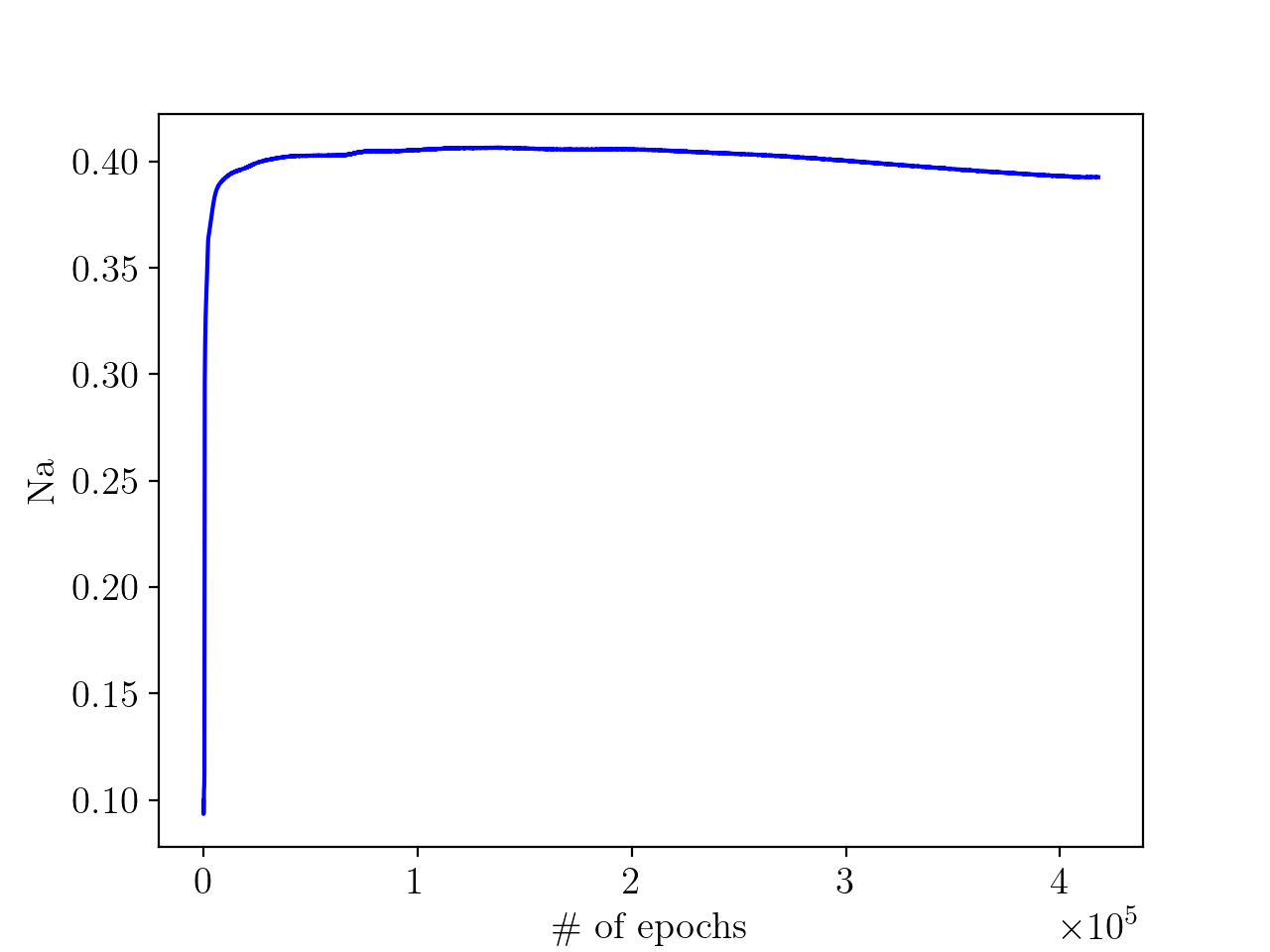}
}
\caption{Results from training of the neural network model for data acquired at incidence angle of $90^o$ , where (a) and (b) represents the snapshot of the particle displacement from actual and PINN recovered data, respectively. (c) shows a comparison of traces obtained from actual and PINN simulated data and (d) represents the speed of wave as a function of space, where crack is reflected as low speed zone. (e) shows the loss function corresponding to neural network model and (f) shows variation in $a$ with $N=10$.}
\end{figure}
Now, we train the PINN to recover $v(x,y)$ to detect the location the crack in the aluminum substrate. To achieve this we used two neural networks, the first one to approximate the speed of wave and then this speed is fed into equation (\ref{eq1}) while computing the loss function defined in (\ref{eq5}) for PINN. The input to the PINN includes 40 snapshots, and again we use only 20\% of the total points in each snapshot. The PINN and velocity networks are  fully connected networks with width of 64 neurons and depth of 4 layers, and an initial learning rate of $5\text{e}-4$.  Figures 6a and 6b represent snapshots at $t=12.38~\mu s$ obtained from real data and PINN model. A video (movie$\_$ds1.mp4) showing the real and PINN simulated data against the time stamp is uploaded as an additional material. A comparison of traces extracted from Figures 6a and 6b at $x=6~\text{mm}$ is shown in Figure 6c. Figure 6d shows the variation of speed of wave $v(x,y)$. The crack is characterized by the zone of low speed with the speed decreasing to  $0.5~\text{mm}/\mu \text{s}$ from the surrounding, which is 2.9 $\text{mm}/\mu \text{s}$. Figure 6e represents the loss function showing that the error reduces to 1.2 \% even with the 20\% data, which also includes the effect of crack. Figure 6f represents the variation of $a$  against number of epochs during the training process. 
\begin{figure}
\centering
\subfloat[PCA filtered data at $t= 12.38 \mu\text{s}$ ]{
\includegraphics[trim=0cm 0cm 0cm 0cm, clip, width=0.33\textwidth]{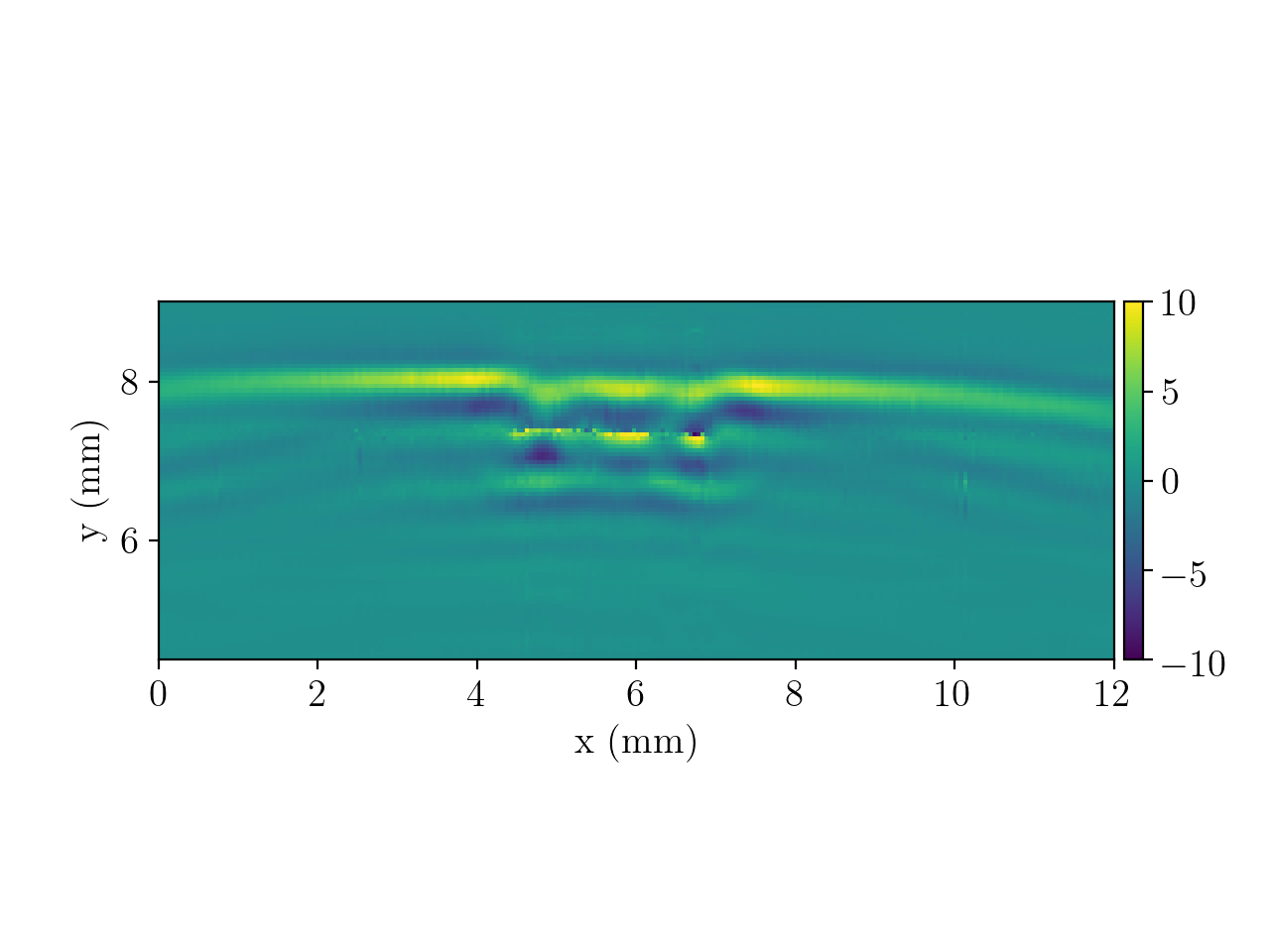}
}
\subfloat[PINN simulation using PCA filtered data at $t= 12.38 \mu\text{s}$ ]{
\includegraphics[trim=0cm 0cm 0cm 0cm, clip, width=0.33\textwidth]{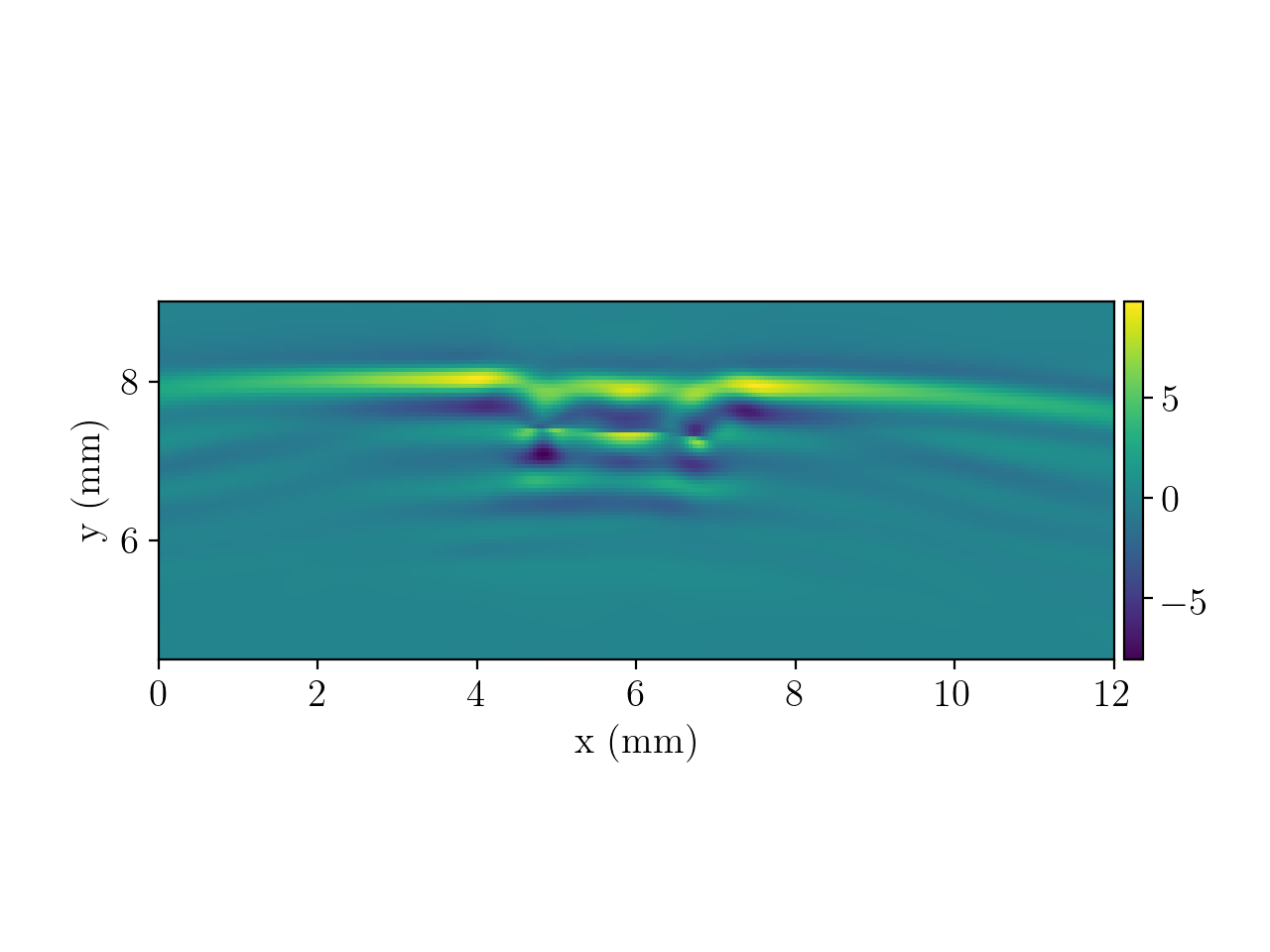}
}
\subfloat[Relative error between (a) and (b) .]{
\includegraphics[trim=0cm 0cm 0cm 0cm, clip, width=0.33\textwidth]{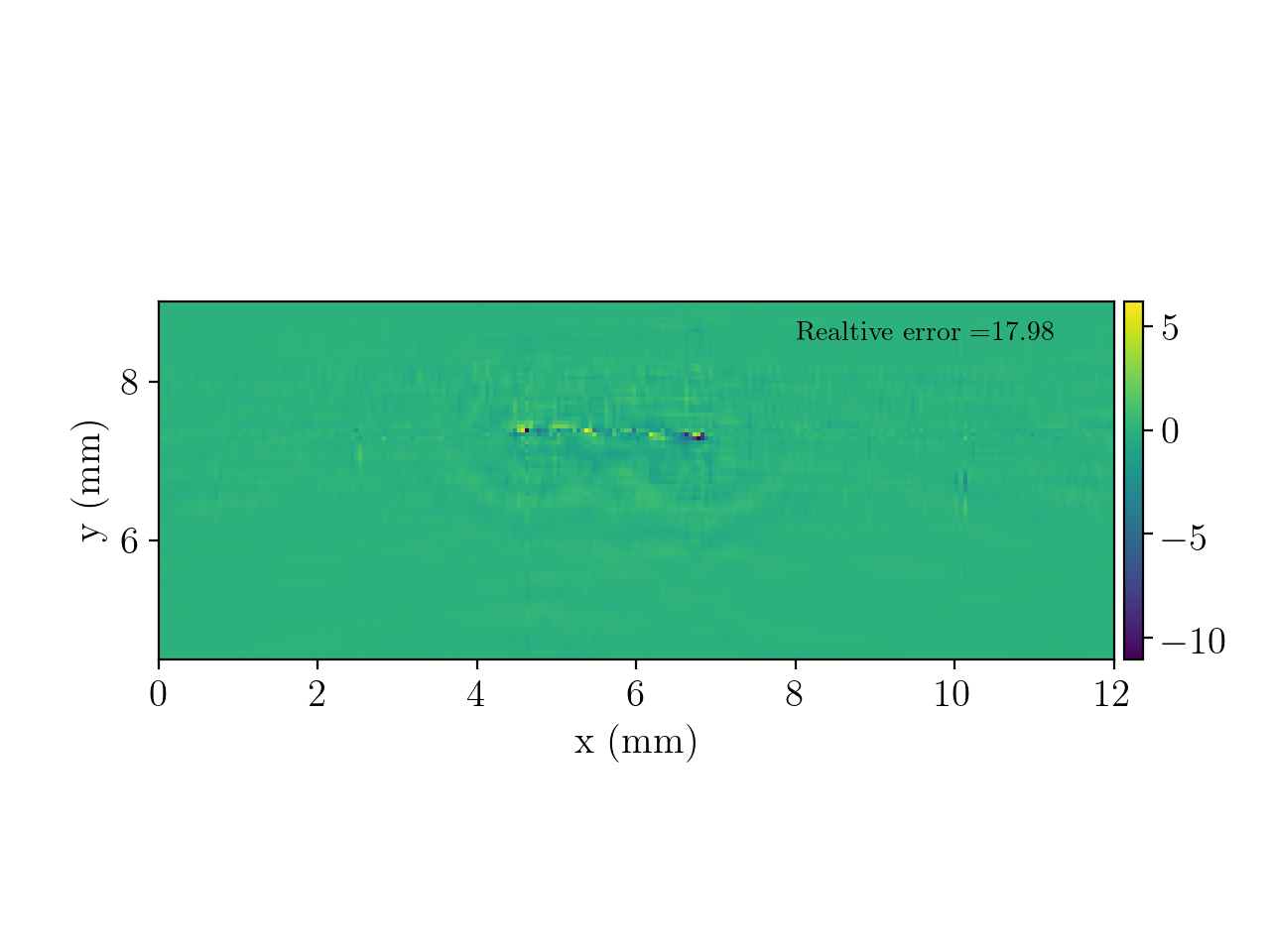}
}\\
\subfloat[Raw data at $t= 12.38 \mu\text{s}$ ]{
\includegraphics[trim=0cm 0cm 0cm 0cm, clip, width=0.33\textwidth]{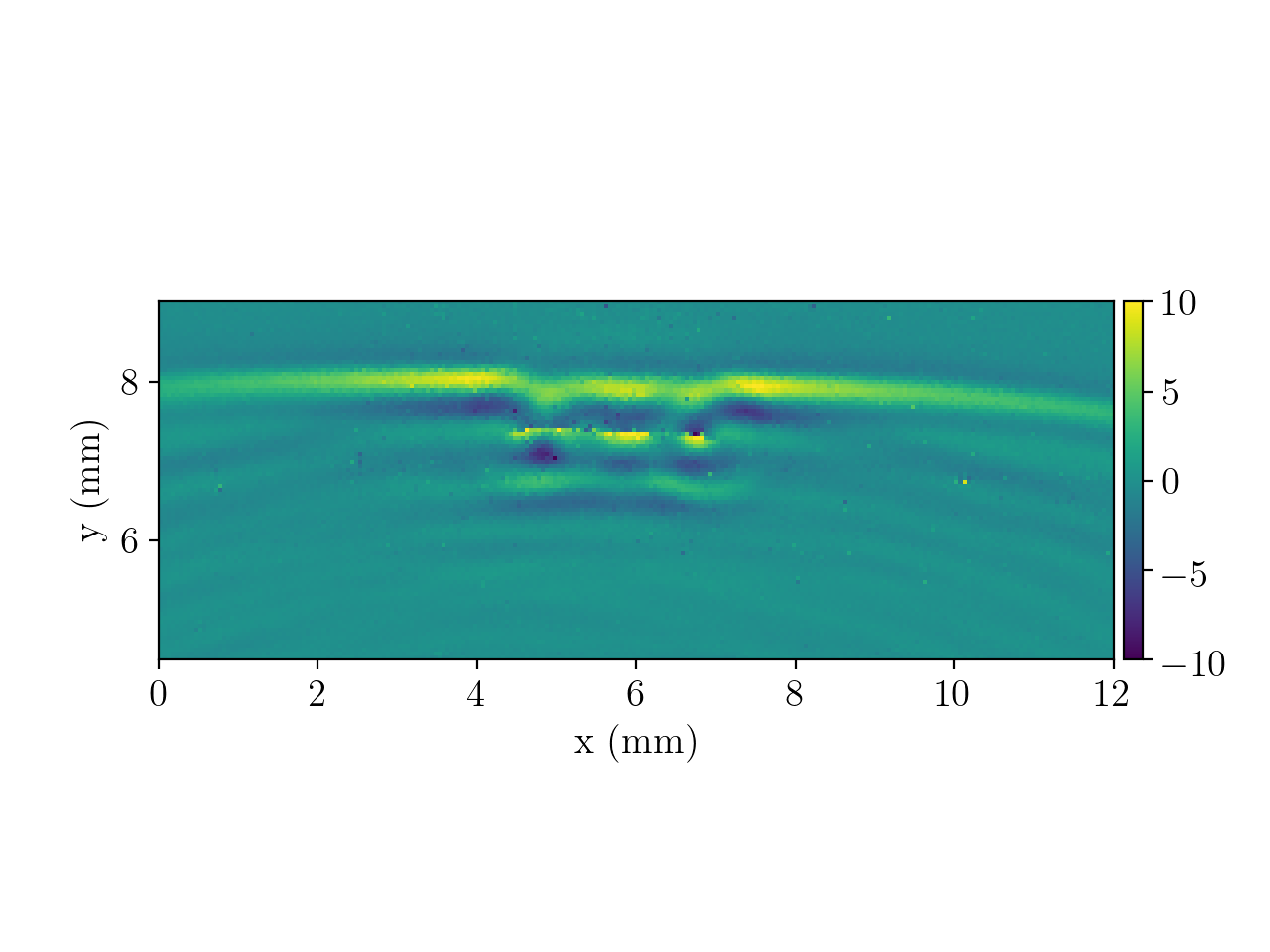}
}
\subfloat[PINN simulation using raw data at $t= 12.38 \mu\text{s}$]{
\includegraphics[trim=0cm 0cm 0cm 0cm, clip, width=0.33\textwidth]{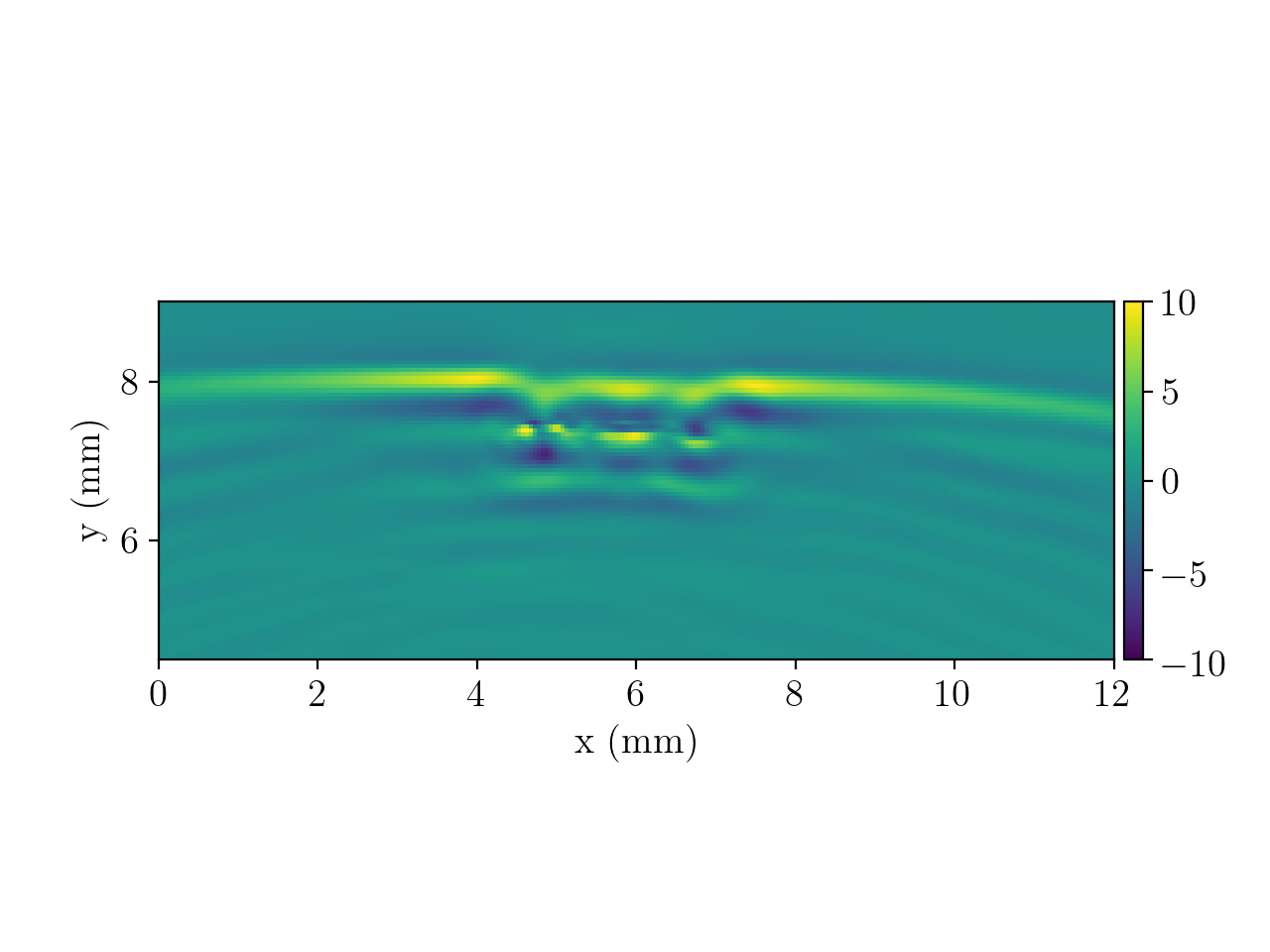}
}
\subfloat[Relative error between (d) and (e).]{
\includegraphics[trim=0cm 0cm 0cm 0cm, clip, width=0.33\textwidth]{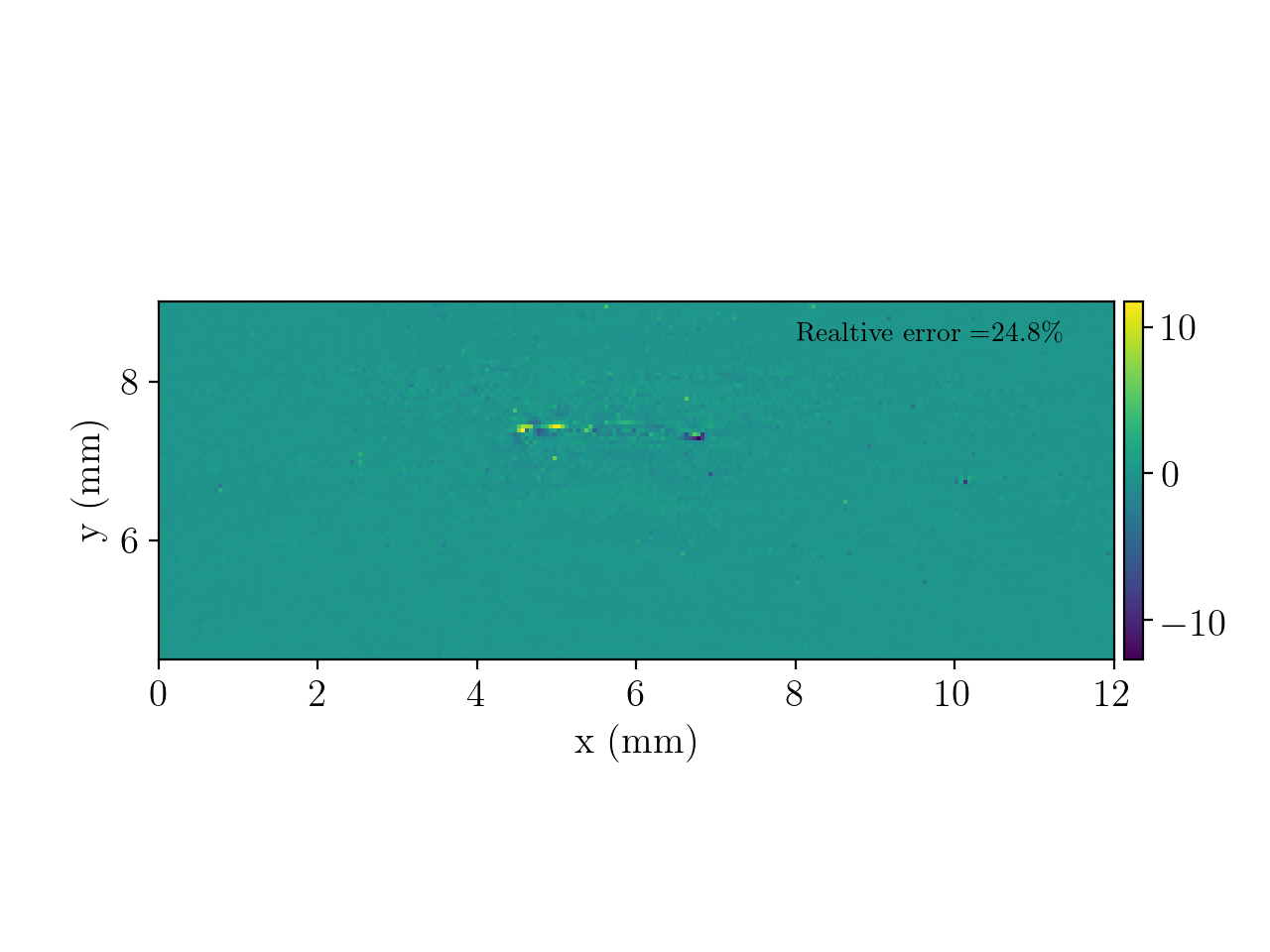}
}
\caption{Application of PCA on the data where (a) and (b) PCA filtered data and PINN simulated data using filtered data, respectively. (c) represents the error between (a) and (b) with a relative error of $17.98\%$. (d) and (e) represent raw and PINN simulated data using raw data, respectively. (f) represents the error between (d) and (e) with a relative error of $24.80\%$. }
\end{figure}

Next, we train the PINN to recover $v(x,y)$ to detect the location of crack from the data acquired at incidence angle of 45$^o$. We consider 80 snapshots and each snapshot is sampled with 10\% of total points from entire domain i.e $(12~\text{mm} \times 12~\text{mm})$. The PINN and velocity network are a fully connected network with width of 32 neurons and depth of 6 layers, and an initial learning rate of $5\text{e}-4$. Figures 7a and 7b represent the snapshots of the wavefields at $t=11.18~\mu s$ extracted from  real data and PINN simulated data. A video (movie$\_$ds2.mp4) showing the real and PINN simulated data against the time stamp is uploaded as an additional material. Figure 7c shows a comparison between the traces of data extracted at $x=6~\text{mm}$ from real (Figure 7a) and PINN simulated data (Figure 7b), which shows a very good agreement. Figure 7d shows spatial variation of wave speed in the entire domain with a zone of low speed of 0.5 $\text{mm}/\mu \text{s}$, representing the presence of the crack. Figure 7e represents the loss function showing that error reduces to 1 \% even with the 10\% data. Figure 7f represents the variation of $a$  against number of epochs during the training process
\begin{figure}
\centering
\subfloat[Data discovered from PINN simulation using 10 \% data.]{
\includegraphics[trim=0.5cm 2.4cm 0cm 0cm, clip, width=0.5\textwidth]{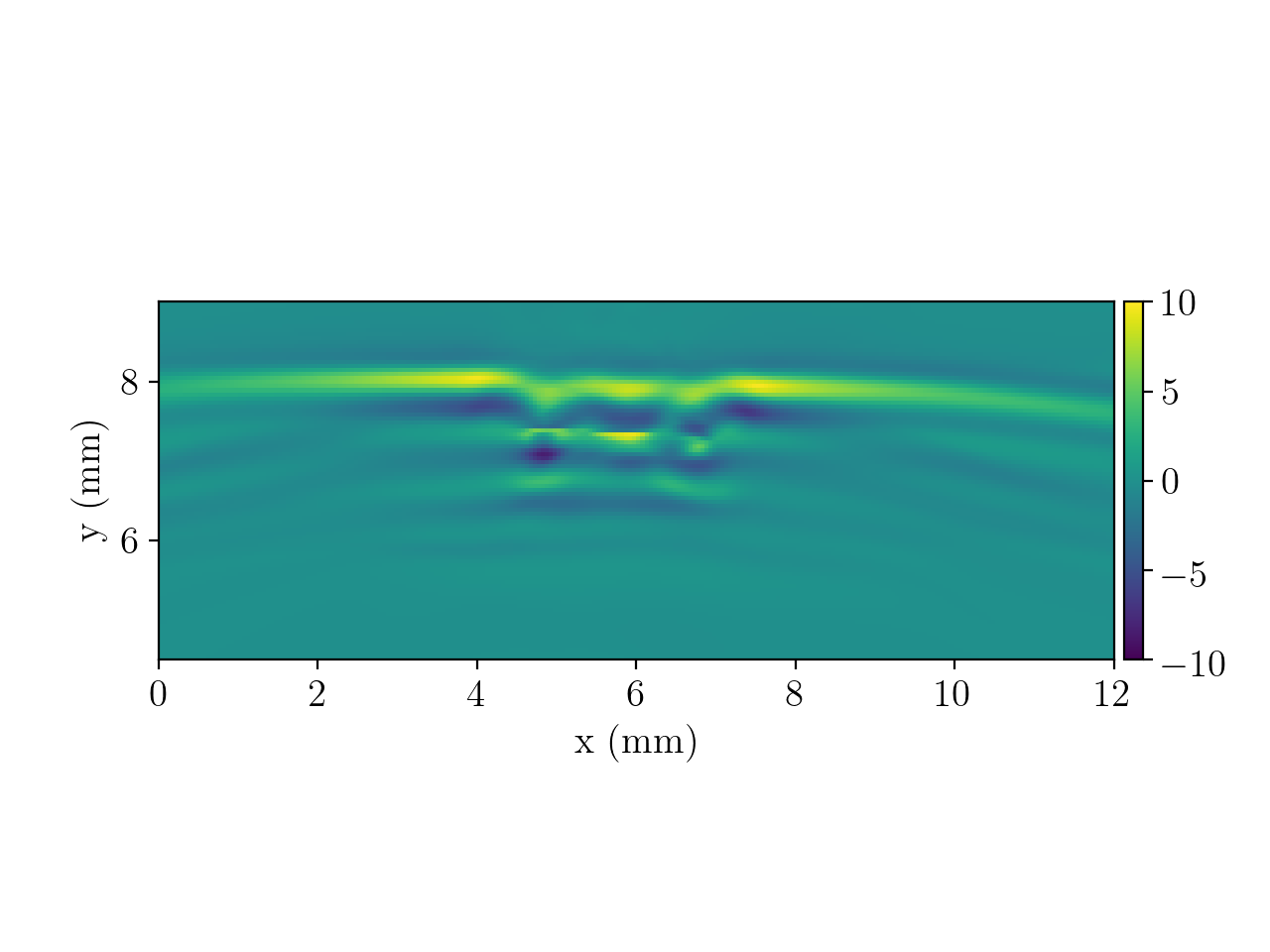}
}
\subfloat[Data discovered from PINN simulation using 20 \% data.]{
\includegraphics[trim=0.5cm 2.4cm 0cm 0cm, clip, width=0.5\textwidth]{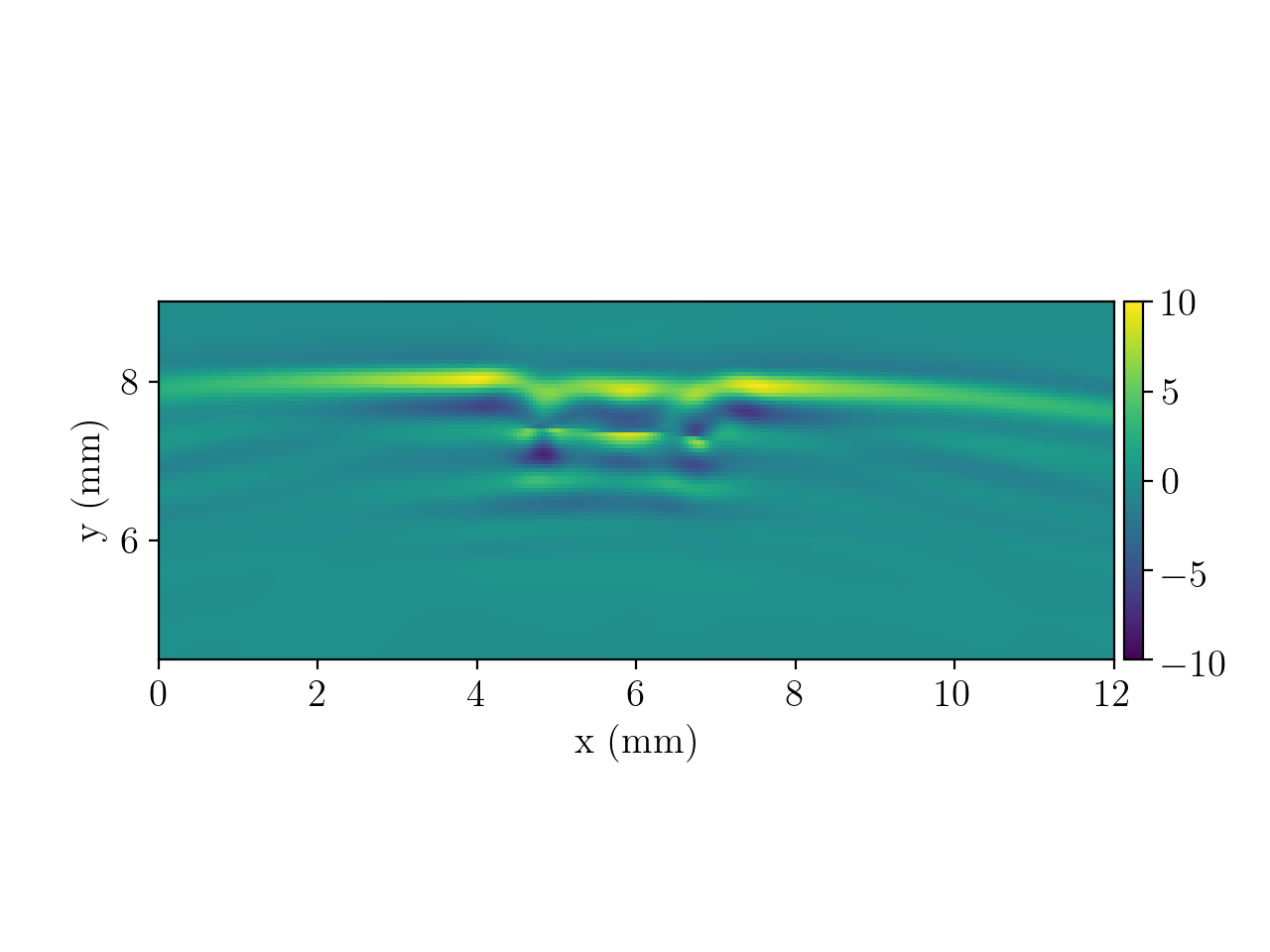}
}\\
\subfloat[$v(x, y)$ discovered from PINN simulation using 10 \% of data.]{
\includegraphics[trim=0.5cm 2.4cm 0cm 0cm, clip, width=0.5\textwidth]{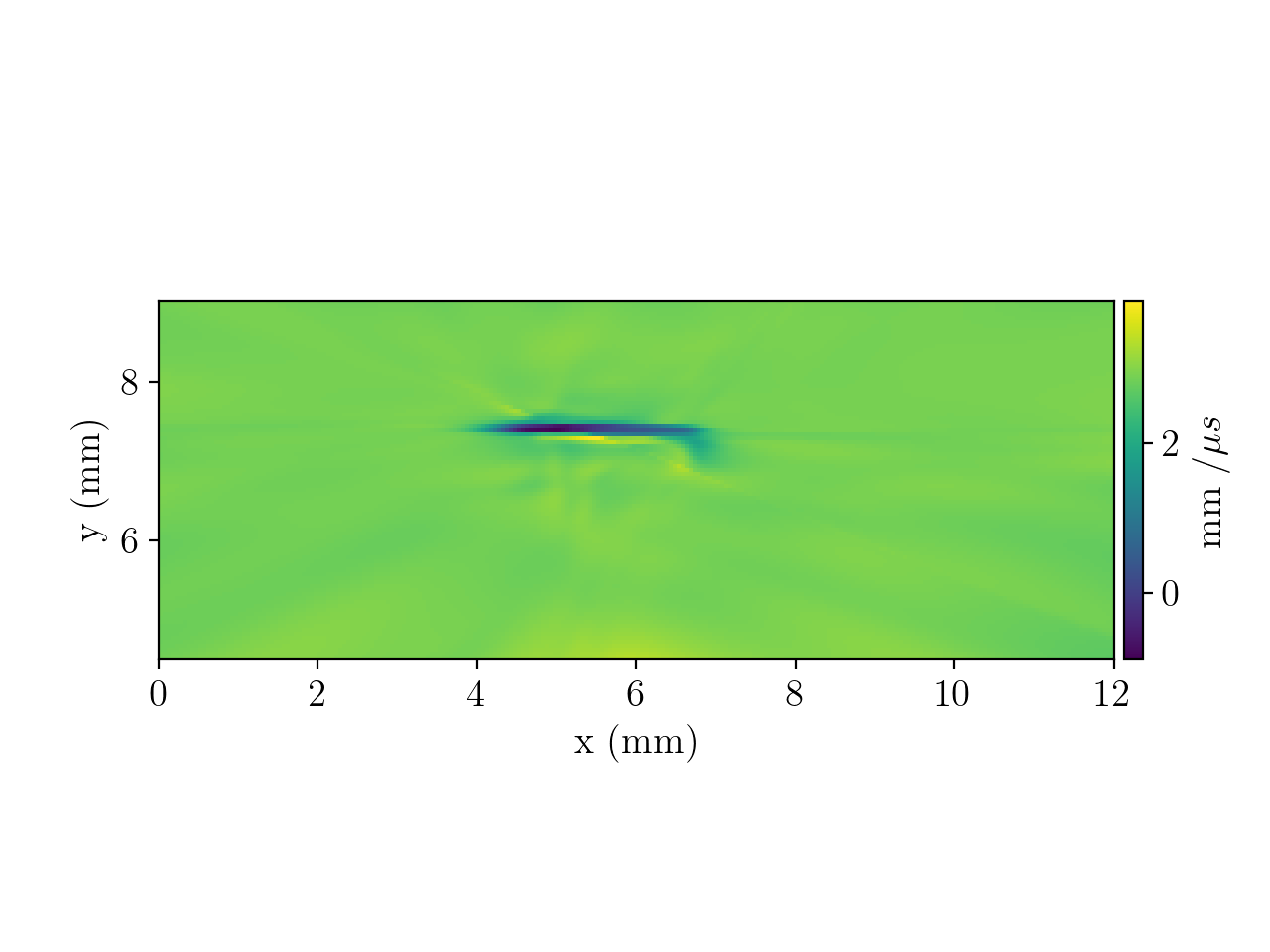}
}
\subfloat[$v(x, y)$ discovered from PINN simulation using 20 \% of data.]{
\includegraphics[trim=0.5cm 2.4cm 0cm 0cm, clip, width=0.5\textwidth]{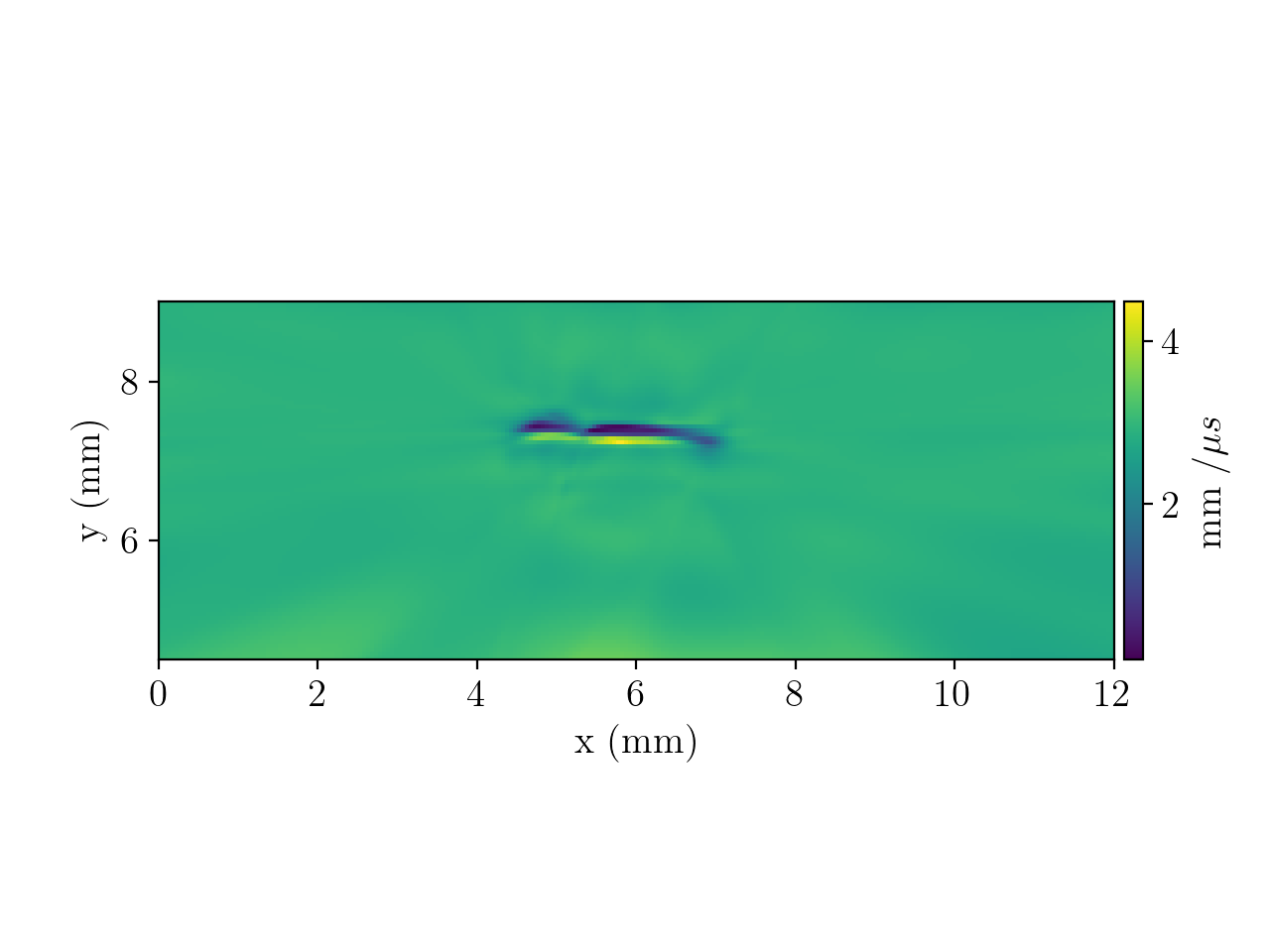}
}\\
\subfloat[Loss function of neural network model.]{
\includegraphics[trim=0cm 0cm 0cm 0cm, clip, width=0.5\textwidth]{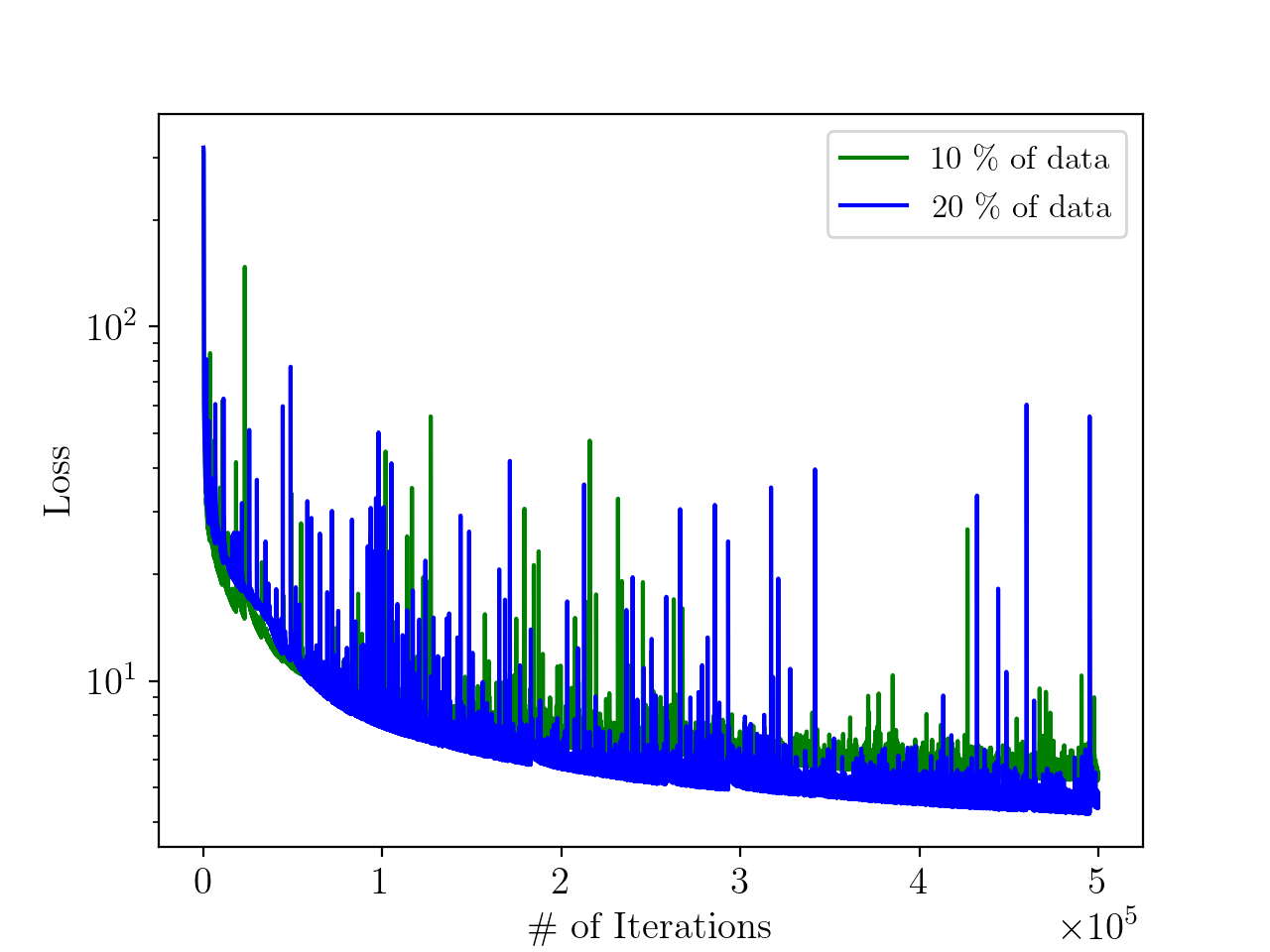}
}
\subfloat[Variation in a with $n=10$.]{
\includegraphics[trim=0cm 0cm 0cm 0cm, clip, width=0.5\textwidth]{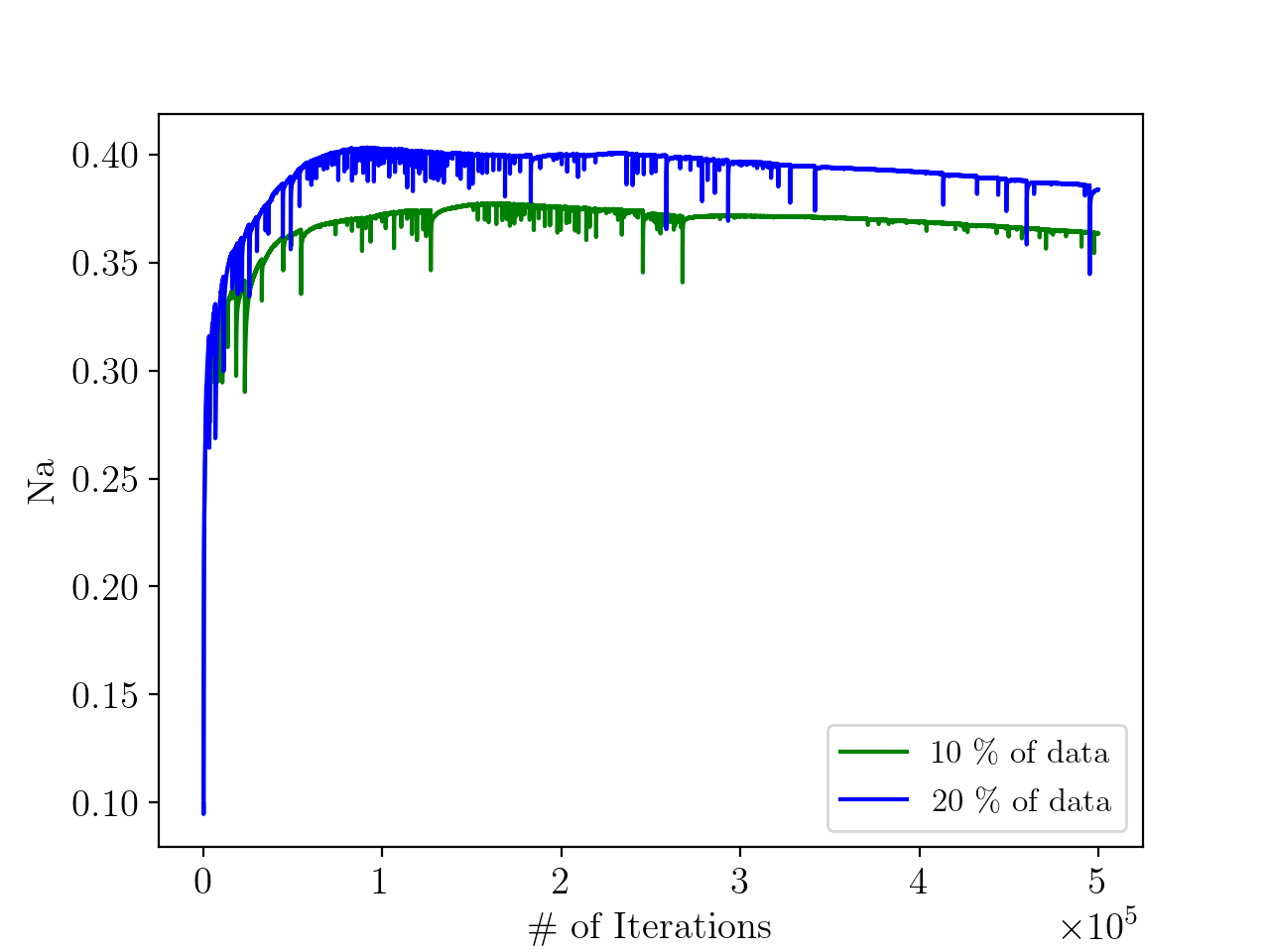}
}\\
\caption{Results from training of neural network model for subsampeld data acquired at incidence angle of $0^o$ , where (a) and (b) represents the snapshot of the particle displacement from using 10 \% and 20\% of data, respectively. (c) and (d) represents the speed of wave discovered from PINN simulation using 10\% and 20\% of total data, respectively. (e) shows the comparison of loss function for data usage of 10\% and 20 \%. (f) shows variation in $a$ with $n=10$ for the both sampling cases. }
\end{figure}
Finally, we train the PINN using the data acquired at $90^o$ of incidence angle and results are presented in Figure 8. The PINN and velocity network are a fully connected network with width of 32 neurons and depth of 4 layers with an initial learning rate of $5\text{e}-4$ are used. The input data to the network is comprised of 120 snapshots and each snapshot is sampled with 20\% of total points from entire domain i.e $(12~\text{mm} \times 12~\text{mm})$. The sub-figures of Figure 8 are subject to the same interpretation as those in Figure 7. A video (movie$\_$ds3.mp4) showing the real and PINN simulated data against the time stamp is uploaded as an additional material. The location of crack is clearly represented by the streak of low speed in Figure 7d, where the speed decreases to 2.4 $\text{mm}/\mu \text{s}$ from the surrounding value of 2.9 $\text{mm}/\mu \text{s}$, which is not as strong as the cases for $0^o$ and $45^o$. This could be due to the fact that effects of back-scattering are not as strong as those for other sets of data. 

Table 1 summarizes the hyperparameters used for all computational experiments, shown in Figure 4-8. To demonstrate the efficiency of PCA, we represent a PCA filtered data, PINN simulated data using filtered data and relative error between then in Figure 9a, b, and c, respectively. The relative error between filtered and PINN simulated data is $17.88 \%$. Subsequently, we also recovered the PINN simulated data using raw (non-filtered) data. Figures 9d, 9e and 9f represent the raw, PINN simulated data and error between raw and PINN simulated,respectively. The relative error using raw data is $24.80\%$ which is greater than error computed from the PCA filtered data. Thus, PCA helps here in achieving the greater accuracy.   

To test the requirement of minimum data size used for training of PINN, we further sub-sampled the data and used only 10\% of the data from each snapshot and trained the PINN. The network architecture and hyper-parameters used are the same as those used to generate Figure 6. The results from the sub-sampling process are shown in Figure 10. Figures 10a and 10b represent the snapshots of wavefield recovered from PINN model using 10\% and 20\% of data, respectively. This shows a very good similarity. The speed $v(x,y)$ from 10\% and 20\% data is shown in Figure 10c and d, respectively. The crack in both images can be seen very convincingly. The loss functions from both data sets are plotted in Figure 10e. Figure 10e shows that the convergence with 10\% and 20\% of the data is very similar. The variation of $a$ against the epoch is shown in the Figure 10f. 

\section{Summary}
In this study, we delineated the crack based on the zone marked by low sound speed. Though the data used in this study was acquired at a high frequency of $5 \text{MHz}$ but as reported by Blackshire et al. \cite{blackshire2002near} the effect of dispersion resulting into frequency dependent velocity and attenuation will not be a concern for or the aluminum alloy material system. That could be the case for much higher frequencies. The decrease in the sound speed in the crack zone could be due to the near-field crack feature interactions of wave field with the crack. The acoustic wave equation used in this study is primarily used to model the wave propagation in time domain and to study the effect of near-field and far-field interaction a frequency domain representation of acoustic wave equation will be more useful.   
 
The ultrasound nondestructive quantification and analysis for various material science problems is very important. The PINN model could be very helpful to solve such problems, as acquiring the real data for such analysis is prohibited by cost and logistical issues. To support our claim, we introduced an optimized PINN model to characterize a surface breaking crack in an aluminum alloy substrate material. In the current paper, we hypothesize that the speed of waves, propagating on the surface of a material, could be key indicator for crack identification. To prove our hypothesis, we designed and tested a deep neural network informed by the physics of acoustic wave propagation. The PINN proposed in this study is designed for inferring the system by discovering the data driven PDE. This is achieved by computing the space dependent speed of the wave in the metal plate. The data used in this study was acquired at incidence angles of $0^o$, $45^o$ and $90^o$. The results clearly show that the presence of the crack generate lower effective speeds in the crack affected zone. This lower speed is due to the back-scattering of waves from the crack, which eventually result in a loss of wave energy. The PINN model accurately predicts the speed of the wave in metal plate, not affected by crack, as 2.9 $\text{mm} / \mu \text{s}$ which is corroborated by the data base of material properties estimated from non-destructive testing. The results shown in the present study used only 10-20\% of the total data, which provided an accuracy of  1.2-1\% error. In order to reduce the cost of acquiring the data, we also have shown the results from further sub-sampling by reducing the size of input data from 20\% to 10\%.

\section{Acknowledgment}
The work is supported by DARPA-AIRA grant HR00111990025. This research was conducted using computational resources and services at the Center for Computation and Visualization, Brown University. KS would like to acknowledge Dr. Helen Kershaw from CCV, Brown University for providing the help and feedback at various stages of the research especially for problems concerning to the computation. 

\begin{table}
\label{Table: Hyperparameters}
\caption{Hyper-parameters of neural network}
\begin{tabular}{ |p{2cm}||p{1cm}|p{1cm}|p{2cm}| |p{2cm}| |p{2cm}| |p{2cm}| |p{1cm}| |p{1cm}| }
 \hline
 \multicolumn{8}{|c|}{Hyperparmeters used in this study} \\
 \hline
 Reference & Depth & Width & \# of epochs $( \times 10^5 )$ & Activation function & Learning rate & \% of data used & $\lambda$\\
 \hline
 Figure 4   &  2 & 32  & 2 & $\sin$ & $5\text{e}-4$ & 20\% & 100 \\
 \hline
 Figure 5   &  4 & 96  & 15 & $\sin$ & $5\text{e}-4$ & 40\% & 100 \\
 \hline
 Figure 6   &  4 & 64  & 5 & $\tanh$ & $5\text{e}-4$ & 20\% & 100 \\
 \hline
 Figure 7   &  6 & 32  & 10 & $\tanh$ & $5\text{e}-4$ & 10\% & 100\\
 \hline
 Figure 8   &  4 & 32  & 20 & $\tanh$ & $5\text{e}-4$ & 20\% & 100\\
  \hline
\end{tabular}
\end{table}
\bibliographystyle{unsrt}
\bibliography{references} 
\appendix
\section{Python code snippet of PCA process}\label{PCA1}
\lstset{language=Python}
\lstset{frame=lines}
\lstset{caption={PCA of wave field data}}
\lstset{label={lst:code_direct}}
\lstset{basicstyle=\footnotesize}
\begin{lstlisting}
import numpy as np 
import scipy
from sklearn.decomposition import PCA
# Routine for choosing the number of components
def nComponent(d, maxTh):
    wvf=scipy.io.loadmat('wvf.mat')['wvf']
    pca=PCA(n_components=240)
    pca.fit(d)
    variance=np.cumsum(pca.explained_variance_ratio_)
    nComp=0
    for i in range(0, len(variance)):
        if (variance[i]-0.95*max(variance)) < eps:
            nComp +=1
    return nComp
    
# Routine for filtering the data and constructing the filtered data
def pcaWd(d, nComponent):
    nComp=nComponent(d, maxTh)
    pca = PCA(n_components=nComp) 
    d1_pca = pca.fit_transform(d1)
    inv_pca = pca.inverse_transform(d1_pca)
    return inv_pca
\end{lstlisting}

\newpage
\section{PCA filtered data acquired at $0^o$ and $90^0$}
\renewcommand{\thefigure}{B\arabic{figure}}
\setcounter{figure}{0}
\begin{figure}[H]
\centering
\subfloat[Actual data at $t=13.2~\mu s$.]{
\includegraphics[trim=0cm 0cm 0cm 0cm, clip, width=0.50\textwidth]{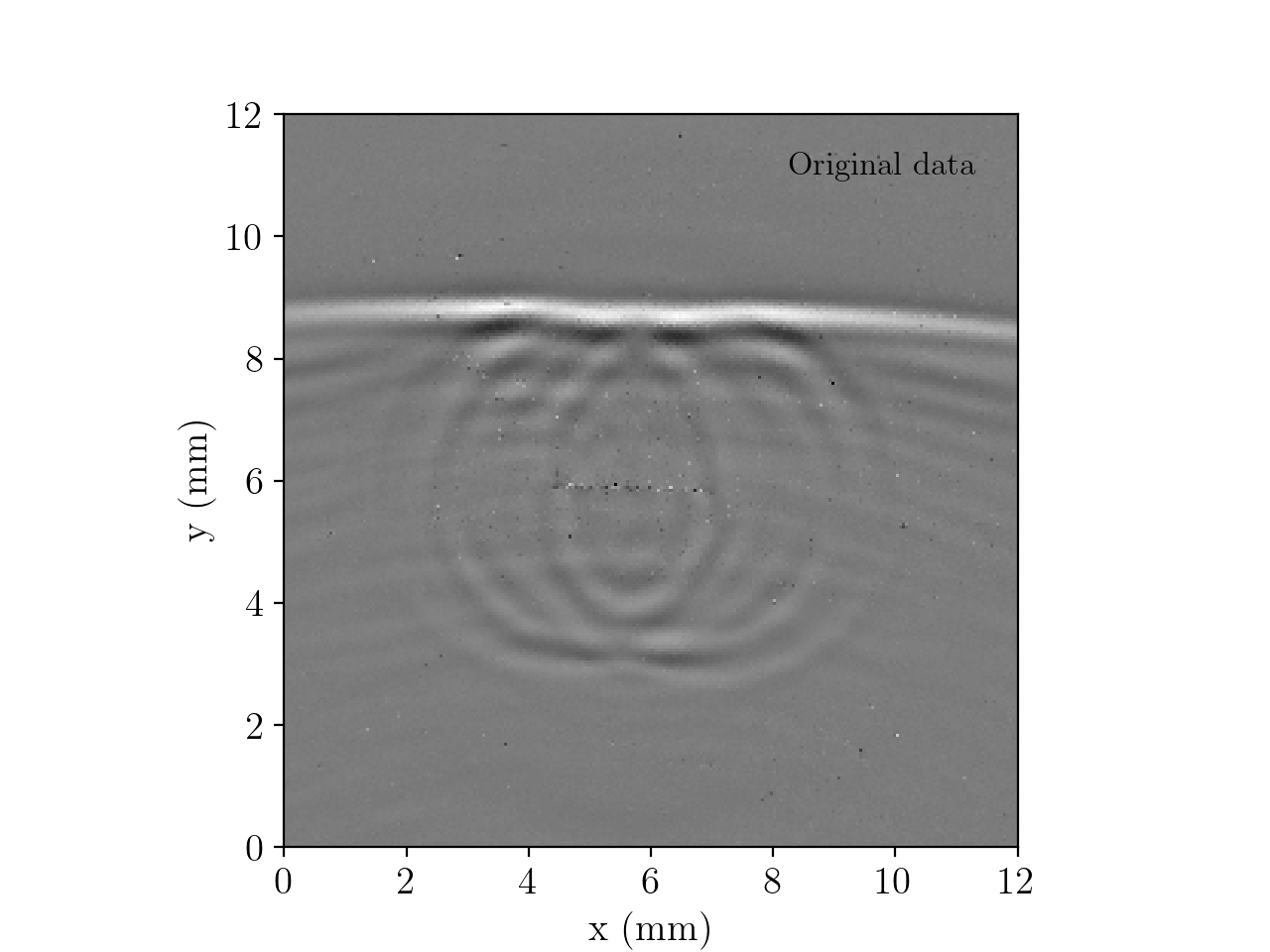}
}
\subfloat[Cumulative explained variance of (a).]{
\includegraphics[trim=0cm 0cm 0cm 0cm, clip, width=0.50\textwidth]{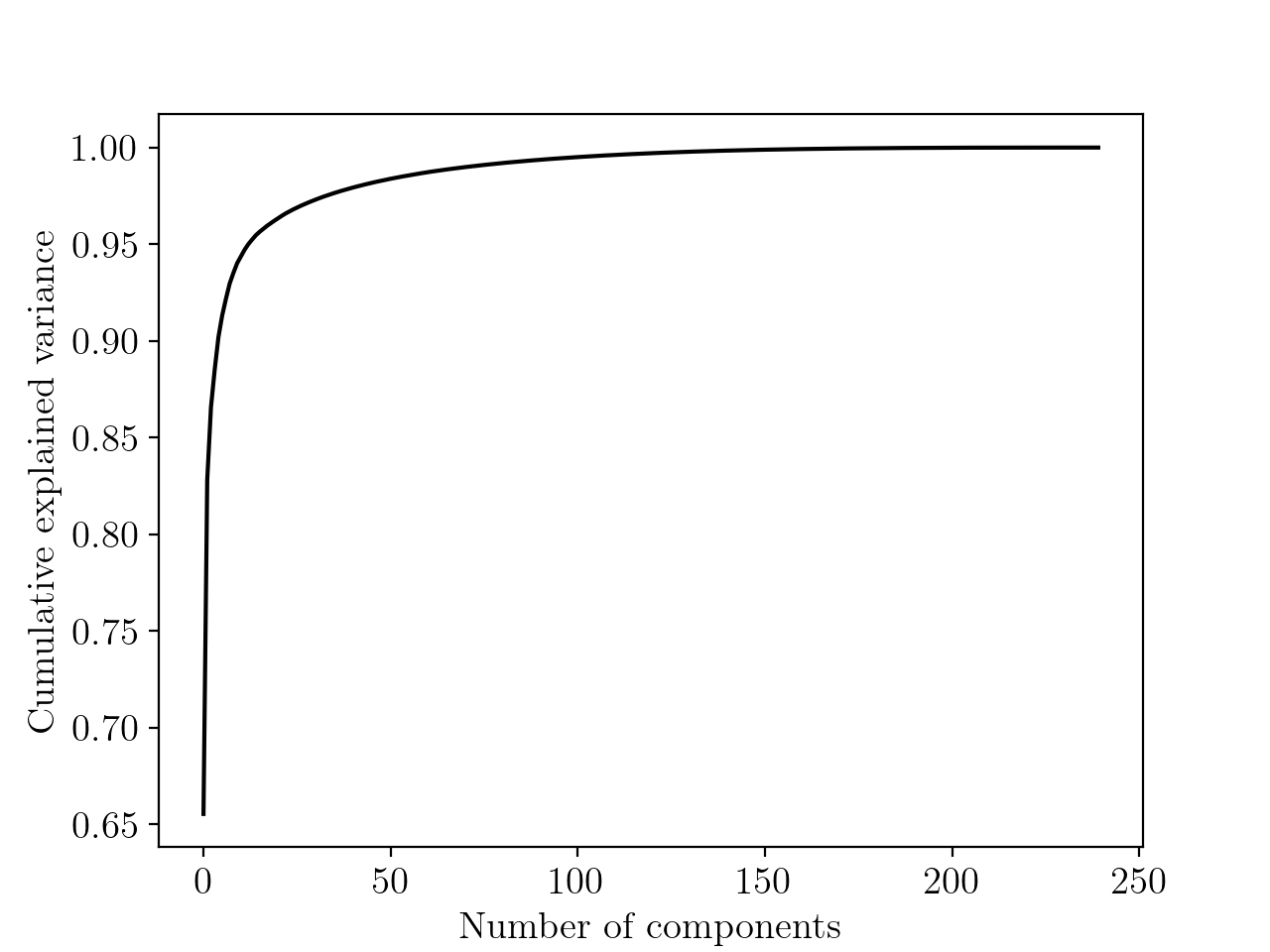}
}\\
\subfloat[Filtered data at $t=13.2~\mu s$.]{
\includegraphics[trim=0cm 0cm 0cm 0cm, clip, width=0.50\textwidth]{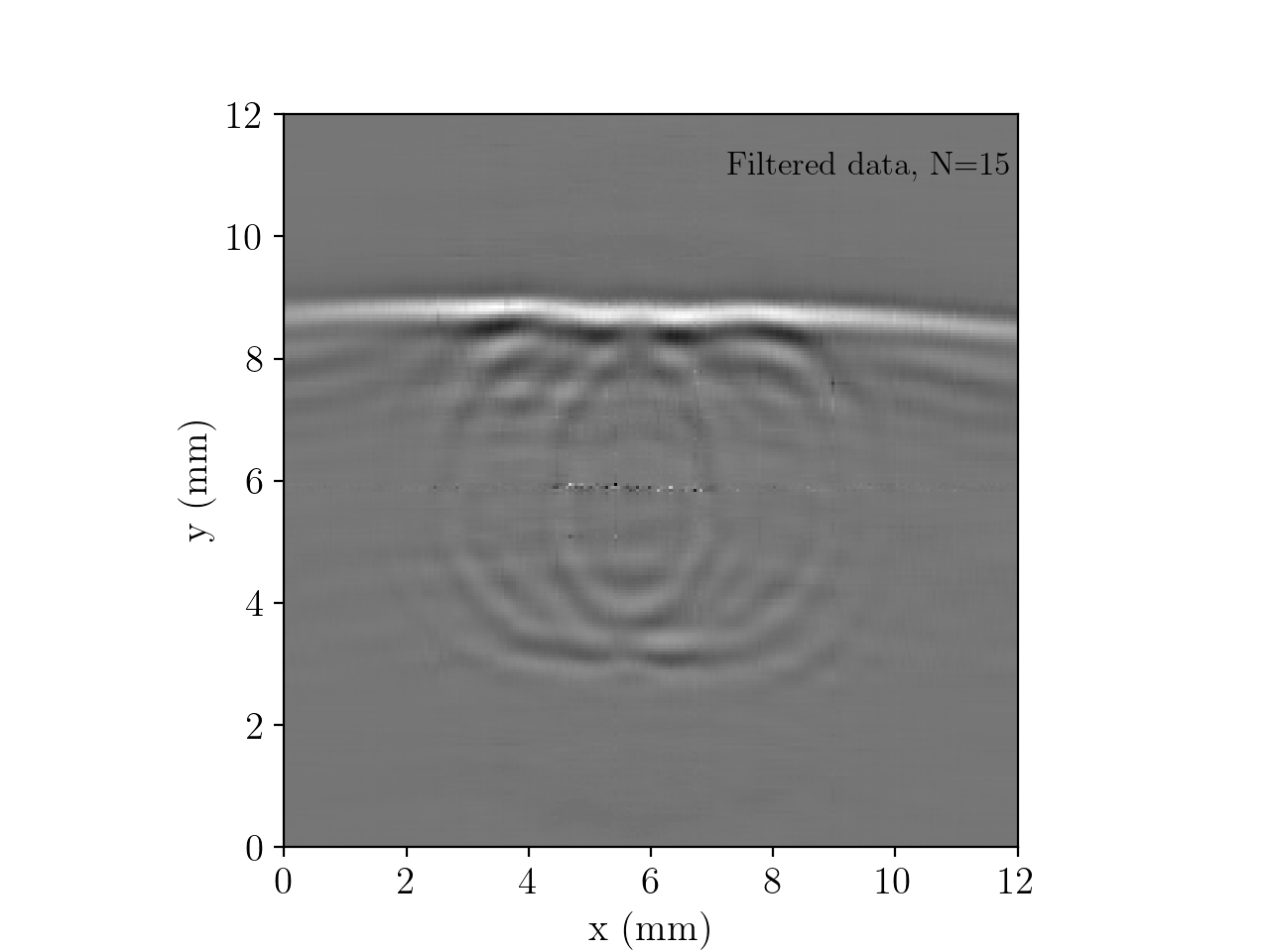}
}
\subfloat[Original vs filtered trace at $x=7~\text{mm}$ at $t=13.2~\mu s$.]{
\includegraphics[trim=0cm 0cm 0cm 0cm, clip, width=0.50\textwidth]{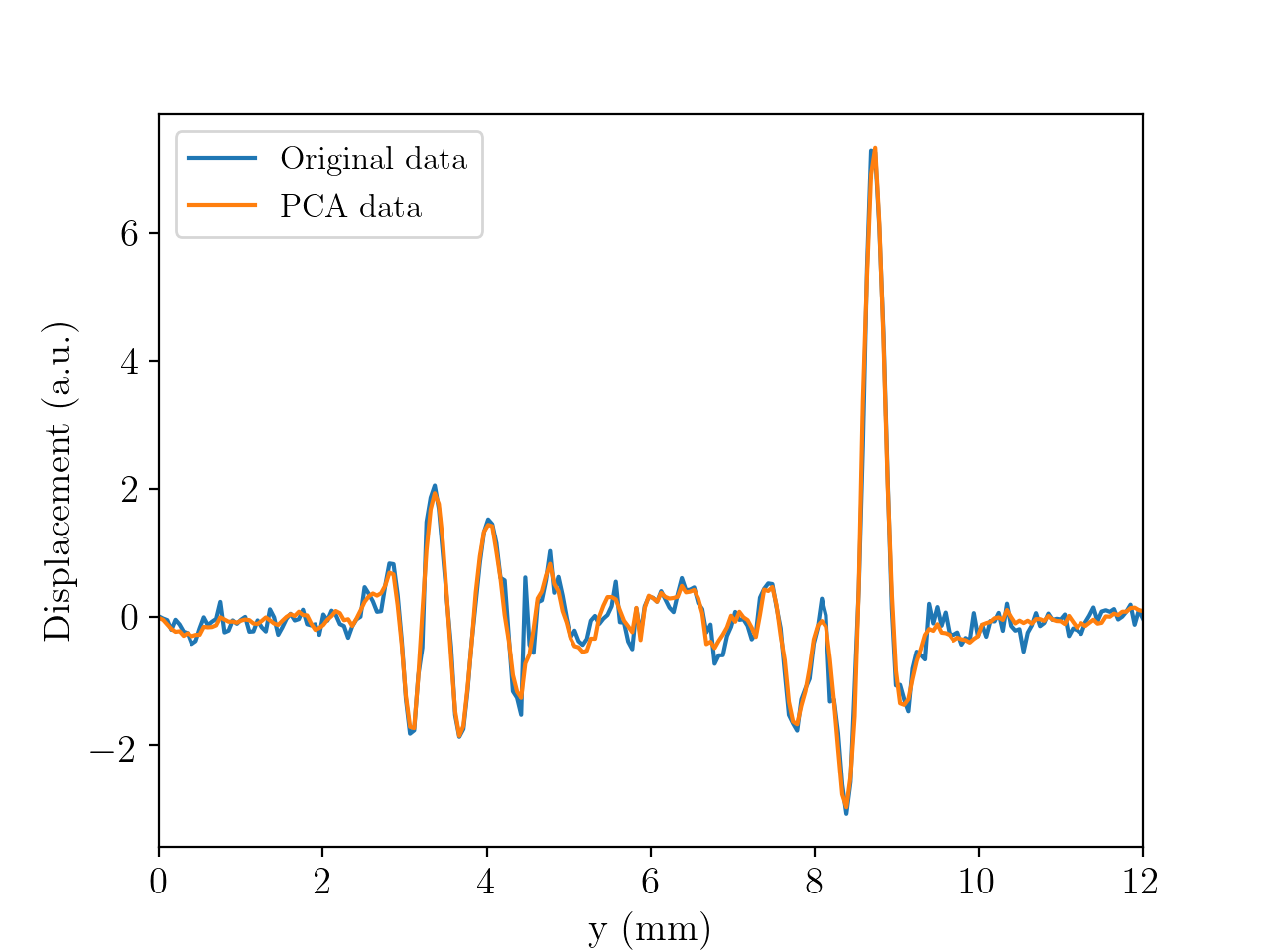}
}
\caption{Filtering of wavefield data acquired at $0^o$, using principle component analysis (PCA). PCA requires only the first 15 components to construct the original data by zeroing out the insignificant components.}
\end{figure}

\newpage
\begin{figure}[H]
\centering
\subfloat[Actual data at $t=11.38~\mu s$.]{
\includegraphics[trim=0cm 0cm 0cm 0cm, clip, width=0.5\textwidth]{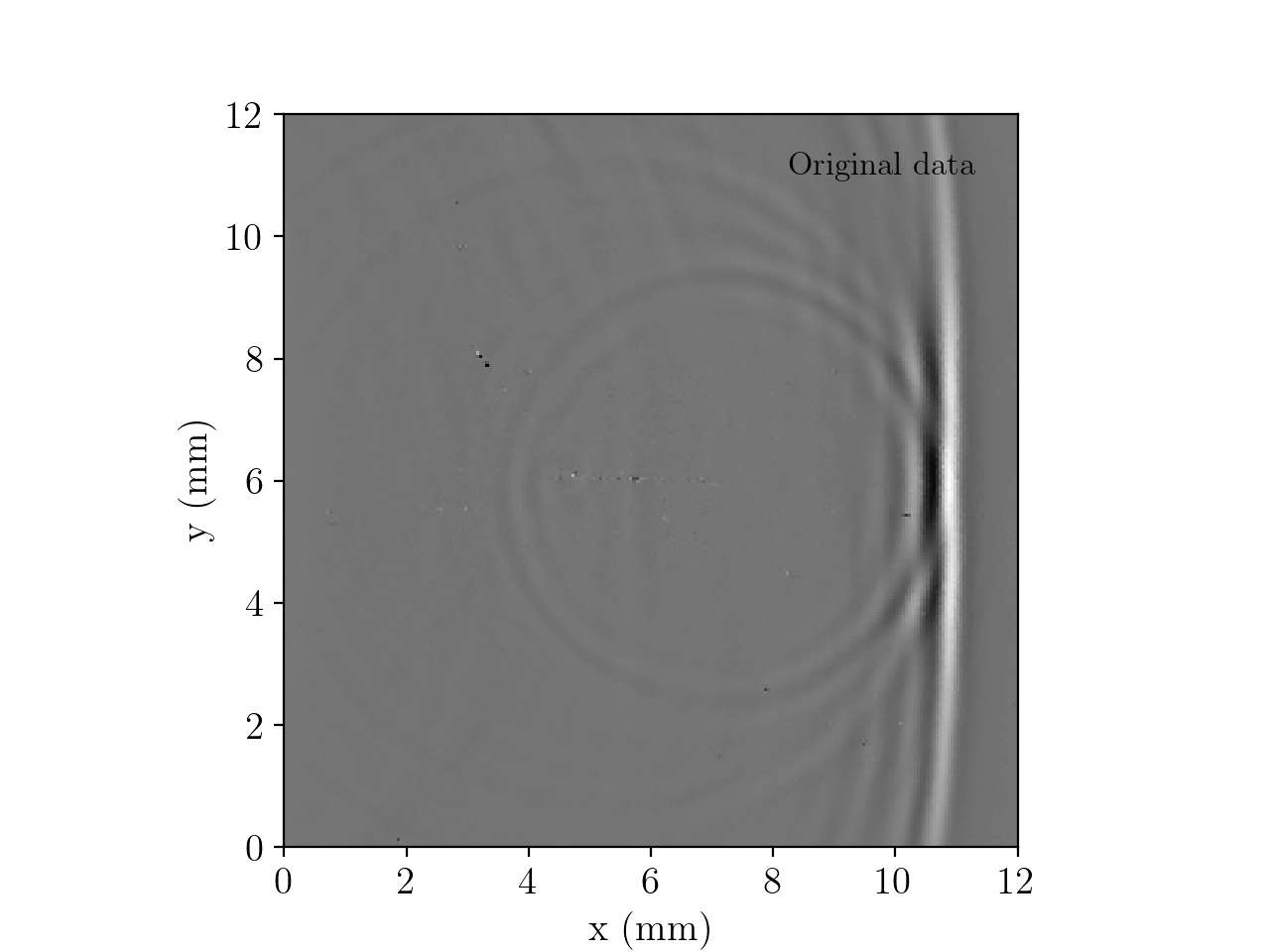}
}
\subfloat[Cumulative explained variance of (a).]{
\includegraphics[trim=0cm 0cm 0cm 0cm, clip, width=0.5\textwidth]{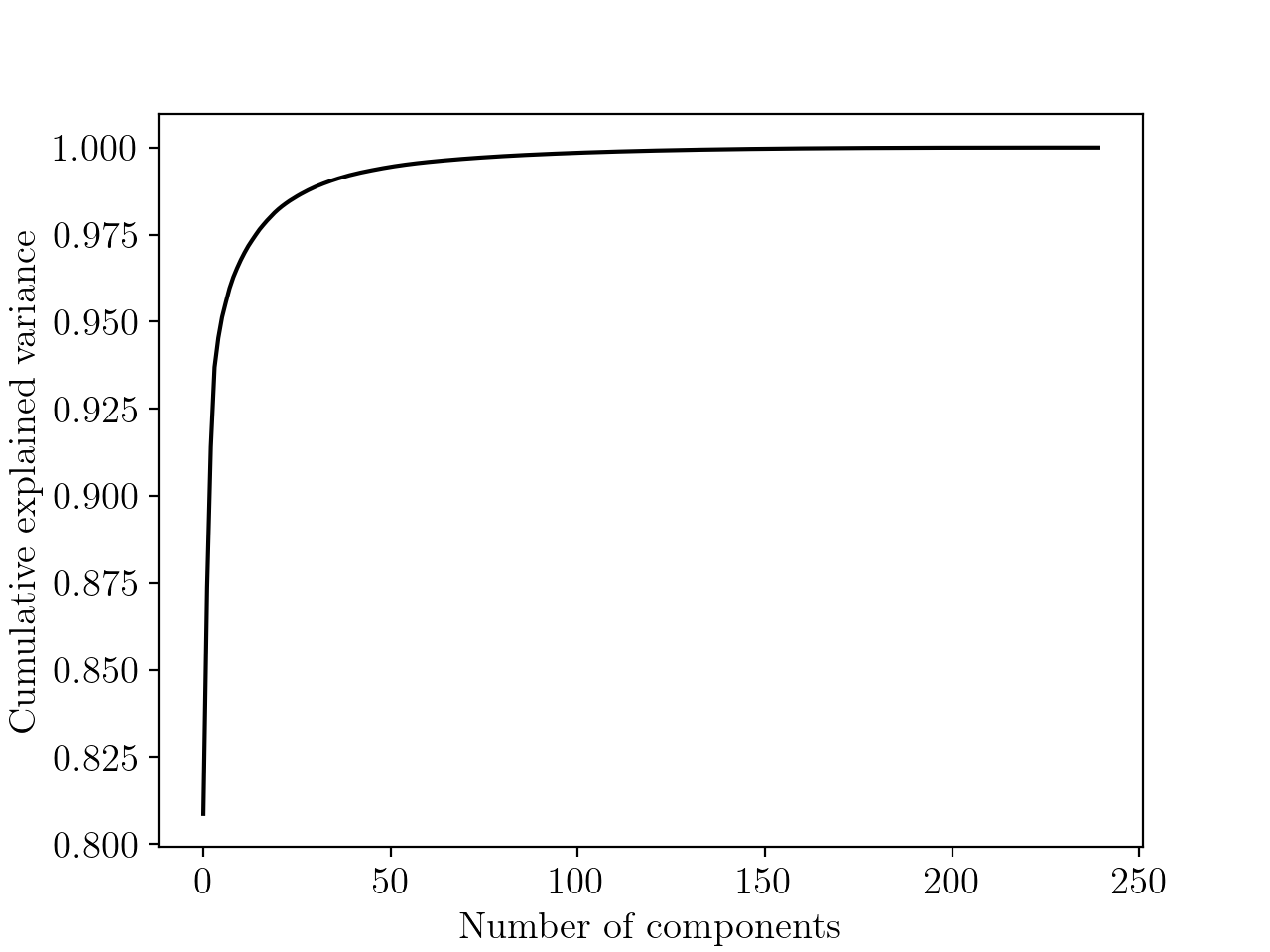}
}\\
\subfloat[Filtered data using at $t=11.38~\mu s$]{
\includegraphics[trim=0cm 0cm 0cm 0cm, clip, width=0.5\textwidth]{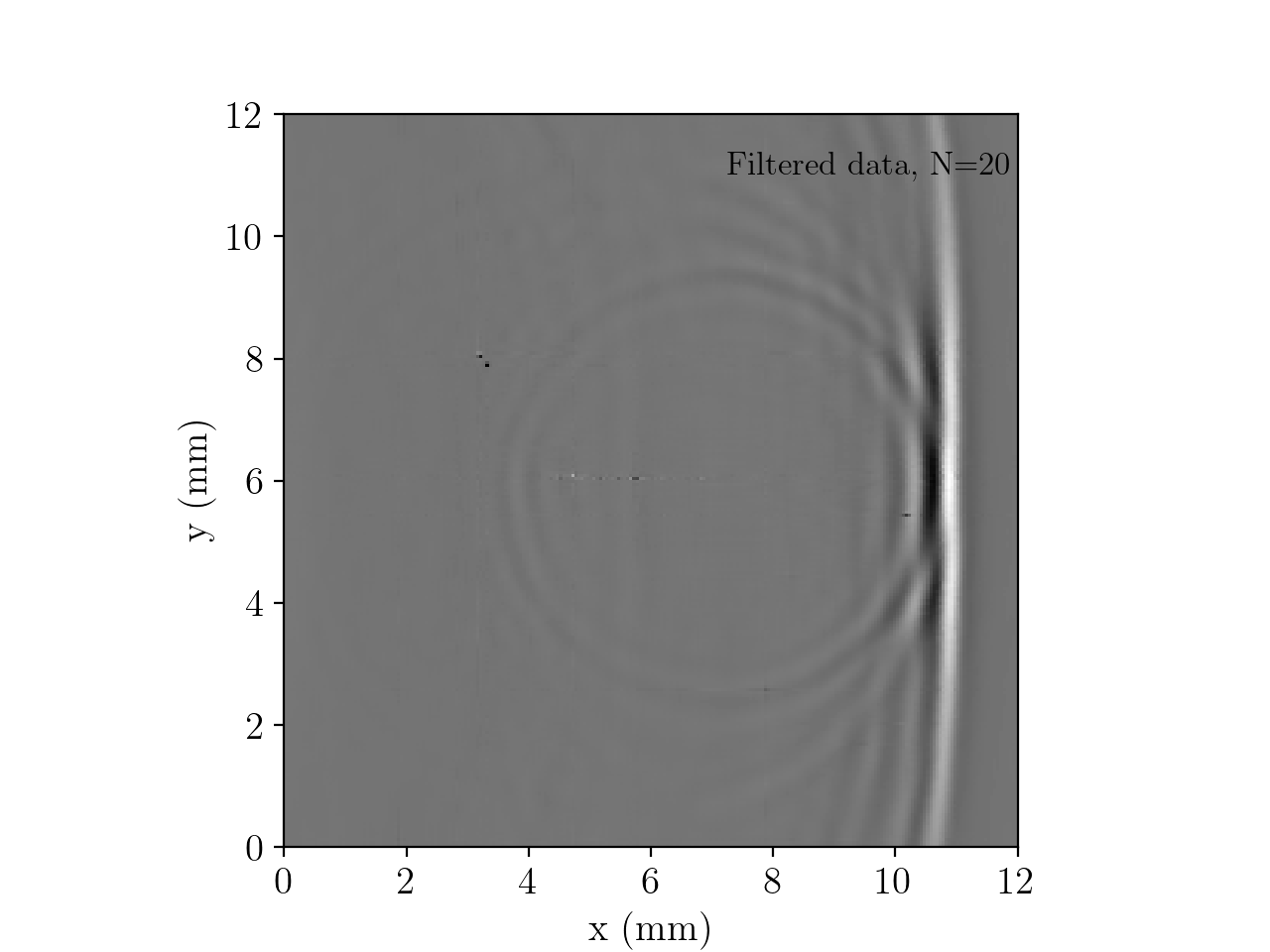}
}
\subfloat[Original vs filtered trace at $x=4.5~\text{mm}$ at $t=11.38~\mu s$.]{
\includegraphics[trim=0cm 0cm 0cm 0cm, clip, width=0.5\textwidth]{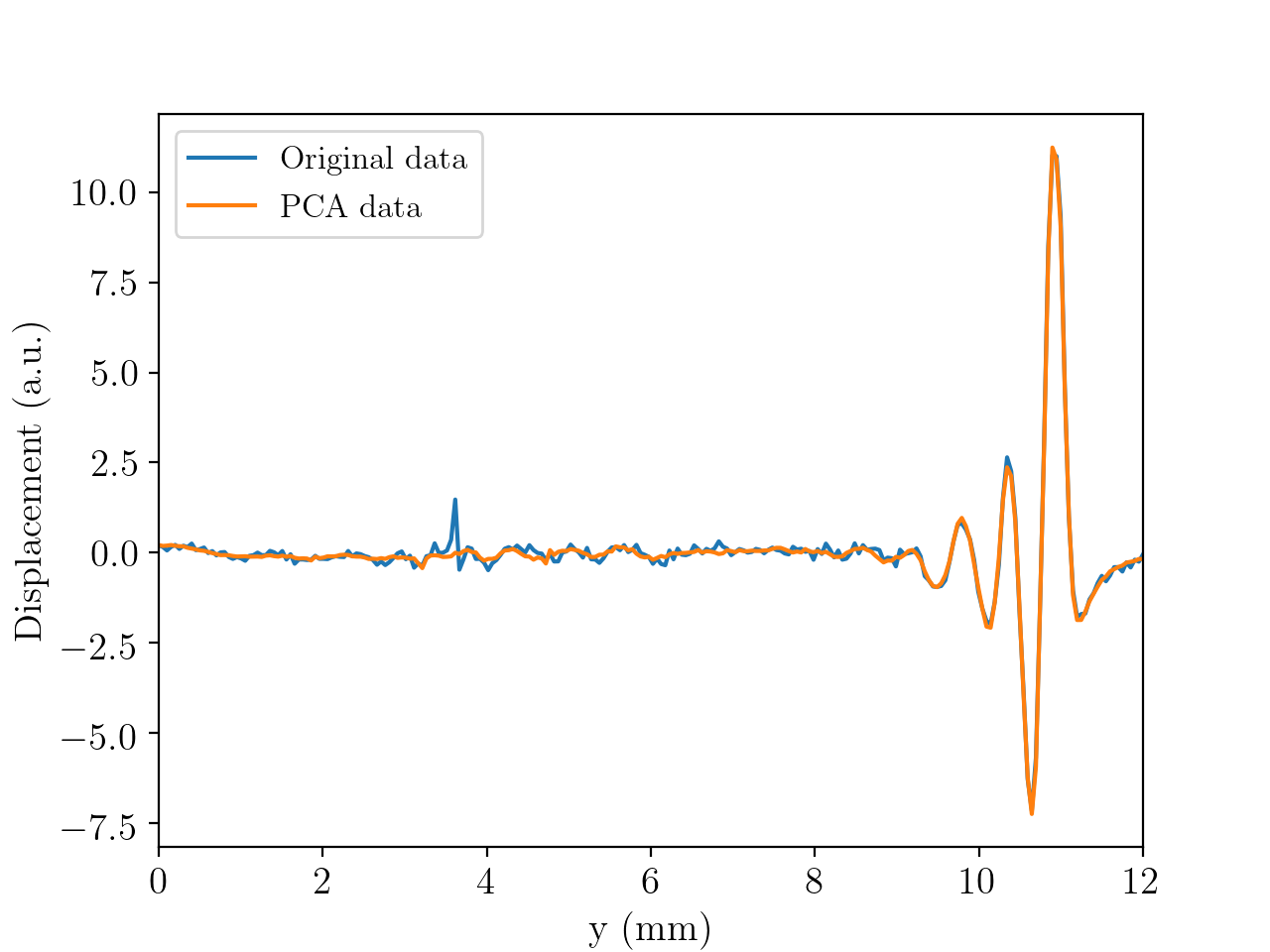}
}
\caption{Filtering of the wavefield data acquired at $90^o$, using principle component analysis (PCA). PCA requires only first 20 components to construct the original data by zeroing out the insignificant components.}
\end{figure}
\end{document}